\documentclass[journal=jcisd8,manuscript=article]{achemso}

\usepackage[T1]{fontenc}
\usepackage[hidelinks]{hyperref}     
\usepackage{url}            
\usepackage{booktabs}       
\usepackage{amsfonts}       
\usepackage{nicefrac}       
\usepackage{microtype}      
\usepackage{xcolor}         
 \usepackage{amsmath}
 \usepackage{bm}
 \usepackage{tablefootnote}
 \usepackage{float}
 \usepackage{graphicx}
 \usepackage{subcaption}
\usepackage{multirow}
\usepackage{filecontents}

\SectionNumbersOn

\usepackage{xr}
\title[FSscore]{FSscore: A Personalized Machine Learning-based Synthetic Feasibility Score}

\author{Rebecca M. Neeser}
\email{{rebecca.neeser,bruno.correia,philippe.schwaller}@epfl.ch}
\affiliation{Laboratory of Artificial Chemical Intelligence (LIAC), EPFL, Switzerland}
\alsoaffiliation{Laboratory of Protein Design and Immunoengineering (LPDI), EPFL, Switzerland}
\author{Bruno Correia}
\affiliation{Laboratory of Protein Design and Immunoengineering (LPDI), EPFL, Switzerland}
\author{Philippe Schwaller}
\affiliation{Laboratory of Artificial Chemical Intelligence (LIAC), EPFL, Switzerland}
\alsoaffiliation{National Centre of Competence in Research (NCCR) Catalysis, EPFL, Switzerland}

\begin{document}

\maketitle

\begin{abstract}
Determining whether a molecule can be synthesized is crucial in chemistry and drug discovery, as it guides experimental prioritization and molecule ranking in \textit{de novo} design tasks. Existing scoring approaches to assess synthetic feasibility struggle to extrapolate to new chemical spaces or fail to discriminate based on subtle differences such as chirality. This work addresses these limitations by introducing the Focused Synthesizability score~(FSscore), which uses machine learning to rank structures based on their relative ease of synthesis. First, a baseline trained on an extensive set of reactant-product pairs is established, which is then refined with expert human feedback tailored to specific chemical spaces. This targeted fine-tuning improves performance on these chemical scopes, enabling more accurate differentiation between molecules that are hard and easy to synthesize. The FSscore showcases how a human-in-the-loop framework can be utilized to optimize the assessment of synthetic feasibility for various chemical applications.
\end{abstract}

\section{Introduction}
\label{sec:intro}
Assessing the synthetic feasibility of a small molecule is of great importance in many different areas of chemistry, notably in early drug discovery stages. Trained chemists traditionally perform this task through intuition or retrosynthetic analysis, allowing them to decide which molecules are likely possible to synthesize and prioritize based on synthetic complexity. However, the chemical space that might be accessible is massive, and only a small fraction has been explored.~\cite{polishchuk2013estimation, reymond2010chemical} Furthermore, computational approaches such as virtual screening~(VS)~\cite{walters_virtual_1998} in drug discovery or the recent surge of generative methods for \textit{de novo} molecular design~\cite{olivecrona_molecular_2017,gomez-bombarelli_automatic_2018,NEURIPS2018_d60678e8,blaschke_reinvent_2020,yang_syntalinker_2020,de_cao_molgan_2022,peng_pocket2mol_2022,guo_link-invent_2022,adams_equivariant_2022,schneuing_structure-based_2022, igashov_equivariant_2022} emphasize the requirement for suitable tools to score synthetic feasibility quickly in an automated fashion.~\cite{bilodeau_generative_2022, baillif_deep_2023}.\par 

 The current state-of-the-art methodologies perform well at discriminating feasible from unfeasible molecules in the data distribution they were designed for but often fail to generalize. This is especially true for machine learning~(ML) predictors that cannot capture such an abstract concept as synthesizability and when applying these scores in the context of generative models.~\cite{bilodeau_generative_2022, skoraczynski_critical_2023, ivanenkov_hitchhikers_2023, gao_synthesizability_2020} However, exploring new chemical space is of great interest specifically in the context of \textit{de novo} design or new drug modalities such as synthetic macrocycles or proteolysis targeting chimeras~(PROTACs). On the other hand, synthetic feasibility cannot be merely captured by the structure as it depends on a chemist's available resources and expertise.~\cite{stanley_fake_2023} Thus, incorporating human preference would greatly improve the practical utility of such a score.~\cite{choung_learning_2023}\par 

Various methods to capture synthetic accessibility or complexity have been previously proposed and include structure-based \cite{ertl_estimation_2009, vorsilak_syba_2020, yu_organic_2022} or reaction-based \cite{coley_scscore_2018, thakkar_retrosynthetic_2021, li_prediction_2022, liu_retrognn_2022, kim_dfrscore_2023} methods. The commonly used Synthetic Accessibility score (SA score) is rule-based and penalizes the occurrence of fragments rarely found in a reference dataset and the presence of specific structural features.~\cite{ertl_estimation_2009} Thus, it captures more synthetic complexity than accessibility and fails to identify big complex molecules with mostly reasonable fragments.~\cite{bilodeau_generative_2022, stanley_fake_2023} The SYBA score was trained to distinguish existing synthesizable molecules from artificial complex ones but the performance was found to be sub-optimal.~\cite{skoraczynski_critical_2023, vorsilak_syba_2020} Many of these structure-/fragment-based approaches are unable to capture small structural differences due to low sensitivity. \citet{yu_organic_2022} attempt to address this by using a graph representation to classify molecules into hard~(HS) and easy~(ES) to synthesize similarly to SYBA.
The Synthetic Complexity score~(SCScore) predicts complexity in terms of required reaction steps and is based on 1024-bit Morgan fingerprints. The SCScore was trained on the assumption that reactants are easier to make than products.~\cite{coley_scscore_2018} This score performs well on benchmarks approximating the length of the predicted reaction path but poorly when predicting feasibility in benchmarks using synthesis predictors.~\cite{bilodeau_generative_2022} The Retrosynthetic Accessibility score~(RAscore) predicts synthetic feasibility with respect to a synthesis prediction tool and thus is directly dependent on the performance of the upstream model.~\cite{thakkar_retrosynthetic_2021} Similarly, RetroGNN classifies molecules based on retrosynthetic accessibility with the specific aim of being applied in VS.~\cite{liu_retrognn_2022} Li and Chen suggested a score based on a Communicative Message Passing Neural Network~(CMPNN), which aims at discriminating ES from HS based on number of reaction steps using a reaction knowledge graph.~\cite{li_prediction_2022} Recently, the Drug-Focused Retrosynthetic score~(DFRscore) was introduced by~\citet{kim_dfrscore_2023} predicting the number of reaction steps required based on a limited set of reaction templates relevant to drug discovery.\par 

The incorporation of human feedback in ML has gained increasing attention since the introduction of RL with Human Feedback~(RLHF)~\cite{christiano_deep_2017} by OpenAI, which lead to the development of popular tools such as InstructGPT~\cite{ouyang_training_2022} and ChatGPT~\cite{chatgpt}. 
Learning with human feedback has also found its way into applications in chemistry: ~\citet{sheridan_modeling_2014}. trained a random forest model to predict the complexity based on human rankings. Ranking individual molecules instead of preference labeling is likely to suffer from more bias, which might be reflected in the moderate correlations of the labels to the predicted score.~\cite{kahneman1984choices} This is avoided in the design of MolSkill, where the model learns to rank based on binary preferences made by medicinal chemists.~\cite{choung_learning_2023} However, the scored objective, general preference, is loosely defined, and substantial pre-filtering of the training data makes the model not applicable to synthesizability and likely fails to generalize well.\par

This work presents a novel approach to assess synthetic feasibility and investigates the incorporation of expert knowledge to tune the model towards a desired chemical space. We build on ideas first put forth by~\citet{coley_scscore_2018} when introducing the SCScore. As such, we use reaction data to pre-train our model, which implicitly informs on the difficulty of synthesizing a molecule through the relational nature of the data, while containing more information than just the number of reaction steps. Furthermore, by framing this task as a ranking problem based on comparisons between pairs of structures, we avoid the need for a ground truth score~(ie. an absolute synthesizability measure), but base the model on reported chemical reactions.~\cite{coley_scscore_2018, choung_learning_2023} To ultimately tune to a specific chemical space in low data regimes, we refined the unbiased baseline model with expert chemist-labeled data in an active-learning type framework as inspired by MolSkill \cite{choung_learning_2023}. Besides the various applications as an offline scoring tool, this fully differentiable approach could also be directly used as guidance in generative models or as a reward function in reinforcement learning~(RL) frameworks. Compared to previous work~\cite{coley_scscore_2018, vorsilak_syba_2020, thakkar_retrosynthetic_2021}, our representations consider stereochemistry and repeated substructures, which are crucial for determining the synthesizability of molecules.

Our main contributions are:
\begin{itemize}
    \item We propose a novel approach for assessing synthetic feasibility that uses pairwise preferences and fine-tuning with human feedback to focus the model on a desired chemical space. This allows for incorporating expert knowledge and intuition.
    \item The method is fully differentiable, allowing it to be easily incorporated into generative models.
    \item Using our FSscore fine-tuned to the chemical space of a generative model we show that we can sample at least 40\% synthesizable molecules (according to Chemspace\footnote{\url{https://chem-space.com}}), compared to only 17\% using the popular SA score, while maintaining good docking scores.
    \item Our experiments show the model can be effectively fine-tuned with relatively small amounts of human-labeled data, as little as 20-50~pairs, which is important for practicality.
    \item Fine-tuning improved performance separately on several chemical scopes, including PROTACs, demonstrating the approach’s ability to adapt to new domains.
\end{itemize}

\section*{Results and Discussion}
\label{sec:res_disc}

\begin{figure*}[!ht]
    \centering
    \includegraphics[width=14cm]{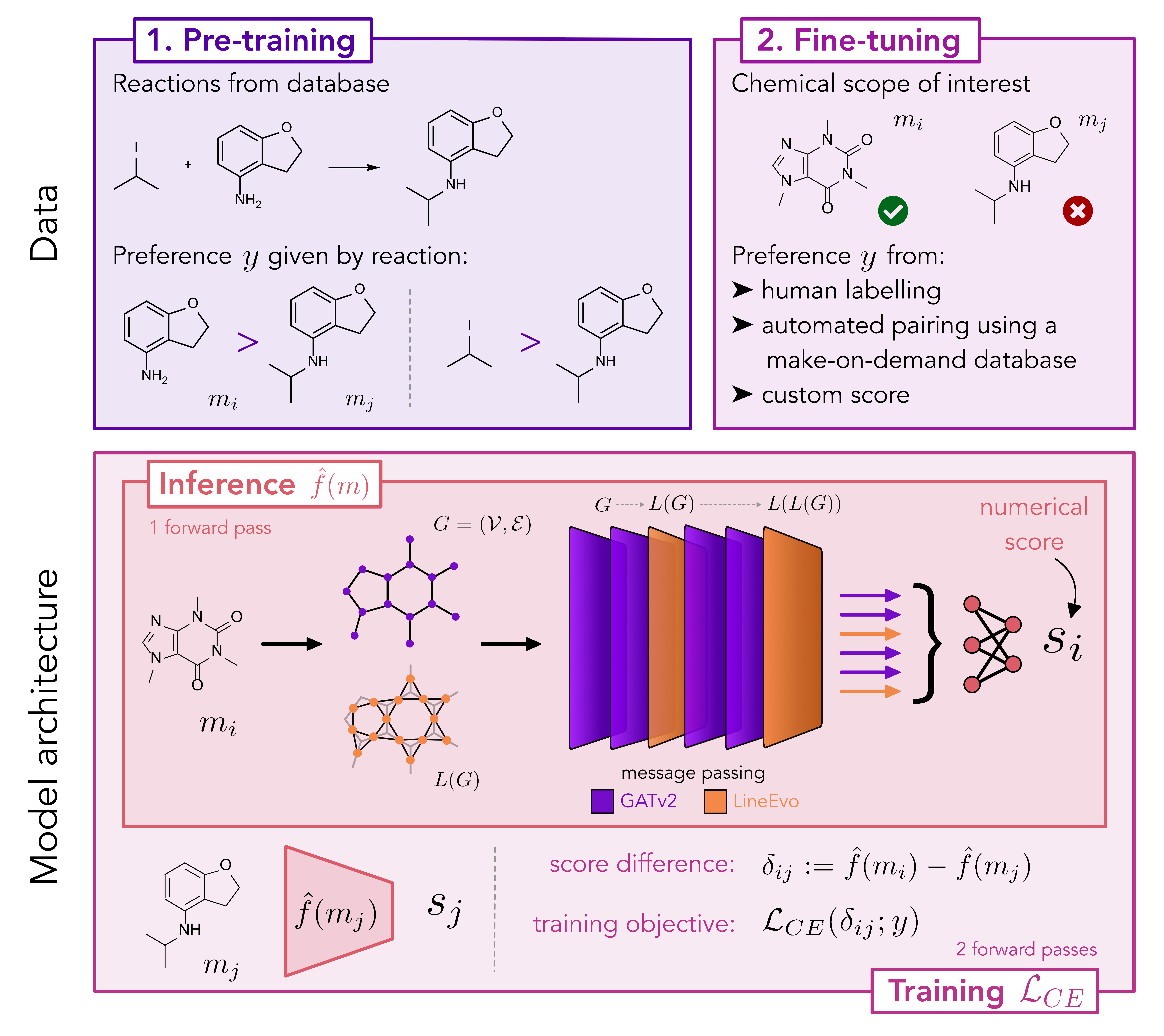} 
    \caption{Schematic overview of the data sources, architecture and training pipeline. For pre-training, data is extracted from reaction datasets, while fine-tuning is with a custom chemical scope of interest that can be labeled in various ways. Training the FSscore requires two forward passes for both molecules in the binary preference pair, while at inference time, one forward pass is sufficient to obtain the score.}
    \label{fig:scheme}
\end{figure*}


The novel ML-based Focused Synthesizability score (FSscore) is trained in two stages. First, we pre-train on a large dataset of reactions to establish an unbiased baseline assessing synthesizability using a graph representation and suitable message passing scheme with local attention (i.e. GATv2~\cite{brody2022how}) improving expressivity over similar frameworks such as the SCScore.~\cite{coley_scscore_2018} Secondly, we introduce our fine-tuning approach using human expert knowledge, allowing us to focus the score towards a specific chemical space of interest. Our approach to learning a continuous score to assess the synthesizability is inspired by ~\citet{choung_learning_2023}, who framed a similar task as a ranking problem using binary preferences, which avoids the need for a ground truth. Specifically, every data point consists of two molecules, for each of which we predict a scalar in separate forward passes (see Fig.~\ref{fig:scheme}). The model learns by minimizing the error~(ie. binary cross entropy) between the model's predicted preference inferred from the score difference and the observed preference from literature for each pair. During the pre-training stage, the pairs consist of reactants \textit{vs.} products, assuming increasing complexity during a reaction, while during fine-tuning these binary labels come from human experts, database look-ups or custom scores. Detailed information on the architecture and training can be found in the Appendix~\ref{sec:FS_meth}. \par 
We compare our approach with established scores, namely SA score, SCScore, SYBA, and RAscore, and report numerous metrics as described in Appendix~\ref{sec:SI_metrics}. We investigate the qualitative performance of the pre-trained model on the MOSES~\cite{polykovskiy_molecular_2020} and COCONUT~\cite{sorokina_coconut_2021} datasets. MOSES contains commercially available drug-like molecules and is often used to benchmark generative models for drug discovery and COCONUT is a collection of natural products extracted from various sources.~\cite{polykovskiy_molecular_2020, sorokina_coconut_2021}\par 

We showcase the model's ability to efficiently focus our score by fine-tuning on several datasets (see Appendix~\ref{sec:SI_data_ft}):
\begin{itemize}
    \item A subset from the pre-training set with chiral tetrahedral centers. The molecule with assigned chirality is labeled as more complex as opposed to the same molecule stripped of the respective assignment.
    \item The MC~(manually curated) and CP~(computationally picked) test sets published with the SYBA score. \cite{vorsilak_syba_2020} Labels are provided and correspond to ES and HS. 
    \item The meanComplexity dataset containing averaged complexity scores from chemists.~\cite{sheridan_modeling_2014} Binary labels are extracted from the continuous score [1,5] based on a set-off of at least~two.
    \item The PROTAC-DB~\cite{weng_protac-db_2022}, which is an open-source collection of PROTACs. Labels were obtained from human experts.
\end{itemize}

Additionally, the applicability to a generative modeling task is assessed using REINVENT with Augmented Memory~\cite{guo_augmented_2023, blaschke_reinvent_2020} as a molecular generator. For details consult Appendix~\ref{sec:reinvent}\par
All code used for training, fine-tuning, and scoring the FSscore is available at \url{https://github.com/schwallergroup/fsscore}. This repository also includes an application that can be run locally and allows a more accessible way to label data, fine-tune, and deploy a model. More information on the application can be found in Appendix~\ref{sec:app}.

\subsection{Base model}

The pre-trained model with varying algorithmic (graph neural network~(GNN) \textit{vs.} multilayer perceptron~(MLP)) and representation (graph \textit{vs.} fingerprint) implementations was evaluated on the hold-out test set (see Tab.~\ref{tab:test_performance}). The graph versions outperform fingerprint representations by a small difference and the best-performing model is used for further investigations. This corresponds to the GNN with GATv2 and LineEvo layers (GGLGGL). Furthermore, preliminary investigations with models trained on a random subset of 100k~data points showed weaknesses of the boolean Morgan fingerprint, resulting in a biased comparison. Not including the counts of fragments in the fingerprint leads to the inability to capture complexity based on recurrence, even if the single fragment is common in the training set. The original publication of the SCScore briefly touches on the impact of varying the Morgan fingerprint. Still, contrary to our results, they found this aspect to not influence the performance, leading them to choose the boolean 1024-bit vector.~\cite{coley_scscore_2018} Furthermore, results on chirality considerations discussed below supported the choice of the graph-based representation. Results with the count-based Morgan fingerprint can be found in the Appendix.\par

\begin{table}[b]
  \caption{Performance on the hold-out test set of graph-based models compared to various fingerprint implementations (binary/counts; \(+\)/\(-\) chirality). Accuracy~(\textit{Acc}) and \textit{AUC} based on the score differences are reported. The best-performing model is highlighted in bold. G~=~GATv2 layer; L~=~LineEvo layer}
  \centering
  \label{tab:test_performance}
  \begin{tabular}{llll}
    \toprule
    Model     & Representation     & \textit{Acc} \(\uparrow\) & \textit{AUC} \(\uparrow\) \\
    \midrule
    GNN \scriptsize{(GGLGGL)}    & graph                 &\bm{$0.905$}   &\bm{$0.971$} \\
    GNN \scriptsize{(GGG)}       & graph                 &0.903          &0.970 \\
    MLP             & binary                &0.87          &0.954  \\
    MLP             & counts         &0.880          &0.959  \\
    MLP             & binary chiral         &0.867         &0.952  \\
    MLP             & counts chiral      &0.875          &0.957  \\
    \bottomrule
  \end{tabular}
\end{table}

\begin{figure*}[t]
	\centering
	\begin{subfigure}[t]{0.59\textwidth}
		\centering
		\includegraphics[width=\textwidth]{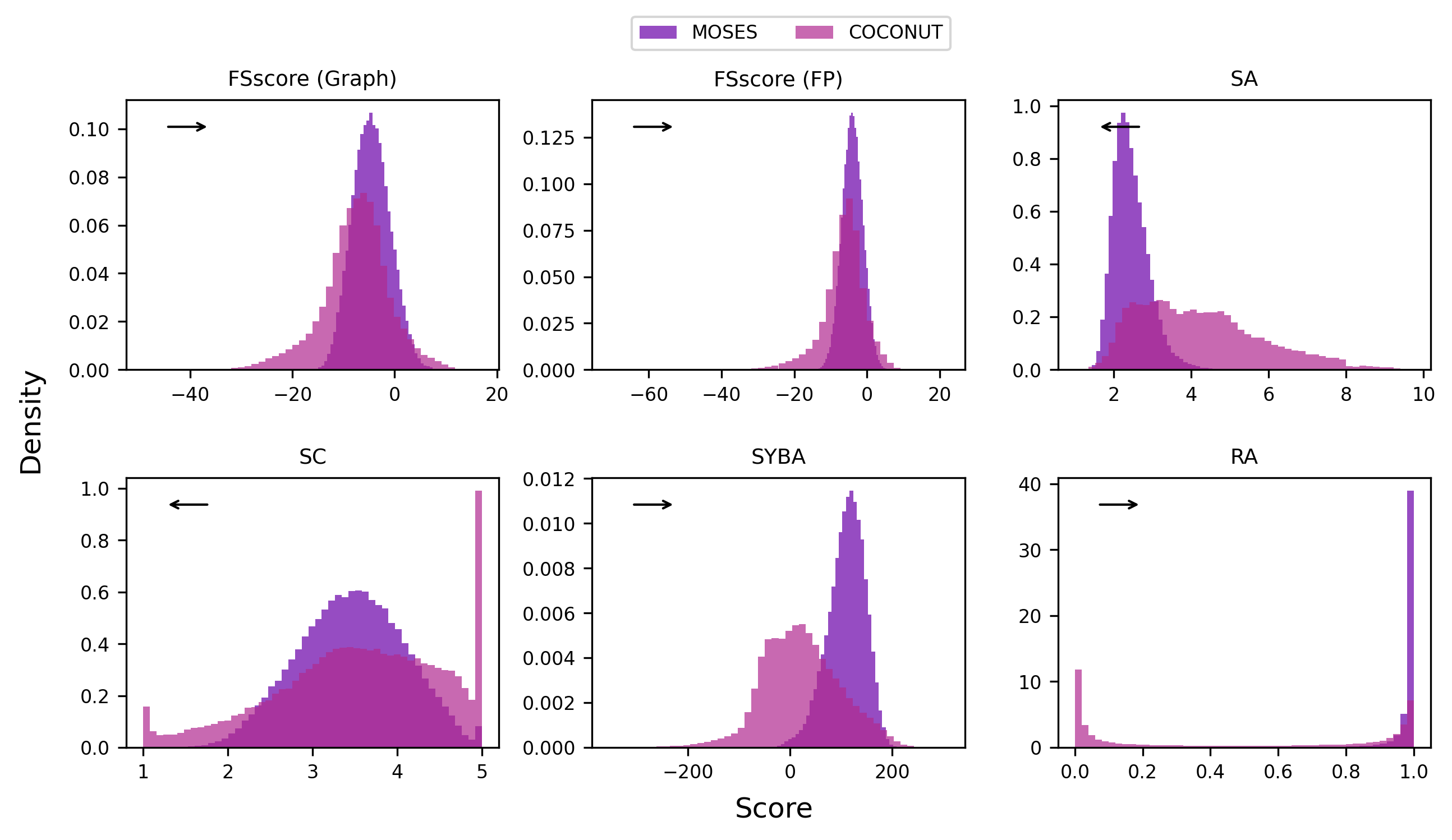}
		\caption{Score distributions.}
		\label{subfig:drugs_dist}
	\end{subfigure} \hfill
	\begin{subfigure}[t]{0.39\textwidth}
		\centering
		\includegraphics[width=0.9\textwidth]{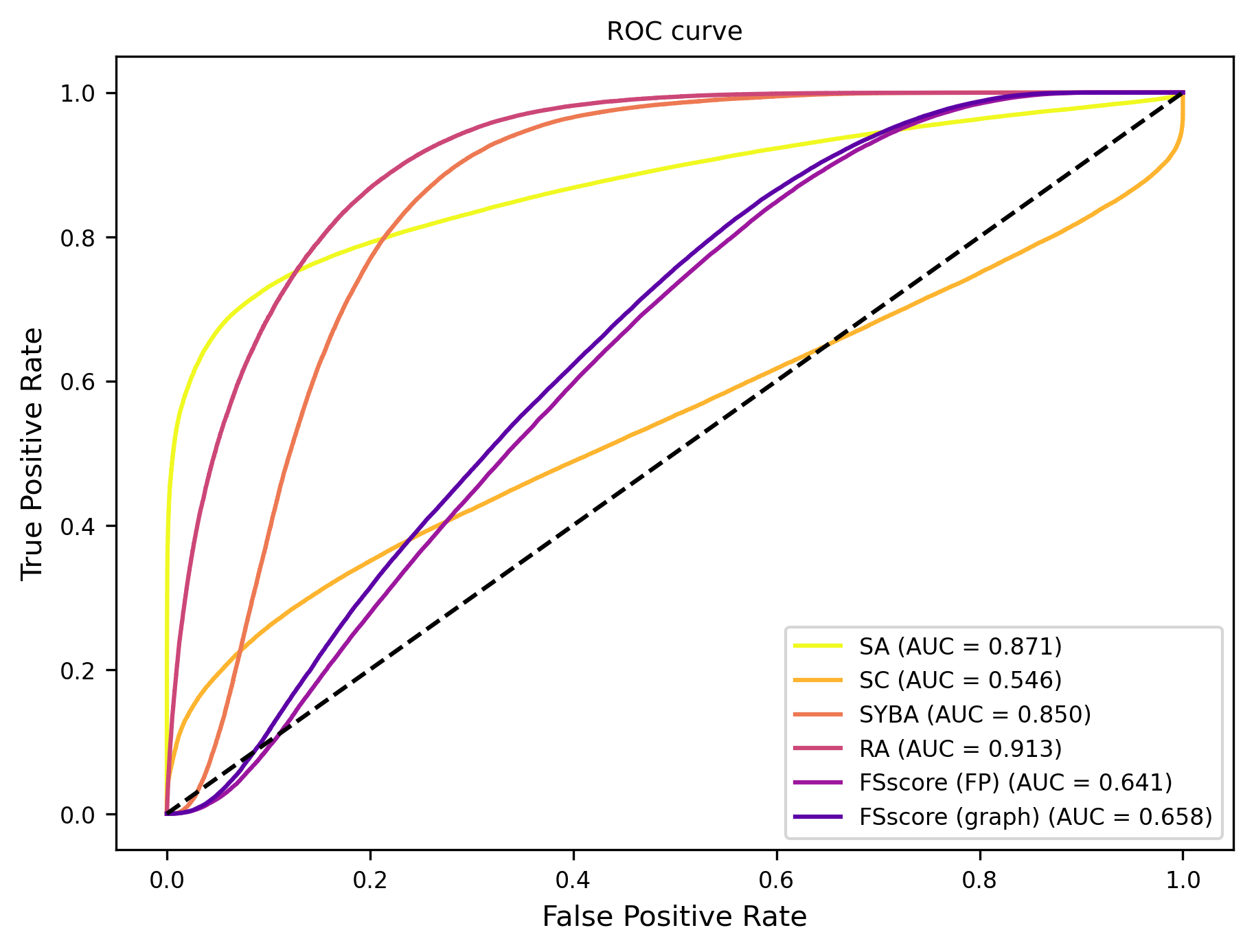}
		\caption{ROC curves.}
		\label{subfig:drugs_ROC}
	\end{subfigure}
	\caption{Results showcasing the ability to differentiate molecules originating from MOSES~\cite{polykovskiy_molecular_2020} from those in COCONUT~\cite{sorokina_coconut_2021}. The latter are expected to be more complex being natural products. The ROC curves in Figure~\ref{subfig:drugs_ROC} detail the power to discriminate MOSES~\cite{polykovskiy_molecular_2020} from COCONUT~\cite{sorokina_coconut_2021}. The arrows in the distribution plot (Fig.~\ref{subfig:drugs_dist}) indicate the direction of higher synthetic feasibility.}
	\label{fig:drugs}
\end{figure*}

The analysis of the pre-trained model's predictions on MOSES (known drugs)~\cite{polykovskiy_molecular_2020} and COCONUT (natural products)~\cite{sorokina_coconut_2021} shows that the pre-trained FSscore is unable to distinguish these molecule classes assuming that COCONUT~\cite{sorokina_coconut_2021} contains more complex structures (see Fig.~\ref{fig:drugs}). While both the fingerprint- and graph-based FSscore outperform the SCScore in the area under the receiver operating characteristic curve (\textit{AUC}), the SA score, SYBA and RAscore yield better results. The structures found in COCONUT~\cite{sorokina_coconut_2021} are likely out-of-distribution for the FSscore highlighting the opportunity for fine-tuning, but it also emphasizes the power of rule-based methods such as the SA score.

\subsection{Application-specific model refinement}

Given the absence of a ground truth and the experimental evaluation being disproportionate, we evaluated the refined (i.e. fine-tuned) model with several case studies. First, we show proof of concept with simple tasks. Next, the usefulness on "uncommon" chemical spaces is demonstrated, and subsequently, the application to generative modeling is showcased.

\subsubsection{Proof of concept case studies}
\begin{figure*}[!ht]
	\centering
	\begin{subfigure}[t]{0.59\textwidth}
		\centering
		\includegraphics[width=\textwidth]{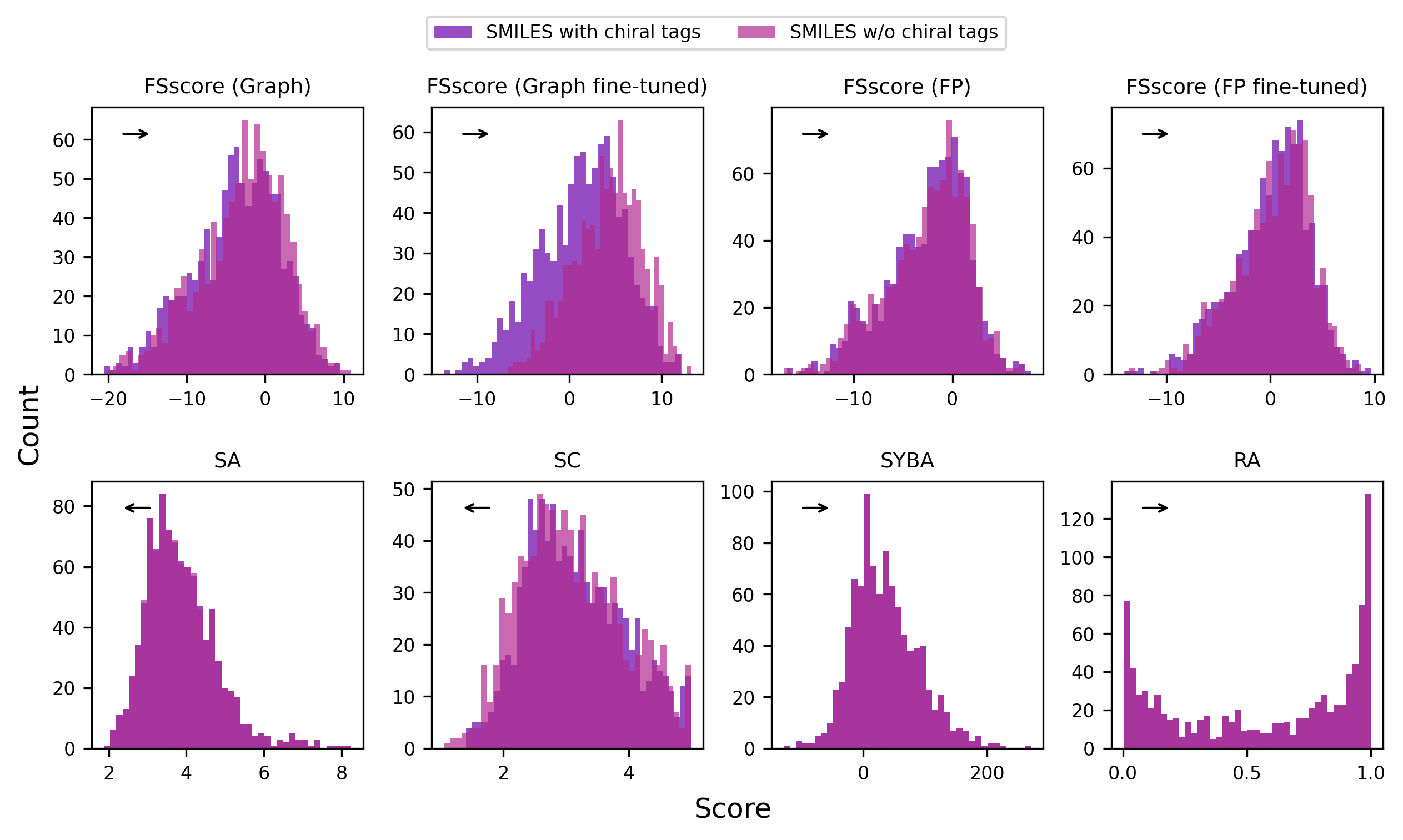}
		\caption{Score distribution.}
		\label{subfig:chiral_dist}
	\end{subfigure} \hfill
	\begin{subfigure}[t]{0.39\textwidth}
		\centering
		\includegraphics[width=0.9\textwidth]{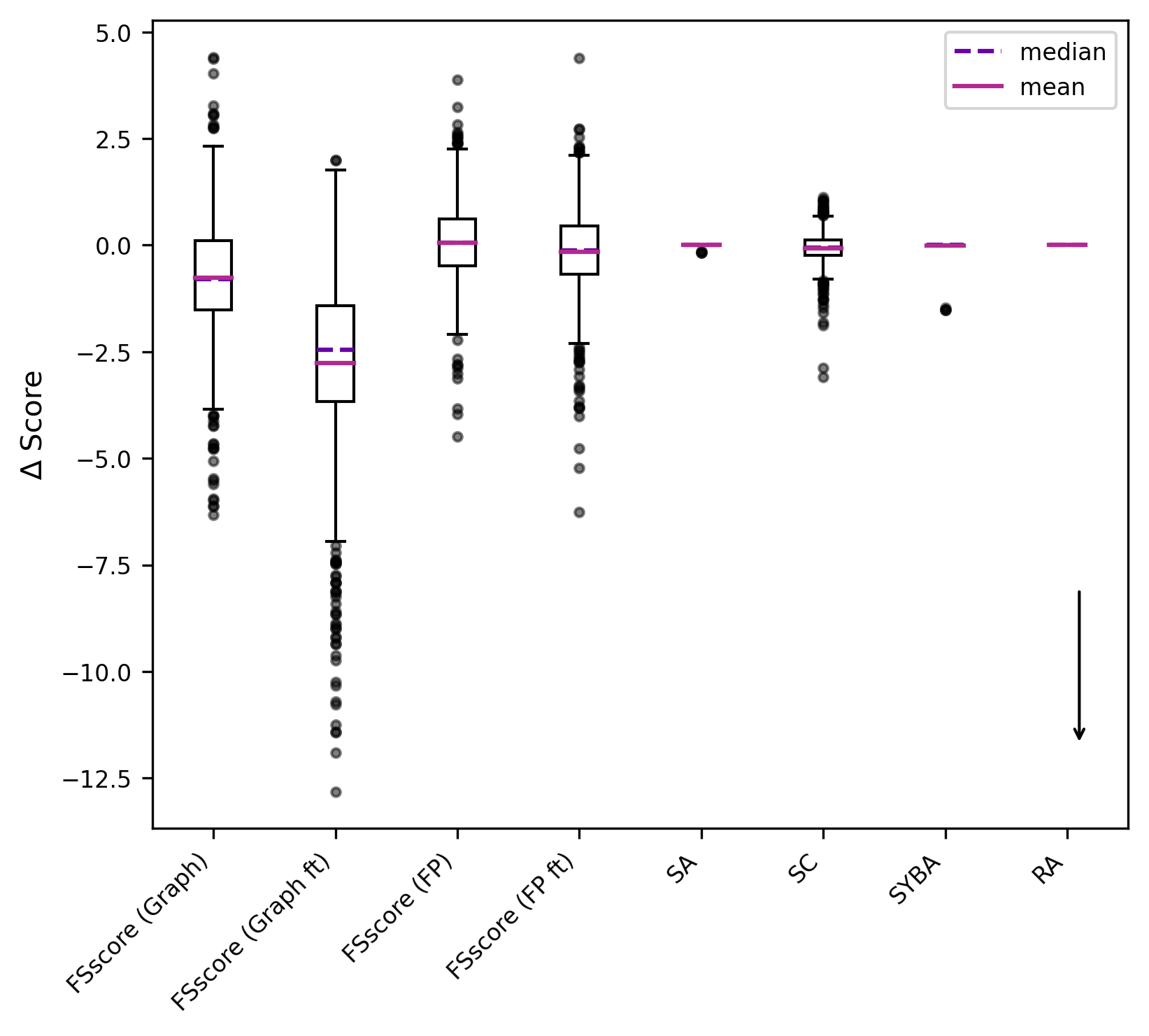}
		\caption{Score difference distribution.}
		\label{subfig:chiral_delta}
	\end{subfigure}
	\caption{Distributions showing the ability to differentiate molecules with assigned tetrahedral chirality from their unassigned counterpart. The desired prediction would score the assigned molecules as more complex resulting in negative delta values (assigned - unassigned) in Figure~\ref{subfig:chiral_delta}.}
	\label{fig:chirality}
\end{figure*}

Figure~\ref{fig:chirality} showcases the inability of all baseline scores, including our pre-trained models, to differentiate structures with assigned chirality from those without assignment. Synthesizing a predefined stereoisomer is a much more challenging task than being able to choose or even leave it up to chance. Figure~\ref{subfig:chiral_delta} shows that fine-tuning on the chirality test set allows the differentiation of molecules in terms of their chirality assignment resulting in predicting molecules with a given isomer as more difficult to synthesize. The performance could likely be improved by increasing the dataset size as opposed to the 50~pairs used here. While the fingerprint-based models (Morgan \emph{chiral} counts) are able to predict different scores, the meaning of the different representations is not captured by either the pre-trained or fine-tuned model.\par

\begin{figure}[!ht]
  \centering
  \includegraphics[width=8.6cm]{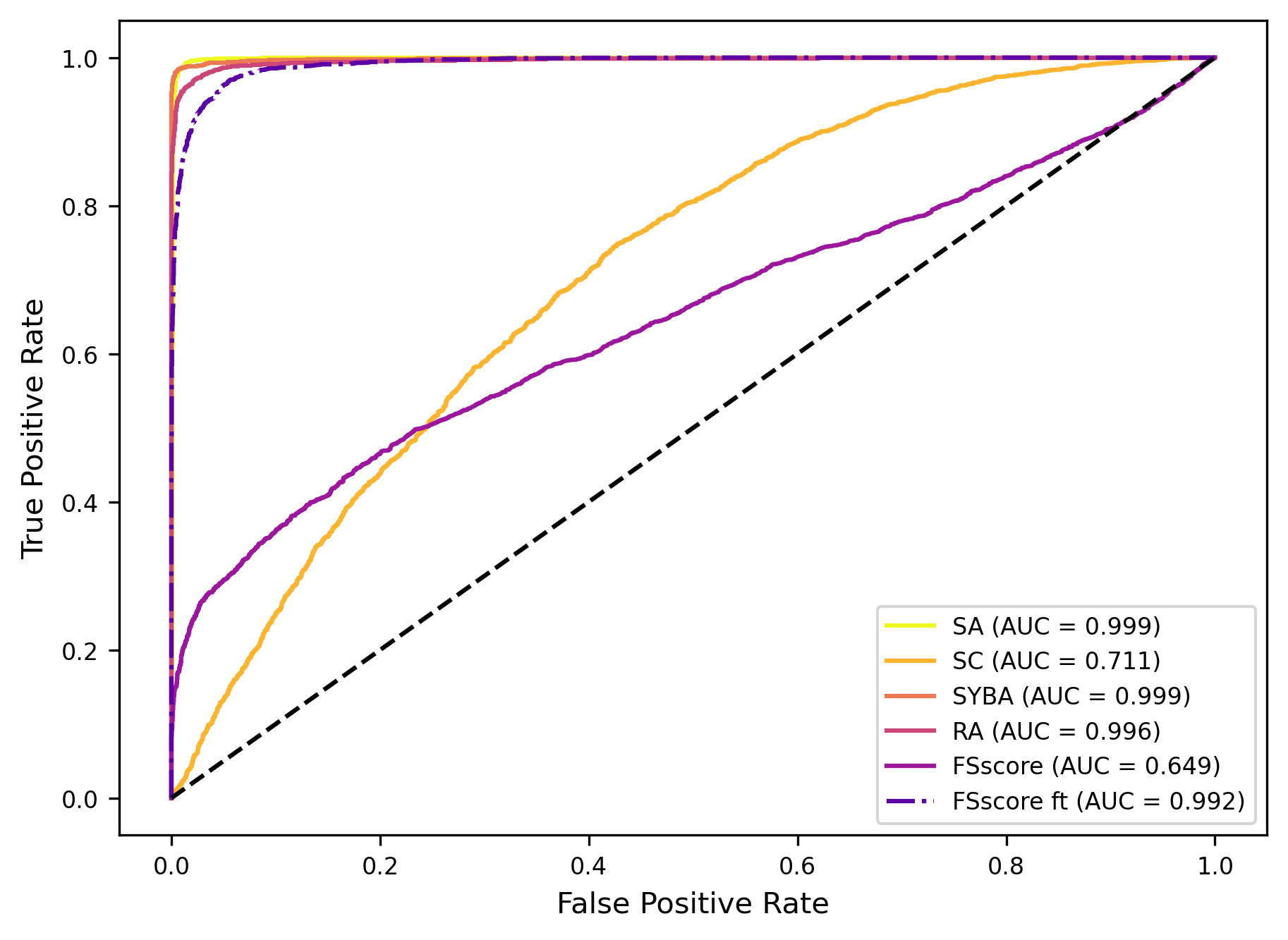}
  \caption{ROC curves showcasing the classification power of the various models to separate hard~(HS) from easy~(ES) to synthesize in the CP test set.}
  \label{fig:ROC_CP}
\end{figure}

Both SYBA test sets show room for improvement from the pre-trained baseline while the RAscore, SA score and SYBA perform well at distinguishing hard~(HS) from easy~(ES) to synthesize. The RAscore, being a classifier, has a structural advantage on this task. The same is true for SYBA, which was purposely trained to perform well on such datasets. The high \textit{AUC} with the SA score is to be expected on the CP set, where all HS molecules contain a ring-bridging atom. Thus, the CP dataset having clearly distinct homogeneous classes represents an ideal proof of concept fine-tuning task, which resulted in substantial increase in performance yielding an \textit{AUC} of 0.992 when fine-tuning on only 50~pairs. Good performance gain was even achieved with fewer data points as seen in Table~\ref{tab:SI_ft}. The MC test set is more diverse and the limited dataset size (40~pairs) makes evaluation challenging. The good performance of the SA score on the MC set is more surprising (see Fig.~\ref{fig:SI_MC_ROC}) and showcases how well-formulated rules are valuable on in-distribution datasets. However, the FSscore can also improve on such a heterogeneous dataset as shown in Table~\ref{tab:SI_ft}. When using overlapping pairs (molecules can appear in multiple pairs) the performance is improved substantially improved over the unique setting, however with the trade-off of reduced generalizability measured as performance on the pre-training test set (see Tab.~\ref{tab:SI_ft_rpt}). Furthermore, our goal is to keep the fine-tuning size to a minimum in order to facilitate human labeling. This set of experiments accentuates how the heterogeneity dictates the data demand but our results also indicate that it is still possible to gain performance on as little as 20~data points on the MC set.\par

The ability to score molecules, whose complexity was assigned by chemists, was assessed on the meanComplexity dataset by~\citet{sheridan_modeling_2014}. The correlations between all scores and the meanComplexity are shown in Figure~\ref{fig:SI_sheridan_pcc}. The correlation to meanComplexity was improved through fine-tuning (see Tab.~\ref{tab:SI_ft_sheridan}) but still lacks behind scores such as SA~score or RAscore. This could likely be rescued by increasing the fine-tuning dataset size from 50~pairs, which we aimed at keeping small. Only by using 500~overlapping pairs (molecules can appear in multiple pairs) can the fine-tuned FSscore outperform the SA score in terms of PCC (0.84 \latin{vs.} 0.8 -- see Figure~\ref{fig:SI_sheridan_barplot}).\par

\subsubsection{Beyond Rule-of-Five drugs: PROTACs}
\begin{figure}[!ht]
  \centering
  \includegraphics[width=6cm]{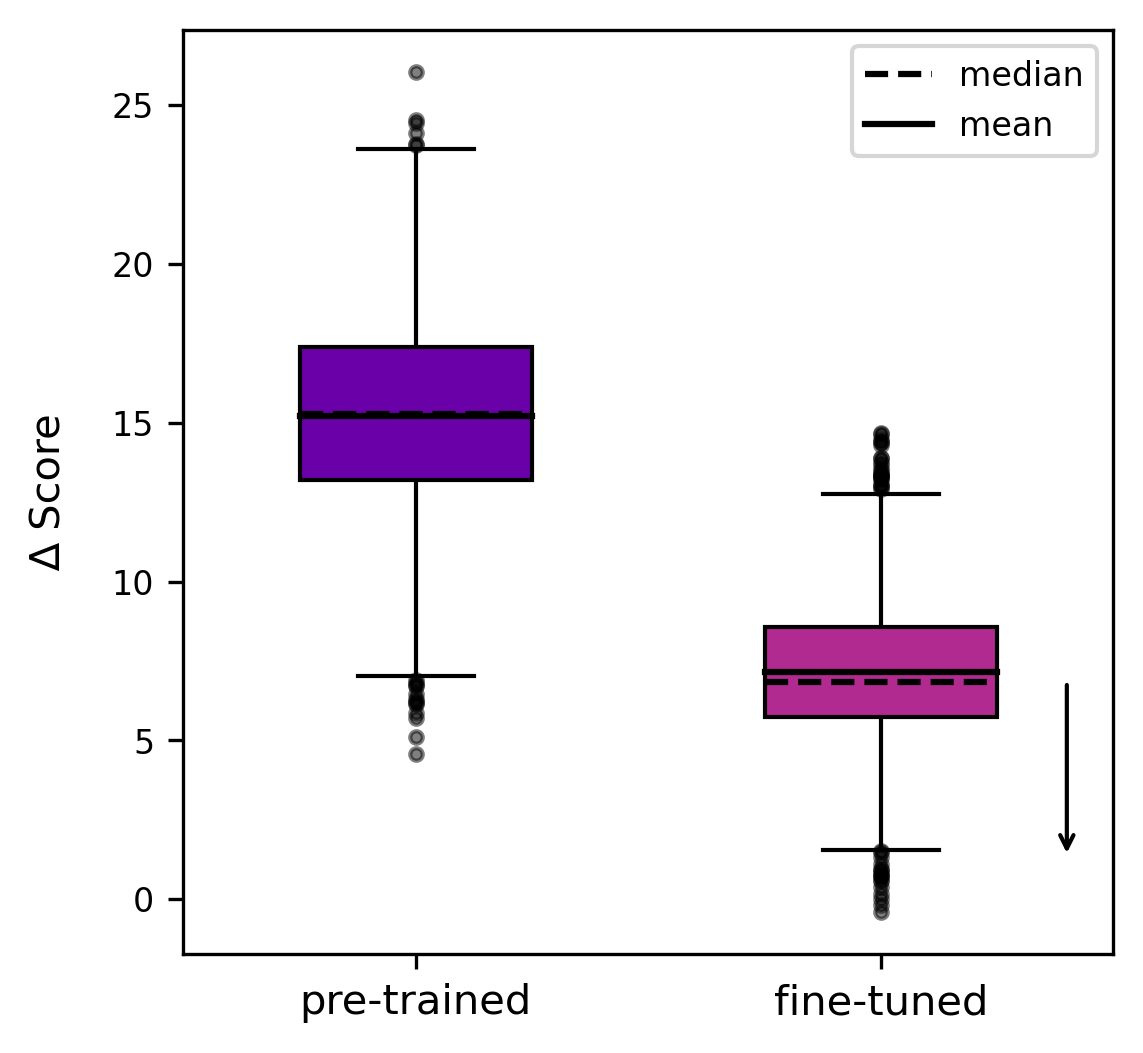}
  \caption{FScore difference between full PROTAC and most complex respective fragment (either of the two ligands or linker).}
  \label{fig:protac_barplot}
\end{figure}

\begin{figure*}[!ht]
    \centering
    \includegraphics[width=17.4cm]{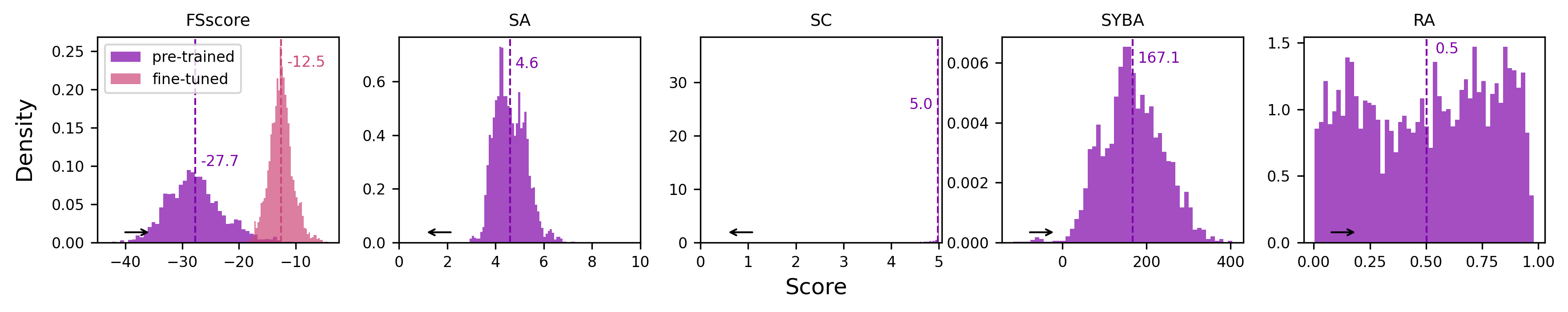}
    \caption{Score distributions obtained for the PROTAC-DB sample. The first plot shows the scores before (grey) and after (purple) fine-tuning on 80~pairs of human-labeled PROTACs. The arrows each indicate direction of higher synthetic feasibility and the dashed line marks the mean.}
    \label{fig:protacdb_hist}
\end{figure*}

PROTACs are large molecules (700-1100~Da) compared to traditional drugs but their fragmented composition allows the synthesis of each ligand and the linker separately before connecting the three~parts. Thus, their size often misleads known scores, ranking them as synthetically hard, generating unreliable scores. Feedback from an expert chemist in PROTACs, remarked that most molecules are relatively easy to synthesize. Figure~\ref{fig:protacdb_hist} shows a distinct shift towards higher predicted synthetic feasibility after fine-tuning the FSscore and Figure~\ref{fig:protac_barplot} highlights the narrowing gap between each PROTAC and the respective most complex fragment (ligands or linker) after fine-tuning. The latter observation is desirable under the assumption that connecting the three fragments is not challenging, the complexity of the full PROTAC is not much higher than its most complex components. However, the increase in performance is small (50~pairs: \textit{Acc} increases from 0.53 to 0.57 and \textit{AUC} from 0.43 to 0.52) on all tested fine-tuning dataset sizes and evaluation is challenging with only 100~labels (see Tab.~\ref{tab:SI_ft}). The performance gain on all PROTACs and the learning curves (see Fig.~\ref{fig:SI_lc_protacdb}) demonstrate the ability to learn relevant features, indicating that with additional labeled data. the performance could be further increased.\par 

\subsubsection{Targeted Improvement of Generated Structures}

\begin{figure*}[t]
    \centering
    \includegraphics[width=1\textwidth]{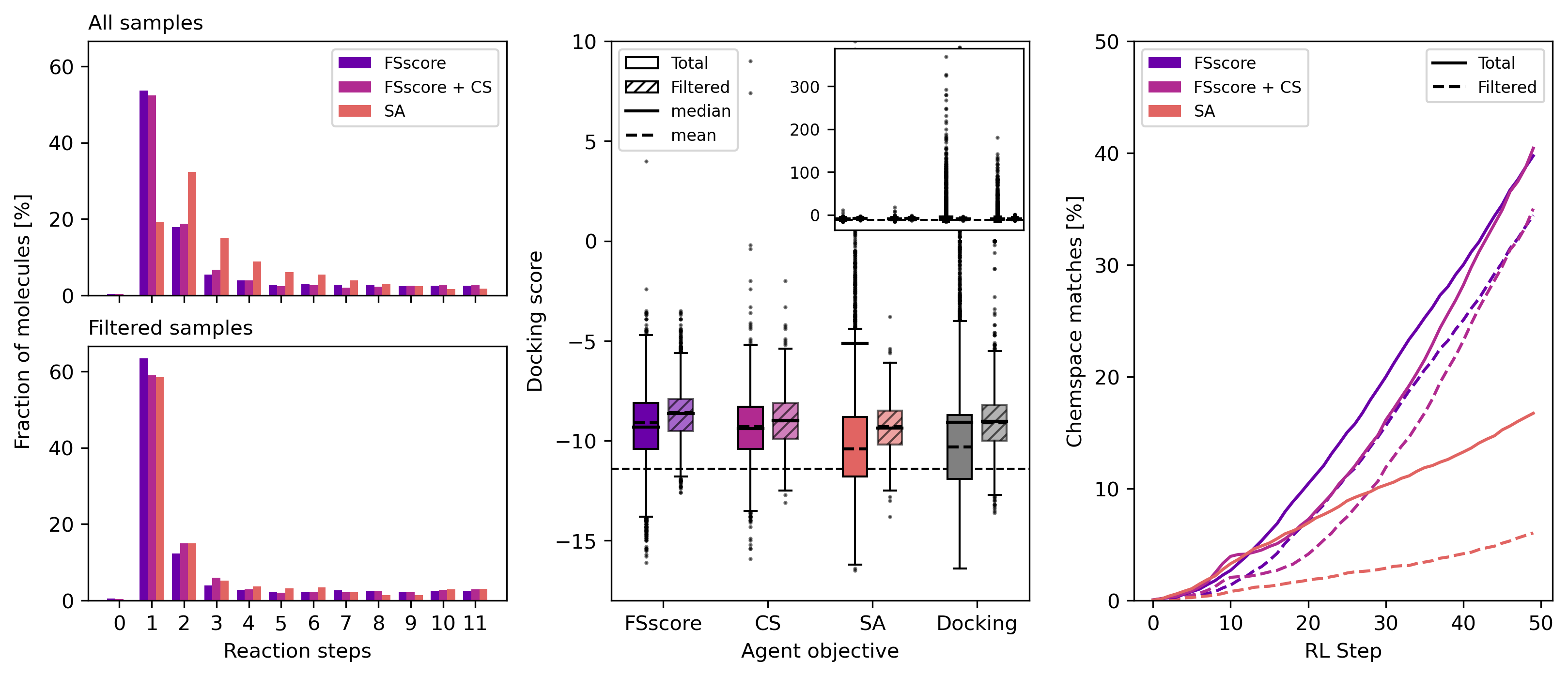}
    \caption{Analysis of the generated molecules optimized for the FSscore (human labeling: \textit{FSscore}; Chemspace labeling: \textit{FSscore + CS}) or the SA score. \textit{Left:} Fraction of reaction steps predicted by AiZynthFinder~\cite{genheden_aizynthfinder_2020} for each generated molecule before (\textit{upper}) and after (\textit{lower}) filtering. Zero reaction steps indicate that the compound was found in commercial providers. \textit{Middle:} Box plots displaying the distribution of docking scores for the four trained agents. The dashed line indicates the docking score of the true ligand Risperidone. \textit{Right:} Cumulative fraction of generated molecules over the optimization steps that have a match in the Chemspace database~\cite{chemspace}. (CS~=~Chemspace)}
    \label{fig:reinvent_eval}
\end{figure*}

\begin{figure*}[!ht]
    \centering
    \includegraphics[width=17.4cm]{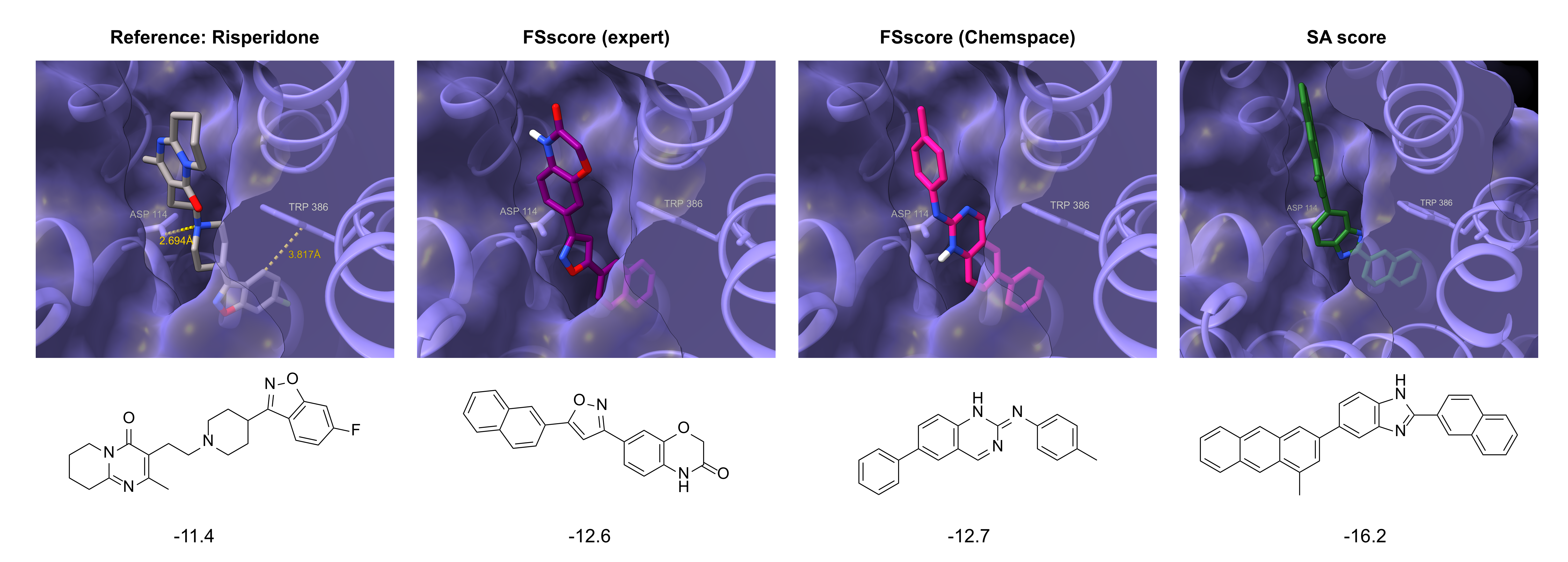}
    \caption{Examples of docked generated molecules with docking scores in the top 10. The first panel shows the reference crystal structure of DRD2 with the ligand Risperidone (PDB~ID:~6CM4). \textit{Upper:} Docked ligands. \textit{Middle:} 2D depiction. \textit{Lower:} Docking score.}
    \label{fig:reinvent_docked_examples}
\end{figure*}

Our last case study couples human feedback to a generative model attempting targeted improvement of synthetic feasibility of the generated molecules. The optimization of the first agent for docking to the D2 Dopamine receptor~(DRD2, PDB ID: 6CM4), which is a target for antipsychotic drugs, leads to the generation of lipophilic structures and polyaromatic ring systems. To improve the synthesizability two other agents are optimized using the FSscore as a reward function: the first approach fine-tunes the FSscore once with the docking-agent output and optimizes the generator agent over 50~steps while the second approach iteratively fine-tunes the FSscore at every step with the newly generated SMILES that are paired with the most similar molecules in Chemspace. Figure~\ref{fig:reinvent_eval} shows the superiority of the FSscore over the SA~score in capturing the synthetic feasibility and generating molecules that are predicted to require fewer reaction steps. This comes at the expense in terms of docking score compared to the SA-optimized agent, which in turn generated many outliers with high docking scores. Interestingly, when applying the common filters using PAINS alerts~\cite{pains} and Lipinski rules (molecular weight < 500~g/mol, less than 5~hydrogen bond donors and less than 10~hydrogen bond acceptors, logP between -5 and 5) only 13\% of the SA-optimized molecules pass while 69\% and 73\% pass for the expert-tuned and Chemspace-tuned FSscore, respectively. The baseline with SA even reduced the fraction passing those filters from 30\% after optimizing for docking indicating that SA is not suited to this task despite countless applications in the literature.~\cite{de_cao_molgan_2022, hjensen_graph-based_2019, cofala_evolutionary_2020, nigam_tartarus_2022, wang_chemistga_2022} This further indicates that the FSscore seemingly captures additional features relevant to this application not solely connected to synthesizability. Most convincingly, however, is the increase in exact matches found in Chemspace when using either approach optimizing for the FSscore compared to when using the SA score (40\% compared to 17\% and 35\% compared to 6\% after filtering). The reported fraction of matches reports the number of molecules that are certain to be synthesizable and is further only the lower bound of that. Figure~\ref{fig:SI_strucs_reinvent} shows examples of generated structures of all three agents. The random subsets of structures highlight that the SA~score is not suited to this task, resulting in a high frequency of undesirable motifs, such as chains of phenyl groups connected by single bonds. Selected examples from the top ten generated structures according to docking scores, are displayed in Figure~\ref{fig:reinvent_docked_examples} alongside the reference crystal structure. All structures with the best docking scores have an aromatic ring in the same position as the reference allowing a \(\pi\pi\)-interaction and a few of them potentially enabling the hydrogen bond to Asp114. This case study highlights how the FSscore can be adapted to a more challenging task allowing the \textit{de novo} generation of molecules with improved synthetic feasibility while being constrained through the primary objective of docking. The comparison to the SA score emphasizes the advantage of tailoring our score to this specific data distribution.

\section*{Conclusion}
\label{sec:conclusion}
We have introduced the Focused Synthesizability score (FSscore), a novel machine learning approach to evaluate the synthetic feasibility of molecules. The FSscore represents a powerful tool for practicing chemists working in areas like drug discovery, organic synthesis, and molecular design. By leveraging pairwise preference learning on a large dataset of chemical reactions, the FSscore establishes a robust baseline for assessing synthesizability. However, the true strength of our approach lies in the ability to fine-tune the model using human feedback, allowing chemists to personalize and focus the FSscore towards specific chemical spaces or synthesis objectives of interest.
Through several case studies, we have demonstrated the FSscore's efficacy in learning from relatively small amounts of labeled data provided by expert chemists. Fine-tuning improved performance across diverse chemical domains, such as chiral molecules, PROTACs, and drug-like compounds, highlighting the method's versatility and adaptability. 

Importantly, the FSscore represents a practical and user-friendly tool. Using our provided application, chemists can readily score molecules, provide preference labels, fine-tune the model to their needs, and deploy a personalized synthesizability model--all without requiring specialized coding expertise. Furthermore, as a fully differentiable approach, the FSscore could be seamlessly integrated into generative models, reinforcement learning frameworks, or serve as a filter for virtual screening studies.

Our experiments optimizing a molecular generator with the FSscore demonstrated a substantial increase in the fraction of produced molecules that are known to be synthesizable according to commercial catalogs, compared to using traditional metrics. This emphasizes the FSscore's potential to guide computational molecular design towards synthetically accessible regions.

In summary, the FSscore provides chemists with a flexible, data-driven tool to assess and prioritize molecules based on synthetic feasibility considerations, bridging the domains of computational and experimental chemistry. We envision this approach will accelerate molecular discovery efforts by effectively leveraging both machine learning capabilities and human chemical intuition.

\begin{acknowledgement}
RMN thanks VantAI (USA) for their support. PS acknowledges support from the NCCR Catalysis (grant number 180544), a National Centre of Competence in Research funded by the Swiss National Science Foundation. We also thank Chalada Suebsuwong and Zlatko Jončev for their expert chemistry feedback and Théo Neukomm for his help with the application. We thank Yurii Moroz and Chemspace for the API access to their molecule search.
\end{acknowledgement}

\bibliography{ref}

\makeatletter\@input{xx.tex}\makeatother

\end{document}


\maketitle

\renewcommand\thesection{S\arabic{section}}
\renewcommand\thesubsection{\thesection.\arabic{subsection}}
\renewcommand{\thefigure}{S\arabic{figure}}
\renewcommand{\thetable}{S\arabic{table}}
\renewcommand{\theequation}{S\arabic{equation}}

\tableofcontents

\section{Code and data availability}
\label{sec:code}
All code used for training, fine-tuning, and scoring the FSscore is available at \url{https://github.com/schwallergroup/fsscore}. This repository also includes an application that can be run locally and allows a more intuitive and accessible way to label data, fine-tune, and deploy a model.
\subsection{Application}
\label{sec:app}
\begin{figure}
    \centering
    \includegraphics[width=0.9\textwidth]{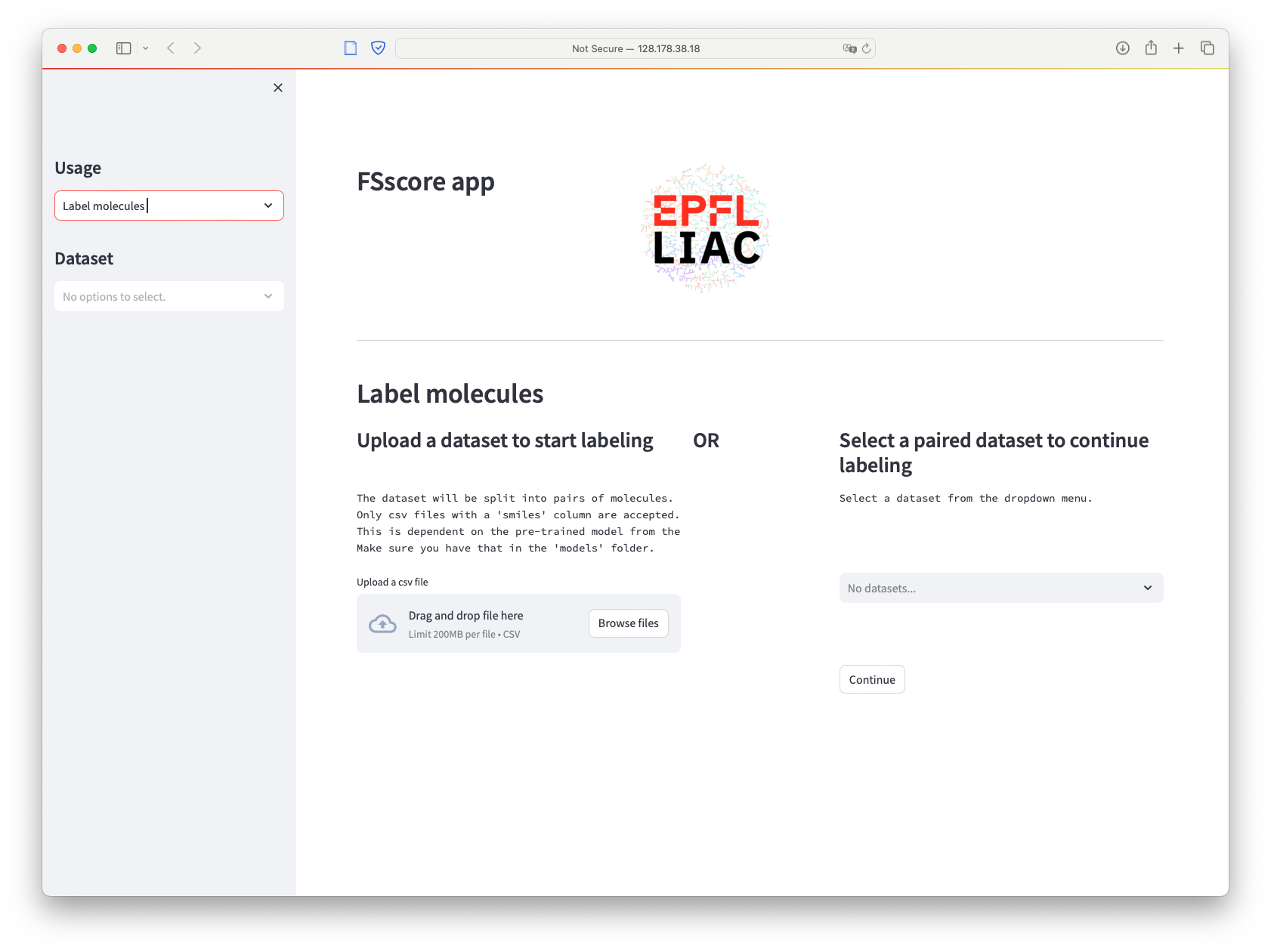}
    \caption{Selection window to either pair molecules of a new dataset or continue with an already paired dataset.}
    \label{fig:app_pair}
\end{figure}

\begin{figure}
    \centering
    \includegraphics[width=0.9\textwidth]{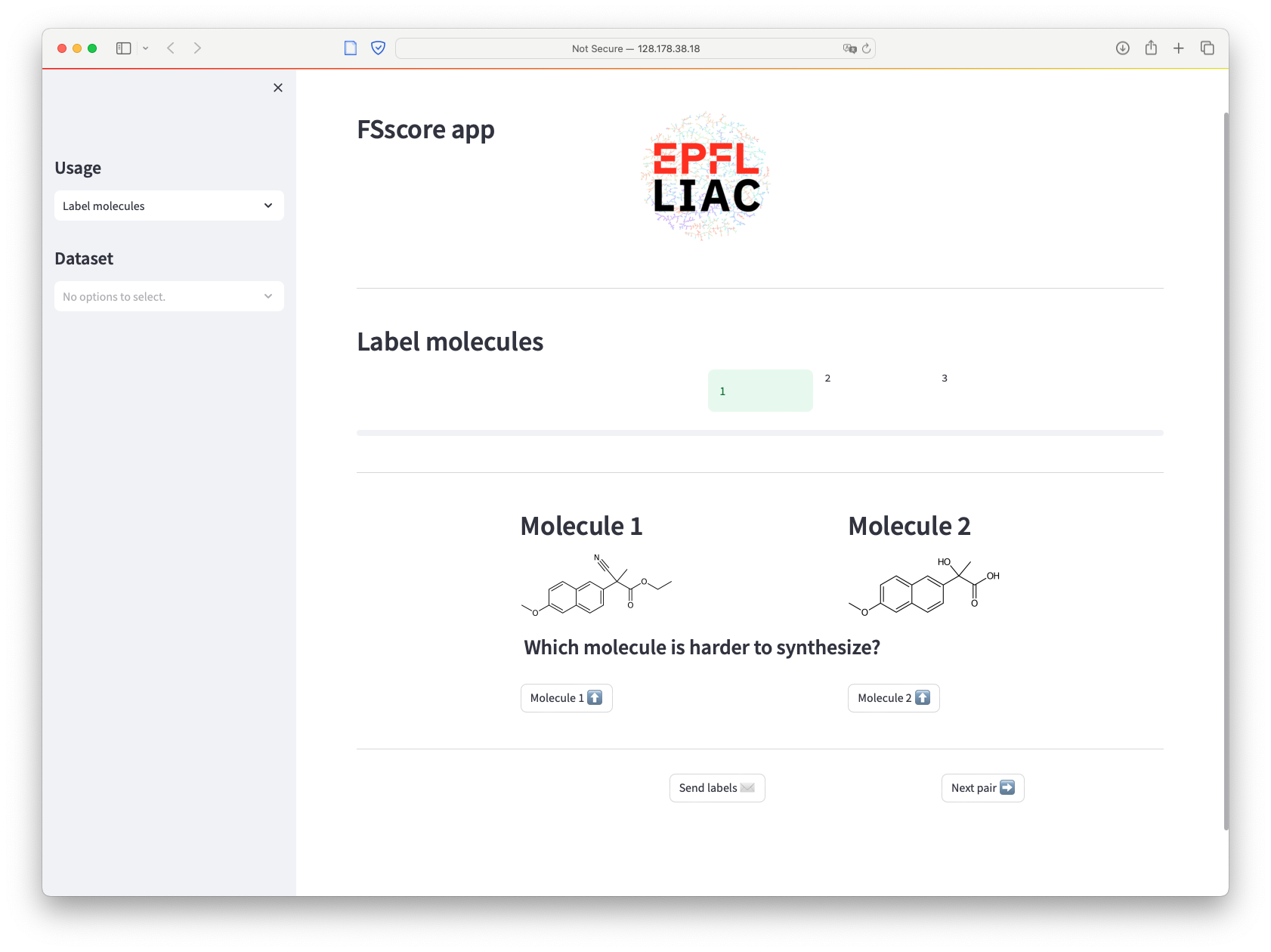}
    \caption{Screenshot of labeling interface. The user is prompted to select the molecule that is harder to synthesize using the buttons below before proceeding to the next pair. Once finished, the user can send the labels with the respective button.}
    \label{fig:app-label}
\end{figure}

\begin{figure}
    \centering
    \includegraphics[width=0.9\textwidth]{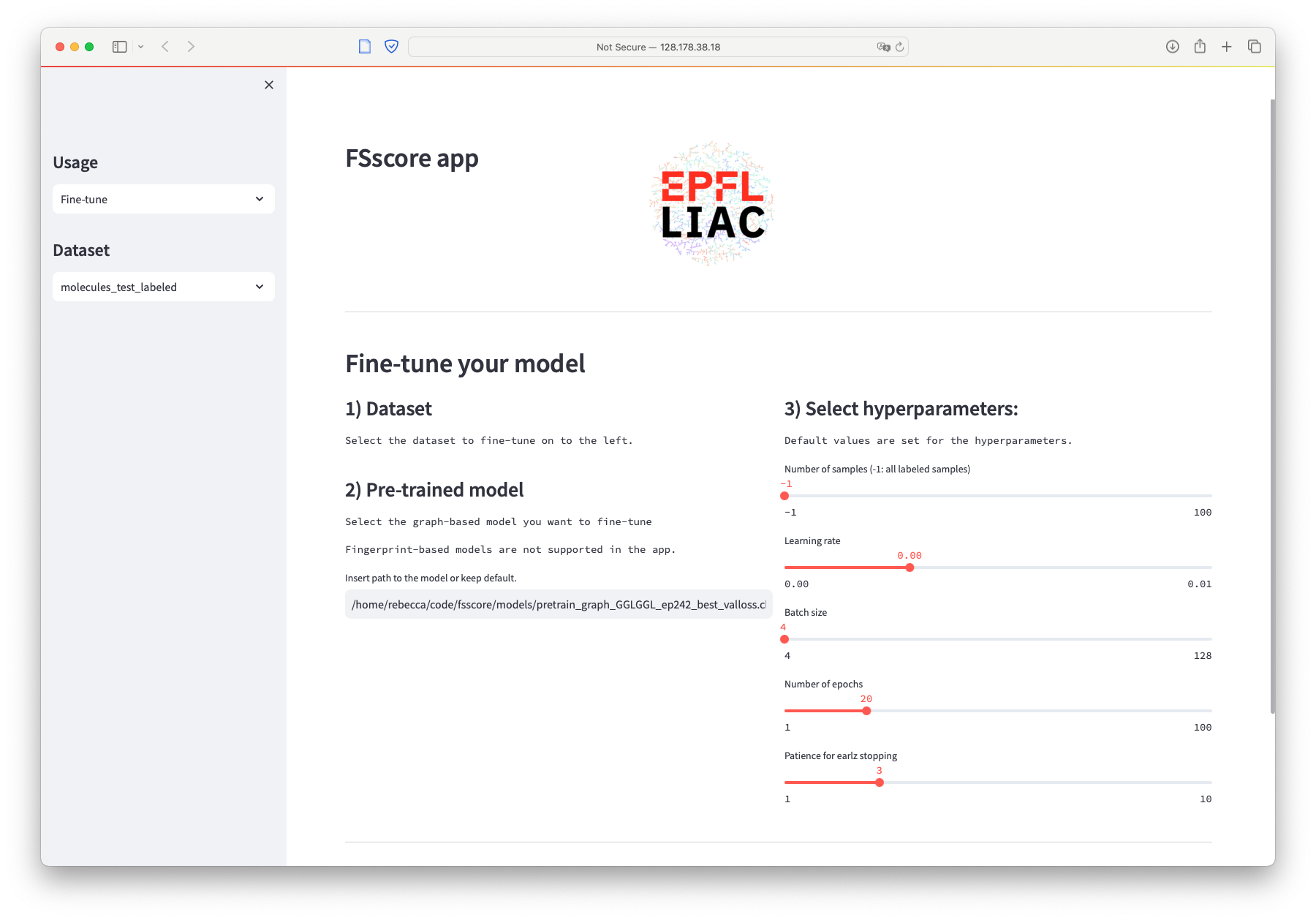}
    \caption{Screenshot of window initializing fine-tuning. The dataset to fine-tune with is selected in the drop-down menu left and the path to the pre-trained model can be changed under point 2. The rulers under point 3 allow the user to tune the hyperparameters.}
    \label{fig:app-ft}
\end{figure}

\begin{figure}
    \centering
    \includegraphics[width=0.9\textwidth]{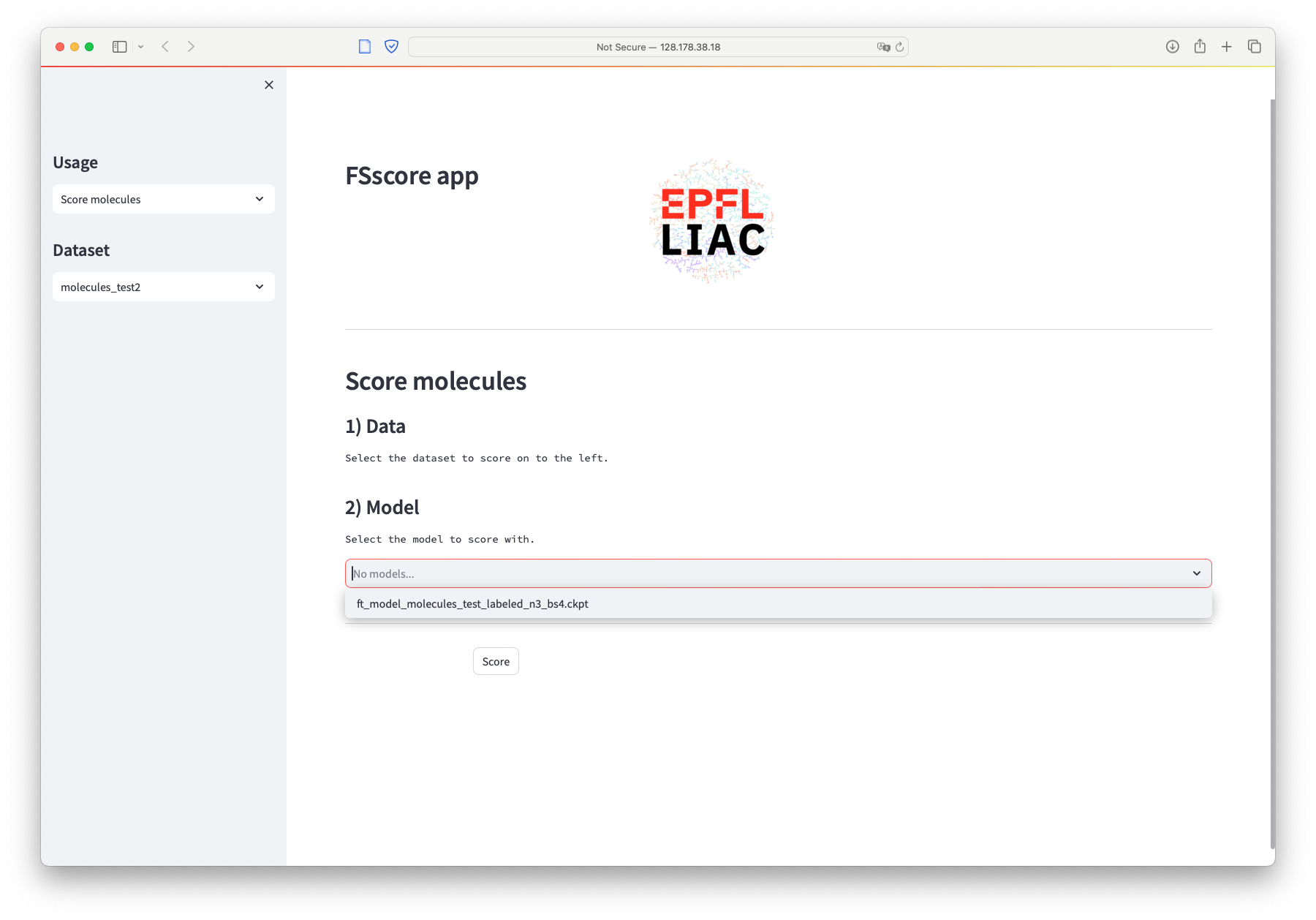}
    \caption{Screenshot of settings to score molecules given a dataset and a fine-tuned model.}
    \label{fig:app-score}
\end{figure}

The application allows a user to perform all steps necessary to get a personalized model as intended. These steps involve: pairing molecules, labeling pairs, fine-tuning the pre-trained model, and ultimately scoring molecules. The application is built using Streamlit~\cite{streamlit} but is deployed locally. This means that the speed is limited by the computer on which the user is running the app. All models were trained on one GPU~(NVIDIA~GeForce~RTX~3090) but the application will run also on a computer without GPUs at the cost of a much increased running time. Furthermore, running it locally also requires the user to install the package, which we recommend doing using a \texttt{conda} environment and \texttt{pip}. All the necessary steps and usability are outlined below.\\
\textbf{1. Installation}\\
We recommend running the app with the packages stored in a \texttt{conda} environment\footnote{Conda installation guide: \url{https://conda.io/projects/conda/en/latest/user-guide/install/index.html}}. First, clone the GitHub repository from the link above (alternatively download the zip file):

\begin{minted}
[
framesep=2mm,
baselinestretch=1.2,
fontsize=\footnotesize,
]
{bash}
git clone https://github.com/schwallergroup/fsscore.git
\end{minted}
Next, change the directory to the repository and install all required packages, like so:
\begin{minted}
[
framesep=2mm,
baselinestretch=1.2,
fontsize=\footnotesize,
]
{bash}
cd fsscore
conda create -n fsscore python=3.10
conda activate FSscore
pip install -r requirements.txt
pip install -e .
\end{minted}
\textbf{2. Start the application}\\
Starting the application only requires one line of code (from the root directory of the repository), which will lead to a window to pop up or return an active link:
\begin{minted}
[
framesep=2mm,
baselinestretch=1.2,
fontsize=\footnotesize,
]
{bash}
streamlit run streamlit_app/run.py
\end{minted}
From the landing page, select \emph{Label molecules} in the upper drop-down menu on the left. This will lead to selecting the next steps. If a dataset with labels is available, go to step 4., otherwise continue with the next step. To directly score molecules and a fine-tuned model is available one can jump to step 6.\\
\textbf{3. Pair molecules}\\
In this step, a list of molecules provided as SMILES first are paired, and then ranked based on uncertainty of the pre-trained model. This second step will take some time if not performed on a machine with GPU-support. Select the option on the left and provide the path to the \texttt{csv} file (Figure~\ref{fig:app_pair}). This file should have \texttt{smiles} as a column header for the SMILES structures to be paired.
\\
\textbf{4. Label molecule pairs}\\
If a dataset with paired molecules is available, place it in \texttt{streamlit\_app/data/unlabeled}. The dataset should appear in the drop-down menu on the right. Make sure to provide the paired molecules with column headers \texttt{smiles\_i} and \texttt{smiles\_j}.\\
The labeling window~(see~Figure~\ref{fig:app-label}) prompts the user to select the molecule that is harder to synthesize, before going to the next pair. Once the user is finished labeling, the button \emph{Send labels} can be pressed. The labeling process can also be continued later and the labeled data can be downloaded. \textbf{Note:} The more labeled pairs, the better the performance of the fine-tuned model.
\\
\textbf{5. Fine-tune}\\
Once data is labeled, one can proceed to fine-tuning. This is by default done using the best pre-trained model from this work. The user can also place their own checkpoint at the same location (folder \texttt{models}). How to pre-train a new model is described on our GitHub repository. Before initializing fine-tuning, one can tune several hyperparameters. If nothing is adjusted, the (recommended) default parameters as used in this work are selected (Figure~\ref{fig:app-ft}). Fine-tuning in general is very fast, but depending on your system this can take a few minutes. Proceeding to score molecules can be achieved by selecting the respective action on the upper drop-down menu. This process can cause some issues in which case we would recommend returning to the home page before trying again.\\
\\
\textbf{6. Score molecules}\\
Lastly, one can deploy the previously fine-tuned model and score molecules. For this, the user may place a \texttt{csv} file in \texttt{streamlit\_app/data/scoring}, which contains the column header \texttt{smiles}. Then, the user can select the correct dataset in the drop-down menu on the left as well as the desired fine-tuned model (Figure~\ref{fig:app-score}).

\section{Machine Learning Methods}
\label{sec:FS_meth}

The novel Focused Synthesizability score~(FSscore) is trained in two stages. First, we train a baseline score assessing synthesizability using a graph representation and suitable message passing scheme improving expressivity over similar frameworks such as the SCScore. Secondly, we introduce our fine-tuning approach using human expert knowledge, allowing us to focus the score towards a specific chemical space of interest.

\subsection{Architecture and representation}
\label{sec:architecture}
Our approach to learning a continuous score to assess the synthesizability is inspired by \citet{choung_learning_2023}, which framed a similar task as a ranking problem using binary preferences. Specifically, every data point consists of two molecules for each of which we predict a scalar in separate forward passes. The minimization of the binary cross entropy between the true preference and the learned score difference~\(\delta_{ij}:=\hat{f}(m_i)-\hat{f}(m_j)\) (scaled using a sigmoid function) constitutes our training objective.\par 

The function \(f:\mathcal{M}\rightarrow\mathbb{R}\) learns to parametrize molecules as an expressive latent representation given a set of molecules~\(m_1,..., m_n\in\mathcal{M}\). We represent molecules as graphs~\(G=~(\mathcal{V}, \mathcal{E})\) with atoms as nodes~\(x_1,..., x_n\in\mathcal{V}\) and bonds as edges~\(e_1,..., e_m\in\mathcal{E}\) from which we can compute the line graph~\(L(G)\) offline iteratively to the desired depth. The transformation process to the line graph is defined such that the edges~\(\{e_1,...,e_m\}\) of graph~\(G\) are the nodes of the line graph and these nodes are connected if the corresponding edges of~\(G\) share a node. The graph neural network (GNN) embedding the molecular graph consists of the graph attention network~(GATv2)~\cite{brody2022how} operating on the graph~\(G\) and the Line Evolution~(LineEvo)~\cite{ren_enhancing_2023} layer operating on the line graph~\(L(G)\) as message passing schemes. Both GATv2 and LineEvo use an attention mechanism to update the node representations~\(\{h_1,...,h_n\}\) as follows:
\begin{equation}
    e(h_i,h_j)=a^T\sigma(W_ih_i||W_jh_j)
\end{equation}

where \(a\in\mathbb{R}^{2d'}\) and \(W\in\mathbb{R}^{d'\times d}\) are learned, \(\sigma\) denotes an activation function and \(||\) denotes concatenation. The GNN operates on a hidden size of~128 and the input dimensions depend on the initial featurization of the edges and nodes (see Tab.~\ref{tab:SI_features}. GATv2 layers use 8~heads and \texttt{LeakyReLU} as activation function after updating the hidden feature vector while LineEvo layers use \texttt{ELU}. In GATv2, these local attention scores~\(e_{ij}\) are then averaged across all neighbors~\(\mathcal{N}\) (see Eq.~\ref{eq:norm_attn}) to obtain a normalized attention coefficient~\(\alpha_{ij}\) to compute the updated node representation~\(h_i'\) as a weighted average (see Eq.~\ref{eq:update_h}):
\begin{equation}
\label{eq:norm_attn}
    \alpha_{ij}=\text{softmax}_j(e(h_i,h_j))=\frac{\text{exp}(e(h_i,h_j))}{\sum_{j\in\mathcal{N}_i}\text{exp}(e(h_i,h_j))}
\end{equation}
\begin{equation}
\label{eq:update_h}
    h_i'=\sigma (\sum_{j\in\mathcal{N}_i}\alpha _{ij}Wh_j)
\end{equation}
with \texttt{PReLU}~\cite{prelu} as nonlinearity~\(\sigma\). In LineEvo layers, \(e(h_i,h_j)\) is simply transformed using \texttt{ELU} to obtain the new node representation~\(h_i'\).\par 
These transformation layers are stacked so that two~GATv2~(G) layers are followed by one~LineEvo~(L) layer~(GGLGGL). Each of these layers is followed by a readout function, which consists of a global max pooling layer and global weighted-add pooling as suggested by~\citet{ren_enhancing_2023} obtaining a molecular representation through concatenation of the two readouts. The intermediate molecular representations of all layers are summed, and the final score~\(s_i=\hat{f}(m_i)\) is inferred with a multilayer perceptron~(MLP) (3 hidden layers of size 256, \texttt{ReLU} as activation function). The model described above was compared to five~other implementations: GATv2 layers only~(GGG) and four~fingerprint implementations, namely Morgan (boolean), Morgan counts, Morgan chiral (boolean), and Morgan chiral counts, all with radius~4 and 2048~bits.~\cite{rogers2010extended} The fingerprints are embedded by one~linear layer (hidden size of 256) with \texttt{ReLU} as activation function followed by the aforementioned MLP. More detailed information can be found in Section~\ref{sec:SI_training_details}.

\subsection{Pairing molecules for fine-tuning}
\label{sec:meth_ft}
To focus our scoring model on a desired chemical space, we apply a fine-tuning approach inspired by~\citet{choung_learning_2023}. This approach was optimized to require few data points in order to limit the time required from expert chemists labeling the training examples in case of labels based on human feedback. Datasets to be used for fine-tuning can be of various origins such as a chemical scope suggested by experimentalists or specific chemical spaces encompassing e.g. natural products. Training our model requires pairs of molecules with a binary preference as label. In order to do so, we first cluster a dataset using the \textit{k}-means algorithm based on the Tanimoto distances of Morgan counts fingerprint. The number of clusters \textit{k} was determined for every dataset individually where the mean Silhouette coefficient over a range of \textit{k} was the smallest.~\cite{ROUSSEEUW198753}. The range queried was set from 5 to 29 for datasets larger than 500~pairs and to 3 to 9 for smaller datasets. Secondly, molecules are paired in such a way that they come from different clusters, appear only once in the paired dataset and have opposite labels if available. We also investigated the influence of having overlapping pairs (molecules can appear multiple times) for the SYBA (CP and MC) and the meanComplexity datasets. This approach results in more fine-tuning data but does not satisfy our desire to keep the dataset size small since human labeling could not make use of this advantage. Subsequently, the uncertainty of the prediction of the pre-trained model on those pairs is determined based on the variance of \(\delta_{ij}\) obtained using the Monte Carlo dropout method \cite{gal2016dropout} with a dropout rate of 0.2 on 100~predictions. The dataset is sorted with descending variance so that the pairs with high uncertainty are labeled first or used first for fine-tuning when selecting a subset. If no label is available the top \textit{n}~pairs (\textit{n} depends on the dataset size) are submitted to be evaluated by our expert chemist based on their preference with regard to synthesizability.

\subsection{Training details}
\label{sec:SI_training_details}

 To center the obtained score around zero a regularization factor of 1e-4 was applied to the predicted score like to obtain the regularization loss~\(\mathcal{L}_{reg}(\hat{f};\lambda):=\lambda ||\hat{f}||^2\) as suggested by~\citet{choung_learning_2023}. \(\mathcal{L}_{reg}\) is added to the cross-entropy loss term as described in Section~\ref{sec:architecture} to obtain the final loss~\(\mathcal{L}=\mathcal{L}_{CE}+\mathcal{L}_{reg}\). 
 
 \textbf{Pre-training:} The training set was randomly split in 25~equally sized subsets and the model was trained with each subset for 10~epochs totaling 250~epochs with a batch size of~128. This sequential learning approach was found to work well in increasing speed of training. The model was trained with the Adam optimizer~\cite{kingma2014adam} and an initial learning rate of~3e-4.

 \textbf{Fine-tuning:} We chose a transfer learning type of fine-tuning where all the weights of the pre-trained model can get adapted. All the hyperparameter except learning rate and batch size are kept the same. In order to avoid the forgetting the previously learned we track the performance metrics on a random subset of 5,000~pairs from the hold-out test set of the pre-training data. Of the four~tested initial learning rates (1e-5, 1e-4, 3e-4, 1e-3) we chose 1e-4 for the graph-based version and 3e-4 for the fingerprint-based models. These learning rates are a good trade-off between degradation of the previously learned and learning on the new dataset as can be seen in the learning curves in Appendix~\ref{sec:SI_ft_results}. We observed quick conversion with fewer than 20~epochs and in order to balance the aforementioned trade-off a custom early stopping method was applied. Training is stopped once either the validation loss (or training loss if the full dataset is used for production) has increased for 3 epochs (patience of 3) or the accuracy on the 5k subset of the hold-out test set decreased for 3 epochs with a delta threshold of 0.02. We trained for a maximum of 20~epochs. For evaluating the efficiency, we further varied the number of data points used for training of which we took 5~random pairs for validation. The results in Table~\ref{tab:SI_ft} show improvement in performance the bigger the dataset size. The batch size was set to~4.

\subsection{Initial featurization of graphs}
\label{sec:SI_inti_ft}
\begin{table}[H]
  \caption{Features for initializing the nodes and edges of the graph. One-hot encoding includes one additional bit for types not found in the choices (incl. in size). These properties were determined  using \texttt{RDKit}.~\cite{rdkit}}
  \label{tab:SI_features}
  \centering
  \begin{tabular}{llll}
    \toprule
    \textbf{Localization}     & \textbf{Feature}     & \textbf{Description} & \textbf{Size} \\
    \midrule
    \multirow{8}{*}{atom}& atom type     & one-hot encoded atom type \tablefootnote{Considered atom types: [H, B, C, N, O, F, Si, P, S, Cl, Br, I, Se]} & 14\\
    & charge & one-hot encoded formal charge [-4,4] & 10 \\
    & implicit hydrogens & one-hot encoded number of hydrogens [0,4] & 6 \\
    & degree & one-hote encoded degree [0,4] & 6 \\
    & ring information & 1 if in ring else 0 & 1 \\
    & aromaticity & 1 if aromatic else 0 & 1 \\
    & hybridization & one-hot encoded hybridization state \tablefootnote{Considered hybridization states: [UNSPECIFIED, s, sp, sp2, sp3, sp3d, sp3d2, other]} & 9 \\
    & chiral tag & one-hot encoded chiral tag [S, R, unassigned] & 4\\ \midrule
     & bond type & one-hot encoded bond type \tablefootnote{Considered bond types: [single, double, triple, aromatic]} &5\\
     bond & conjugated & 1 if conjugated else 0 & 1\\
     & ring information & 1 if in ring else 0 & 1 \\
    \bottomrule
  \end{tabular}
\end{table}

\section{Data}
\label{sec:SI_data}
\subsection{Pre-training data}
\label{sec:SI_data_pt}
To pre-train our model on a large collection of reactions, we combine the USPTO\_full~\cite{uspto_lowe} patent dataset with a complementary dataset~\cite{cjhif_data}. Reaction data implicitly contains information on the synthetic feasibility through the relation of reactant to product, with the product being synthetically more difficult. The USPTO\_full dataset was downloaded according to \url{https://github.com/coleygroup/Graph2SMILES/blob/main/scripts/download_raw_data.py} \cite{tu_permutation_2022} using all USPTO\_full subsets and "src" as product and "tgt" as reactants. The dataset was split so that every data point is a pair of one reactant and its respective product and the SMILES strings were canonicalized. The datasets do not contain reactions with multiple products. Data points with empty strings, identical reactants and products (isomeric SMILES), either reactant or product containing less than four heavy atoms or containing element types not in \{H, B, C, N, O, F, Si, P, S, Cl, Se, Br, I\} were removed. The datasets were further deduplicated retaining one instance of replicates. Furthermore, data points were removed ensuring that there are no cycles in the reaction network. This was achieved by removing back edges with the Depth-First Search~(DFS) algorithm as implemented by~\citet{sun2017breaking}. The filtered dataset consisted of 5,340,704~data points. The train test split was performed so that no molecules (note not just reactant:product pairs) are overlapping resulting in a significant data loss to prevent data leakage. The training set consists of 3,349,455~pairs and the hold out test set of 711,550~pairs (17.5\%).\par 

To qualitatively evaluate the pre-trained model the performance on the MOSES~\cite{polykovskiy_molecular_2020} and COCONUT~\cite{sorokina_coconut_2021} dataset were assessed. We downloaded the MOSES test set from ~\citet{polykovskiy_molecular_2020} as is totaling 176,074 SMILES. COCONUT was extracted from ~\citet{sorokina_coconut_2021}, all SMILES returning \texttt{None} with \texttt{RDKit}~\cite{rdkit} were removed and a random subset of 176,000 molecules was selected to match the MOSES fraction.

\subsection{Fine-tuning data}
\label{sec:SI_data_ft}
The data for the fine-tuning case studies come from various sources. The specific processing steps for every dataset is outlined below.\par 

\textbf{Chirality:} The obtain a test set for the chirality case study we filtered the deduplicated molecules from the pre-training training set keeping only those with \texttt{@} or \texttt{@@} tokens in the SMILES string, which define the chirality of a tetrahedral stereocenter. Of those, we sample 1,000 SMILES randomly and obtain their partner by getting the non-isomeric SMILES using \texttt{RDKit}~\citet{rdkit}.\par 

\textbf{SYBA sets:} The CP and MC test sets were extracted from \citet{vorsilak_syba_2020}. No further processing was required.\par 

\textbf{meanComplexity:} The meanComplexity dataset was downloaded from \citet{sheridan_modeling_2014} and cleaned by removing SMILES whose conversion to an \texttt{RDKit} molecule object failed. This processing step removed 44~SMILES yielding a fine-tuning test set of 1,731~molecules.\par 

\textbf{PROTACs:} For the PROTAC case study we extracted all PROTAC SMILES from the PROTAC-DB~\citet{weng_protac-db_2022} and deduplicated the dataset keeping the first instance resulting in 3,270 SMILES. After pairing and ranking by confidence as described in Appendix~\ref{sec:meth_ft}, 100~pairs were labeled by a medicinal chemist with expertise in PROTACs.
In order to compare the scores of the individual components to the full PROTAC as done in Figure~\ref{fig:protac_barplot} the individual fragments (the two ligands and the linker) have to be extracted. For this, we extracted the collections containing information on anchor and warhead (the two ligands) from PROTAC-DB and tried to match an anchor and warhead to every PROTAC by matching protein target ID and find the biggest substructure match. The linker was obtained by removing the ligands' atoms. For the 3270 PROTACs in the database we could find matching ligands for 1920 of them. This approach of extraction is necessary because PROTAC-DB does not cross-reference the PROTACs to the fragments and our extraction cannot guarantee concordance with the reported warhead and anchor in the respective publications but still is a good approximation.\par 

\section{\textit{De novo} design case study} 
\label{sec:reinvent}
To assess the applicability of the FSscore to a generative task we took advantage of the RL-framework of REINVENT~\cite{blaschke_reinvent_2020} and used the recently proposed augmented memory~\cite{guo_augmented_2023} optimization protocol. First, an agent was trained with the composite (equal weights) of the docking score to the D2 Dopamine receptor~(DRD2, PDB ID: 6CM4) and the molecular weight~(MW) as a reward. The docking score was determined using AutoDock Vina~\cite{koes_lessons_2013} on one conformer each. Conformer generation is performed with \texttt{RDKit}~\cite{rdkit} and the Universal force field~(UFF)~\cite{rappe_uff_1992} for energy minimization for a maximum of 600~iterations. The score \(x\) was scaled to obtain reward \(x'\) in a range of [0,1] by applying a sigmoid transformation as follows:
\begin{equation}
\label{eq:sigmoid}
    x'=\frac{1}{1+10^{10k\cdot\frac{x-\frac{a+b}{2}}{a-b}}}
\end{equation}
with \(a\) corresponding to a docking score \(x\) of -1, \(b\) -13 and the steepness \(k\) was set to 0.25 resulting in a reverse sigmoid (a small docking score results in a reward \(x'\) towards 1 while a high score returns a reward \(x'\) close to 0). The score for MW was formulated so that a high reward is returned when having a MW between 0 and 500~Da using a double sigmoid like so:
\begin{equation}
\end{equation}
\begin{equation}
    A=10^{c_{SE}\frac{x}{c_{div}}}
\end{equation}
\begin{equation}
    B=10^{c_{SE}\frac{x}{c_{div}}}+10^{c_{SE}\frac{b}{c_{div}}}
\end{equation}
\begin{equation}
    C=\frac{10^{c_{SI}\frac{x}{c_{div}}}}{10^{c_{SI}\frac{x} \\{c_{div}}}+10^{c_{SI}\frac{a}{c_{div}}}}
\end{equation}
\begin{equation}
    x'=\frac{A}{B}-C
\end{equation}
with coefficients \(c_{SE}=c_{SI}=500\) and \(c_{div}=250\), the lower bound \(a=0\) and upper bound \(b=500\). RL was carried out for 150~epochs, a batch size of~64, learning rate of~1e-4, a \(\sigma\) of~128 and augmented memory as optimization algorithm with 2~augmentation rounds and selective memory purge with a minimum similarity of~0.4 and a bin size of~10 based on the identical Murcko scaffold. For details on augmented memory in REINVENT we refer to \citet{guo_augmented_2023}. For our first approach (FSscore optimization) all valid SMILES (9,246) generated during this optimization process were subsequently used to fine-tune the FSscore as described above. The agent already optimized for docking was next used to optimize for synthetic feasibility either with the fine-tuned FSscore or the SA score as a comparison. Both approaches were trained in identical fashion with the following parameters: 50~epochs, batch size of~128, \(\sigma\) of~128, a learning rate of~1e-4 and using the augmented memory algorithm with 2~rounds of augmentation. The diversity filter (incl. selective memory purge) was based on identical Murcko scaffolds based on a minimal similarity of~0.4, a minimal score of~0.4 and a bucket size of~25. The FSscore was transformed to range from 0 to 1 with a sigmoid according to Equation~\ref{eq:sigmoid} with \(a=-13.08\), \(b=10.07\) and a steepness \(k=0.25\). The agent optimized for the SA score was transformed similarly reverting the sigmoid by setting \(a=10\), \(b=1\) and \(k=0.25\). Thus, a high FSscore result in a high reward, while a low SA score results in high reward and vice versa. As mentioned above, only molecules with a minimal score of~0.4 were collected and subsequently used for evaluation. The second optimization approach using FSscore and the Chemspace API (FSscore + CS) was carried out similarly but instead of fine-tuning the FSscore once, the model was updated at every step for 50~rounds of RL. For all the valid and unique SMILES of each batch (max 128~molecules) the most similar molecule in the Chemspace database of buyable building blocks and buyable/synthesizable screening compounds (databases: CSSB, CSSS, CSMB, CSMS) were extracted and used as less difficult partner in a pair for the FSscore fine-tuning. Any data point was removed if an exact match was found. all other hyperparameters were chosen as outlined above. This yielded 4,779 SMILES from the FSscore-trained agent, 3,519 for the FSscore+CS-trained agent, and 4,915 SMILES from the SA score-trained agent. The results are compared by the number of reaction steps predicted by AiZynthFinder and the fraction of exact matches found in the Chemspace database.~\cite{genheden_aizynthfinder_2020, chemspace}

\section{Metrics}
\label{sec:SI_metrics}
The Pearson correlation coefficient~(\textit{PCC}) between \(x\) and \(y\) is computed using the function \texttt{scipy.stats.pearsonr}~\cite{2020SciPy-NMeth} and is calculated as followed:
\begin{equation}
    PCC=\frac{\sum(x-m_x)(y-m_y)}{\sqrt{\sum(x-m_x)^2\sum(y-m_y)^2}}
\end{equation}
with \(m_x\) the mean of vector \(x\) and \(m_y\) the mean of vector y.\par 

The following metrics for datasets with binary labels (e.g. HS \latin{vs.} ES) are computed using the \texttt{sklearn.metrics} package~\cite{sklearn_api}. Accuracy~(\textit{Acc}), sensitivity~(\textit{SN}) and specificity are calculated based on the relative sizes of true positives~(\textit{TP}), true negatives~(\textit{TN}), false positives~(\textit{FP}) and false negatives~(\textit{FN}):
\begin{equation}
    Acc=\frac{TP+TN}{TP+TN+FP+FN}
\end{equation}
\begin{equation}
    SN=\frac{TP}{TP+FN}
\end{equation}

\begin{equation}
    SP=\frac{TN}{TN+FP}
\end{equation}

The area under the receiver operating characteristic~(ROC) curve~(\textit{AUC}) describes the ability to discriminate data points based on binary labels at various thresholds. It is computed by plotting \textit{SN} against \(1-SP\) at various cut-off values defining the two classes. 

\newpage
\section{Additional results}
\label{sec:SI_add_results}

 \begin{figure}[!ht]
	\centering
	\begin{subfigure}[t]{0.49\textwidth}
		\centering
		\includegraphics[width=\textwidth]{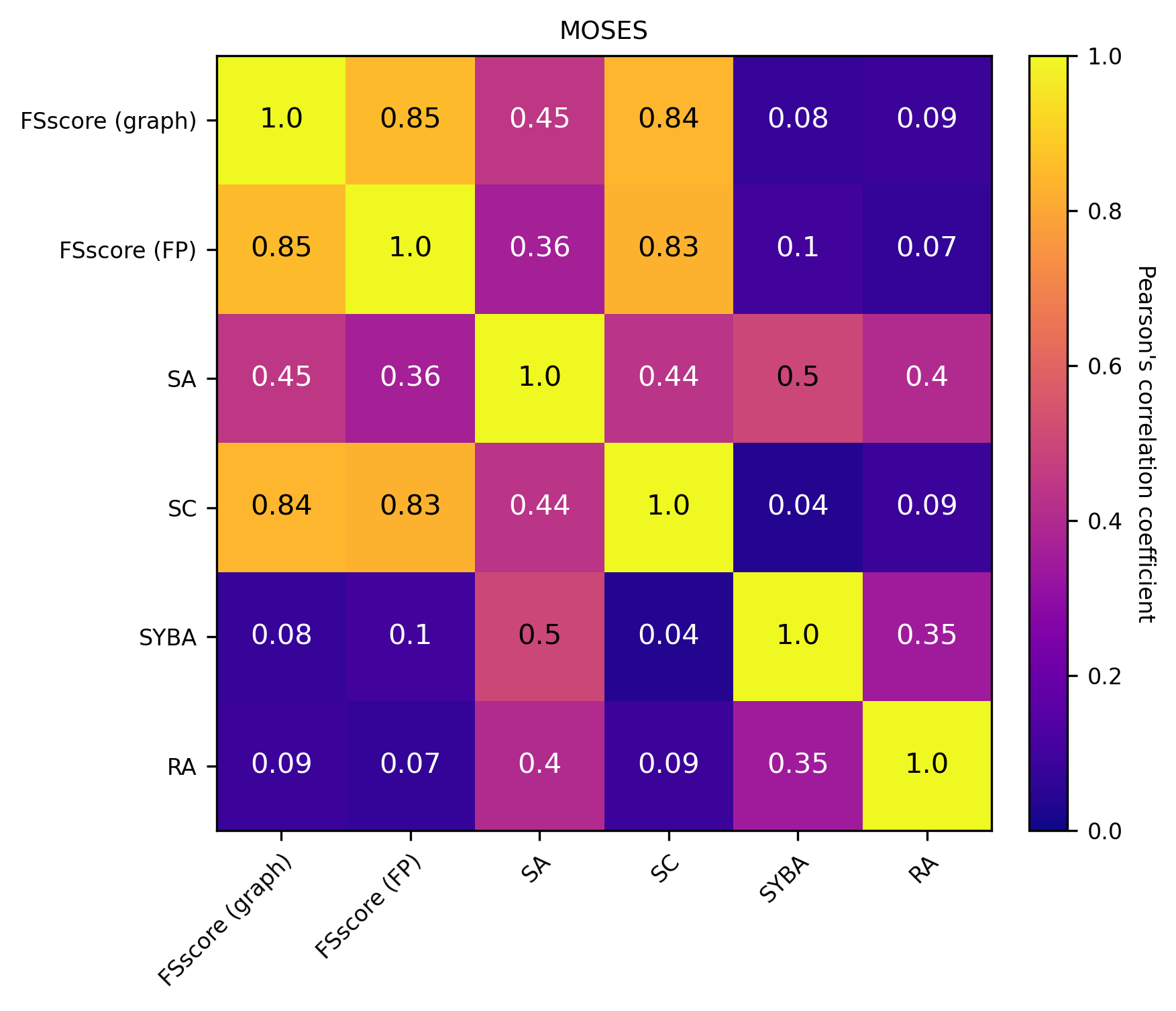}
		\caption{MOSES}
		\label{subfig:SI_moses_pcc}
	\end{subfigure} \hfill
	\begin{subfigure}[t]{0.49\textwidth}
		\centering
		\includegraphics[width=\textwidth]{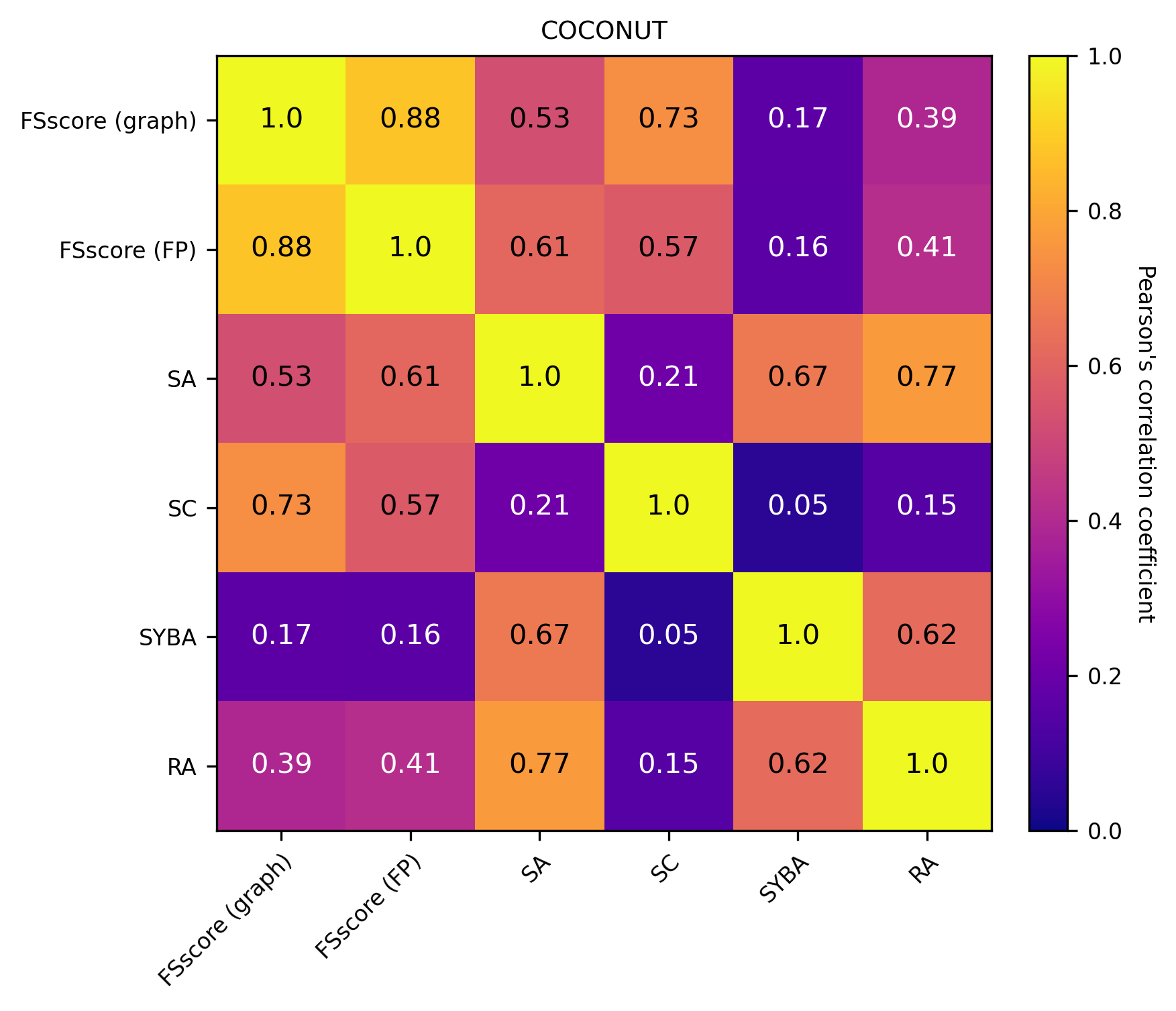}
		\caption{COCONUT}
		\label{subfig:SI_coconut_pcc}
	\end{subfigure} 
	\caption{Heat maps displaying the correlations (\textit{PCC}) between all scores obtained on molecules from MOSES and COCONUT.}
	\label{fig:SI_drugs_pcc}
\end{figure}

\subsection{SYBA test sets}
\label{sec:SI_ft_results}

\begin{table}[H]
\caption{Performance metrics showcasing the improvement on specific datasets after fine-tuning at different fine-tuning dataset sizes (\bm{$p_{ft}$} pairs). No FT refers to the performance of the pre-trained model on those datasets. \(Acc_{pt}\) and \(AUC_{pt}\) are determined on the pre-training test set and are based on the score difference (as during training) not the score itself. The values for the fine-tuned versions always show metrics on the dataset excluding the training molecules and the full dataset in brackets.
}
\begin{scriptsize}
\begin{center}
\label{tab:SI_ft}
\begin{tabular}{lllllll}
\toprule
\textbf{set}                    & \textbf{mode}                  & \bm{$p_{ft}$}        & \bm{$Acc$}      & \bm{$AUC$}     & \bm{$Acc_{pt}$} & \bm{$AUC_{pt}$} \\ \midrule
\multirow{10}{*}{chiral}        & \multirow{5}{*}{graph}         & no FT                & 0.5435          & 0.5391         & 0.905      & 0.971       \\ \cmidrule(l){3-7} 
                                &                                & 20                   & 0.5762 (0.575)  & 0.5972 (0.5947)& 0.9044     & 0.9706     \\
                                &                                & 30                   & 0.5918 (0.592)  & 0.6185 (0.6173)& 0.9035     & 0.97       \\
                                &                                & 40                   & 0.6156 (0.6145) & 0.6539 (0.6521)& 0.9022     & 0.9692     \\
                                &                                & 50                   & 0.6268 (0.6235) & 0.6749 (0.6728)& 0.8983     & 0.9671     \\ \cmidrule(l){2-7} 
                                & \multirow{5}{*}{fp}            & no FT                & 0.509           & 0.4989         & 0.875      & 0.957       \\ \cmidrule(l){3-7} 
                                &                                & 20                   & 0.5163 (0.5165) & 0.5082 (0.5102)& 0.879      & 0.959      \\
                                &                                & 30                   & 0.5180 (0.5195) & 0.5122 (0.5153)& 0.8786     & 0.9583     \\
                                &                                & 40                   & 0.5255 (0.5255) & 0.5180 (0.5212)& 0.8738     & 0.9544     \\
                                &                                & 50                   & 0.5237 (0.5235) & 0.5147 (0.5174)& 0.8783     & 0.9582     \\ \midrule
\multirow{10}{*}{CP~\cite{vorsilak_syba_2020}}& \multirow{5}{*}{graph}& no FT           & 0.6319          & 0.6446         & 0.905      & 0.971      \\ \cmidrule(l){3-7} 
                                &                                & 20                   & 0.9275 (0.9273) & 0.9804 (0.9802)& 0.903      & 0.9700     \\
                                &                                & 30                   & 0.943 (0.9429)  & 0.9868 (0.9865)& 0.9021     & 0.9695     \\
                                &                                & 40                   & 0.9592 (0.9587) & 0.9933 (0.9930)& 0.9012     & 0.9689     \\
                                &                                & 50                   & 0.9567 (0.9560) & 0.9921 (0.9918)& 0.9008     & 0.9687     \\ \cmidrule(l){2-7} 
                                & \multirow{5}{*}{fp}            & no FT                & 0.7008          & 0.7566         & 0.880      & 0.959       \\ \cmidrule(l){3-7} 
                                &                                & 20                   & 0.8683 (0.8685) & 0.9396 (0.9397)& 0.8771     & 0.9568     \\
                                &                                & 30                   & 0.8833 (0.8838) & 0.9517 (0.9521)& 0.8741     & 0.9543     \\
                                &                                & 40                   & 0.9043 (0.9045) & 0.9645 (0.9649)& 0.8732     & 0.9536     \\
                                &                                & 50                   & 0.9167 (0.9178) & 0.9725 (0.9731)& 0.8708     & 0.952      \\ \midrule
\multirow{8}{*}{MC~\cite{vorsilak_syba_2020}}& \multirow{4}{*}{graph}& no FT            & 0.5375          & 0.4894         & 0.905      & 0.971      \\ \cmidrule(l){3-7} 
                                &                                & 20                   & 0.65 (0.7875)   & 0.645 (0.7744) & 0.9034     & 0.9695     \\
                                &                                & 30                   & 0.6 (0.7625)    & 0.55 (0.8150)  & 0.9034     & 0.9696     \\
                                &                                & 40                   & (0.85)          & (0.906)        & 0.8998     & 0.9679     \\ \cmidrule(l){2-7} 
                                & \multirow{4}{*}{fp}            & no FT                & 0.5875          & 0.515          & 0.880      & 0.959      \\ \cmidrule(l){3-7} 
                                &                                & 20                   & 0.6 (0.7125)    & 0.5425 (0.7832)& 0.8820     & 0.9596     \\
                                &                                & 30                   & 0.7 (0.8875)    & 0.61 (0.8994)  & 0.8734     & 0.9529     \\
                                &                                & 40                   & (0.825)         & (0.8513)       & 0.8786     & 0.9579     \\ \midrule
\multirow{8}{*}{PROTAC-DB~\cite{weng_protac-db_2022}}& \multirow{8}{*}{graph} & no FT   & 0.53            & 0.4341         & 0.905      & 0.971     \\ \cmidrule(l){3-7} 
                                &                                & 20                   & 0.5063 (0.515)  & 0.4341 (0.4602)& 0.905      & 0.9715     \\
                                &                                & 30                   & 0.5214 (0.555)  & 0.4587 (0.5474)& 0.9023     & 0.9697     \\
                                &                                & 40                   & 0.5083 (0.51)   & 0.4583 (0.4781)& 0.9048     & 0.9712     \\
                                &                                & 50                   & 0.57 (0.6)      & 0.522 (0.6142) & 0.8993     & 0.968      \\
                                &                                & 60                   & 0.55 (0.615)    & 0.5097 (0.6389)& 0.8984     & 0.9672     \\
                                &                                & 70                   & 0.5167 (0.675)  & 0.4433 (0.6857)& 0.8954     & 0.9659     \\
                                &                                & 80                   & 0.55 (0.685)    & 0.4725 (0.7281)& 0.8975     & 0.9672     \\ \midrule
\multirow{8}{*}{generated}      & \multirow{8}{*}{graph}         & no FT                & 0.5594          & 0.5255         & 0.905      & 0.971     \\ \cmidrule(l){3-7} 
                                &                                & 20                   & 0.5741 (0.5842) & 0.5437 (0.5629)& 0.8971     & 0.9675     \\
                                &                                & 30                   & 0.5775 (0.6139) & 0.5444 (0.6117)& 0.8969     & 0.9673     \\
                                &                                & 40                   & 0.5656 (0.6535) & 0.5612 (0.6686)& 0.8942     & 0.9659     \\
                                &                                & 50                   & 0.5882 (0.604)  & 0.5692 (0.589) & 0.9012     & 0.9694     \\
                                &                                & 60                   & 0.5823 (0.6485) & 0.5756 (0.6566)& 0.8887     & 0.9632     \\
                                &                                & 70                   & 0.6034 (0.6287) & 0.6177 (0.6527)& 0.8907     & 0.9643     \\
                                &                                & 80                   & 0.6316 (0.7376) & 0.6357 (0.7752)& 0.8847     & 0.9616     \\ \bottomrule
\end{tabular}
\end{center}
\end{scriptsize}
\end{table}



\begin{table}[H]
\caption{Performance metrics showcasing the improvement on specific datasets after fine-tuning at different fine-tuning dataset sizes with overlapping pairs (molecules can appear in multiple pairs). No FT refers to the performance of the pre-trained model on those datasets. \(Acc_{pt}\) and \(AUC_{pt}\) are determined on the pre-training test set and are based on the score difference (as during training) not the score itself. The values for the fine-tuned versions always show metrics on the dataset excluding the training molecules (with evaluation size \bm{$n_{eval}$}) and the full dataset in brackets.
}
\begin{center}
\label{tab:SI_ft_rpt}
\begin{tabular}{llllllll}
\toprule
\textbf{set}  & \textbf{mode}           & \bm{$p_{ft}$} & \bm{$n_{eval}$} & \bm{$Acc$}      & \bm{$AUC$}     & \bm{$Acc_{pt}$} & \bm{$AUC_{pt}$} \\ \midrule
\multirow{14}{*}{CP~\cite{vorsilak_syba_2020}}
              & \multirow{7}{*}{graph}  & no FT         & 7162            & 0.6319          & 0.6446         & 0.905           & 0.971        \\ \cmidrule(l){3-8} 
              &                         & 20            & 7122            & 0.8869 (0.8869) & 0.9543 (0.9544)& 0.9037          & 0.9639       \\
              &                         & 50            & 7062            & 0.9598 (0.9598) & 0.9933 (0.9933)& 0.9008          & 0.9622       \\
              &                         & 100           & 6962            & 0.9779 (0.9779) & 0.9978 (0.9978)& 0.8984          & 0.9609       \\
              &                         & 500           & 6162            & 0.9982 (0.9982) & 1.0 (1.0)      & 0.8773          & 0.9469       \\
              &                         & 1000          & 5163            & 0.9983 (0.9983) & 1.0 (1.0)      & 0.8662          & 0.941        \\
              &                         & 2000          & 3371            & 0.9990 (0.9990) & 1.0 (1.0)      & 0.8497          & 0.9318       \\ \cmidrule(l){2-8} 
              & \multirow{7}{*}{fp}     & no FT         & 7162            & 0.7008          & 0.7566         & 0.880           & 0.959        \\ \cmidrule(l){3-8} 
              &                         & 20            & 7122            & 0.8489 (0.8489) & 0.9201 (0.9207)& 0.8761          & 0.9471       \\
              &                         & 50            & 7062            & 0.8833 (0.8830) & 0.9517 (0.9487)& 0.8721          & 0.944       \\
              &                         & 100           & 6962            & 0.9102 (0.9102) & 0.9679 (0.9689)& 0.8688          & 0.9403       \\
              &                         & 500           & 6162            & 0.9789 (0.9789) & 0.9970 (0.9974)& 0.8564          & 0.9285       \\
              &                         & 1000          & 5164            & 0.9901 (0.9901) & 0.9989 (0.9993)& 0.8415          & 0.9168       \\
              &                         & 2000          & 3372            & 0.9965 (0.9965) & 0.9997 (0.9999)& 0.8179          & 0.8981        \\ \midrule
\multirow{10}{*}{MC~\cite{vorsilak_syba_2020}}
              & \multirow{5}{*}{graph}  & no FT         & 80              & 0.5375          & 0.4894         & 0.905           & 0.971        \\ \cmidrule(l){3-8}
              &                         & 20            & 59              & 0.6625 (0.6625) & 0.5829 (0.6656)& 0.8985          & 0.9607       \\
              &                         & 50            & 41              & 0.775 (0.775)   & 0.7273 (0.8175)& 0.8971          & 0.9593       \\
              &                         & 100           & 32              & 0.8125 (0.8125) & 0.8155 (0.8244)& 0.8995          & 0.9609       \\
              &                         & 500           & 20              & 0.9 (0.9)       & 0.6813 (0.9375)& 0.8698          & 0.9414       \\ \cmidrule(l){2-8} 
              & \multirow{5}{*}{fp}     & no FT         & 80              & 0.5875          & 0.515          & 0.880           & 0.959        \\ \cmidrule(l){3-8} 
              &                         & 20            & 60              & 0.6375 (0.6375) & 0.4733 (0.6313)& 0.8797          & 0.9502       \\
              &                         & 50            & 44              & 0.85 (0.85)     & 0.7789 (0.9106)& 0.8669          & 0.9415       \\
              &                         & 100           & 29              & 0.925 (0.9250)  & 0.8389 (0.9484)& 0.8545          & 0.9312       \\
              &                         & 500           & 20              & 0.95 (0.95)     & 0.8542 (0.9612)& 0.8324          & 0.9118       \\ \bottomrule
\end{tabular}
\end{center}
\end{table}

\subsubsection{CP test set}
 \begin{figure}[H]
	\centering
	\begin{subfigure}[t]{0.49\textwidth}
		\centering
		\includegraphics[width=\textwidth]{img/SYBA/roc_curve_CP_dp50_graph.png}
		\caption{Graph (GGLGGL)}
		\label{subfig:SI_CP_ROC_graph}
	\end{subfigure} \hfill
	\begin{subfigure}[t]{0.49\textwidth}
		\centering
		\includegraphics[width=\textwidth]{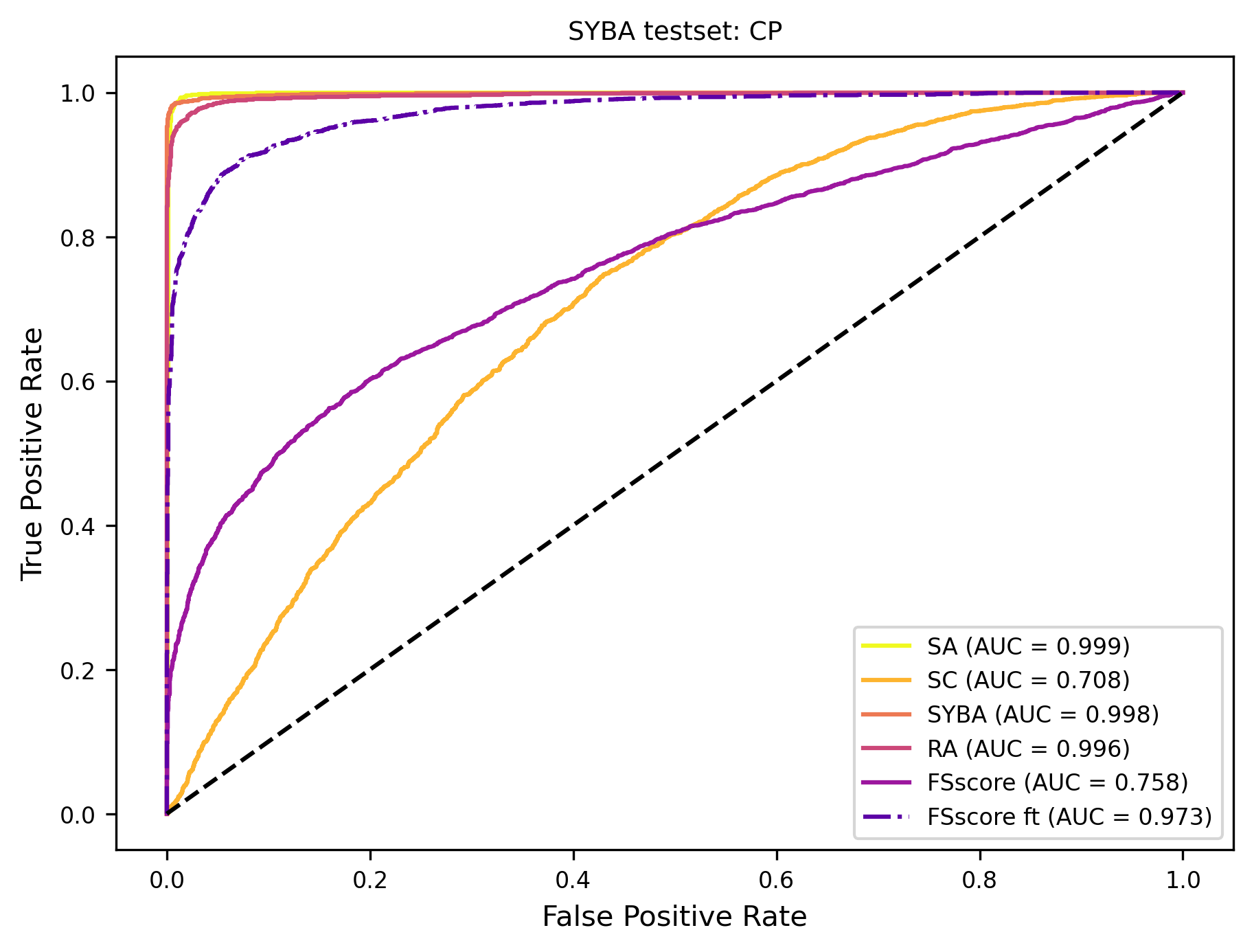}
		\caption{Morgan counts}
		\label{subfig:SI_CP_ROC_fp}
	\end{subfigure} 
	\caption{ROC curves showcasing the ability to distinguish HS from ES in the CP test set using the graph-based FSscore or the fp-based FSscore. The fine-tuning was done with 50~pairs and these 100~molecules were excluded from the plots.}
	\label{fig:SI_CP_ROC}
\end{figure}

 \begin{figure}[H]
	\centering
	\begin{subfigure}[t]{0.49\textwidth}
		\centering
		\includegraphics[width=\textwidth]{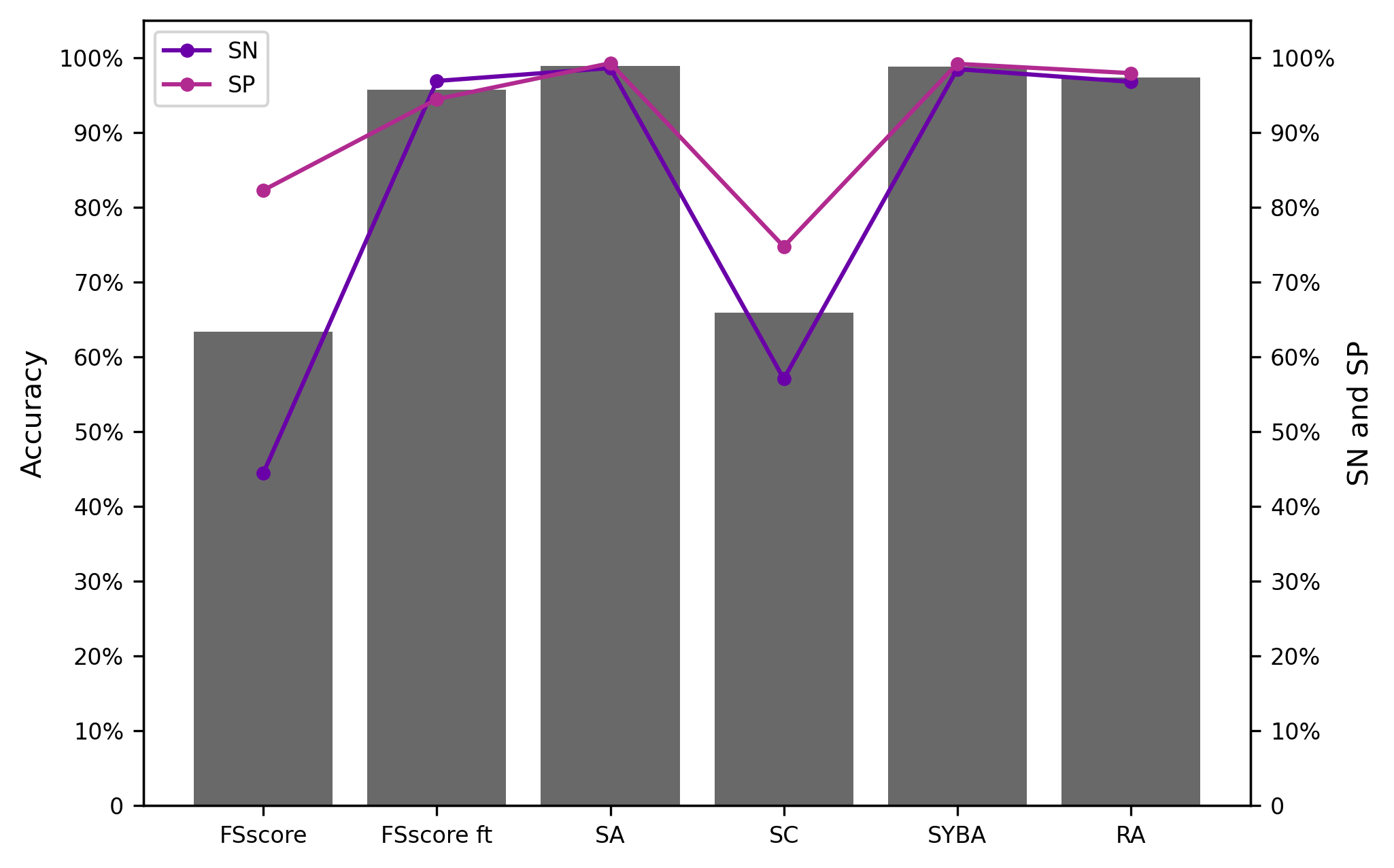}
		\caption{Graph (GGLGGL)}
		\label{subfig:SI_CP_acc_graph}
	\end{subfigure} \hfill
	\begin{subfigure}[t]{0.49\textwidth}
		\centering
		\includegraphics[width=\textwidth]{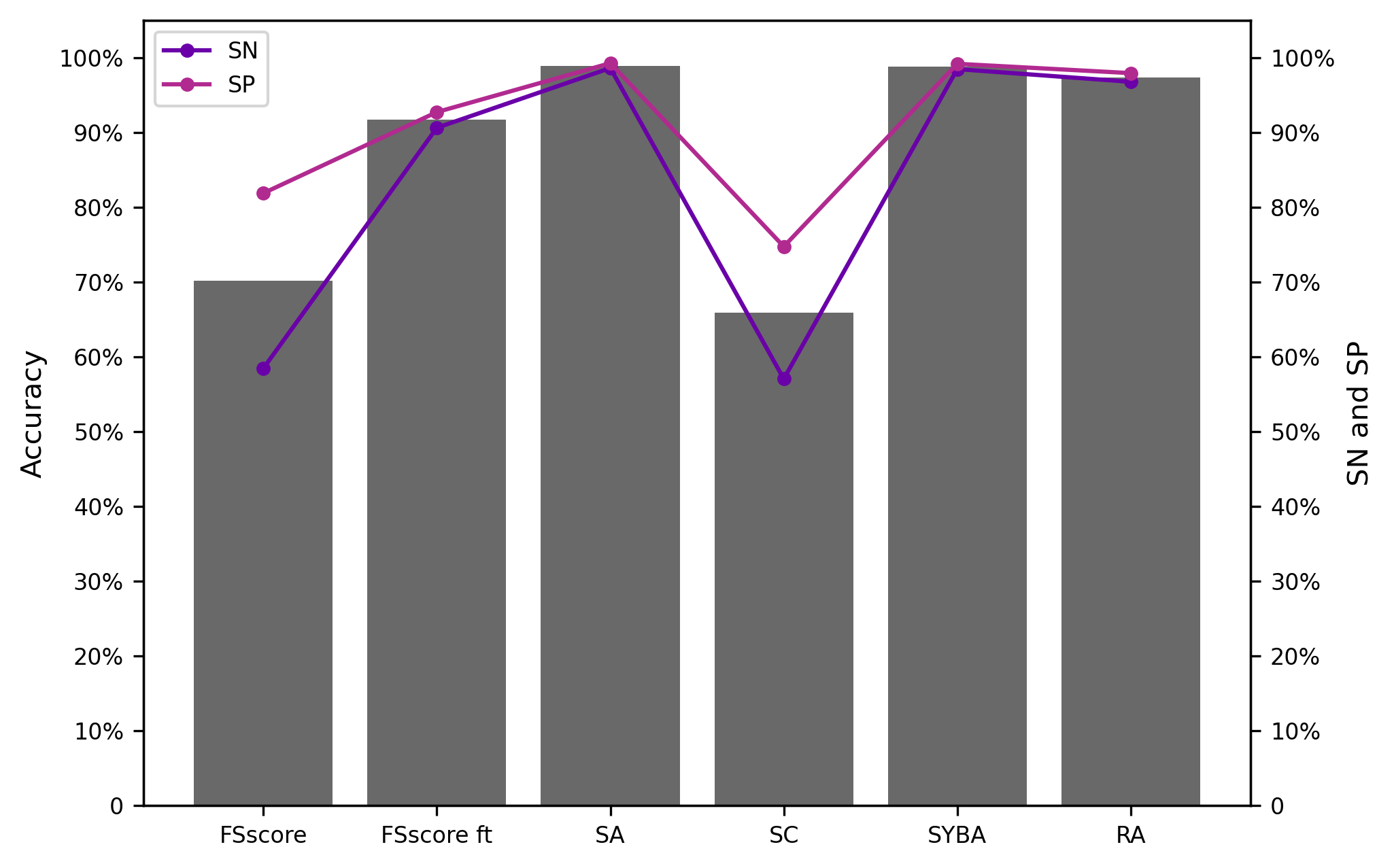}
		\caption{Morgan counts}
		\label{subfig:SI_CP_acc_fp}
	\end{subfigure} 
	\caption{Summary of \textit{Acc}, \textit{SN} and \textit{SP} for classifying the molecules from the CP test set as either HS or ES. The fine-tuning was done with 50~pairs and these 100~molecules were excluded from the plots.}
	\label{fig:SI_CP_acc}
\end{figure}

\subsubsection{MC test set}
 \begin{figure}[H]
	\centering
	\begin{subfigure}[t]{0.49\textwidth}
		\centering
		\includegraphics[width=\textwidth]{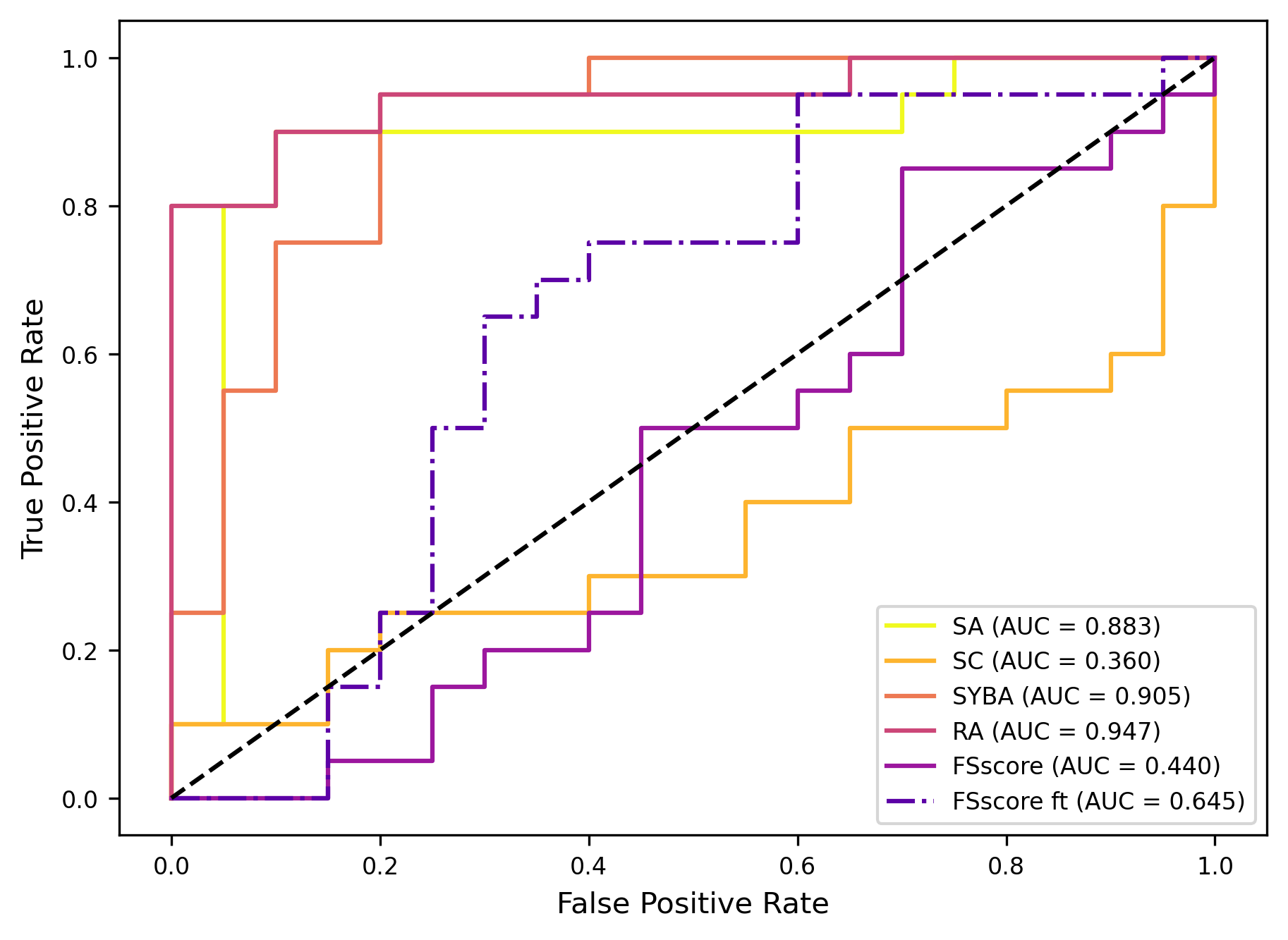}
		\caption{Graph (GGLGGL)}
		\label{subfig:SI_MC_ROC_graph}
	\end{subfigure} \hfill
	\begin{subfigure}[t]{0.49\textwidth}
		\centering
		\includegraphics[width=\textwidth]{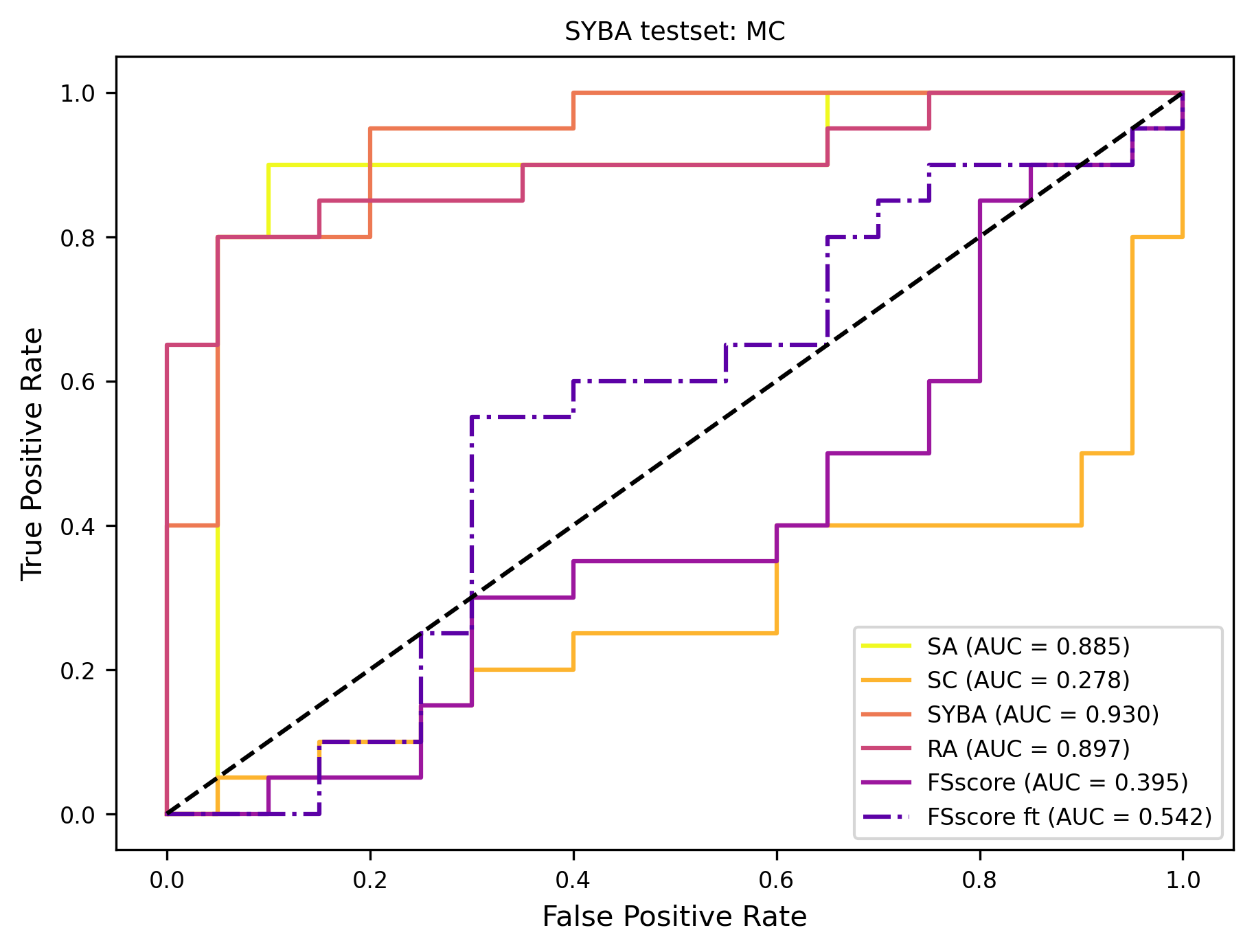}
		\caption{Morgan counts}
		\label{subfig:SI_MC_ROC_fp}
	\end{subfigure} 
	\caption{ROC curves showcasing the ability to distinguish HS from ES in the MC test set using the graph-based FSscore or the fp-based FSscore. The fine-tuning was done with 20~pairs and these 40~molecules were excluded from the plots.}
	\label{fig:SI_MC_ROC}
\end{figure}

 \begin{figure}[H]
	\centering
	\begin{subfigure}[t]{0.49\textwidth}
		\centering
		\includegraphics[width=\textwidth]{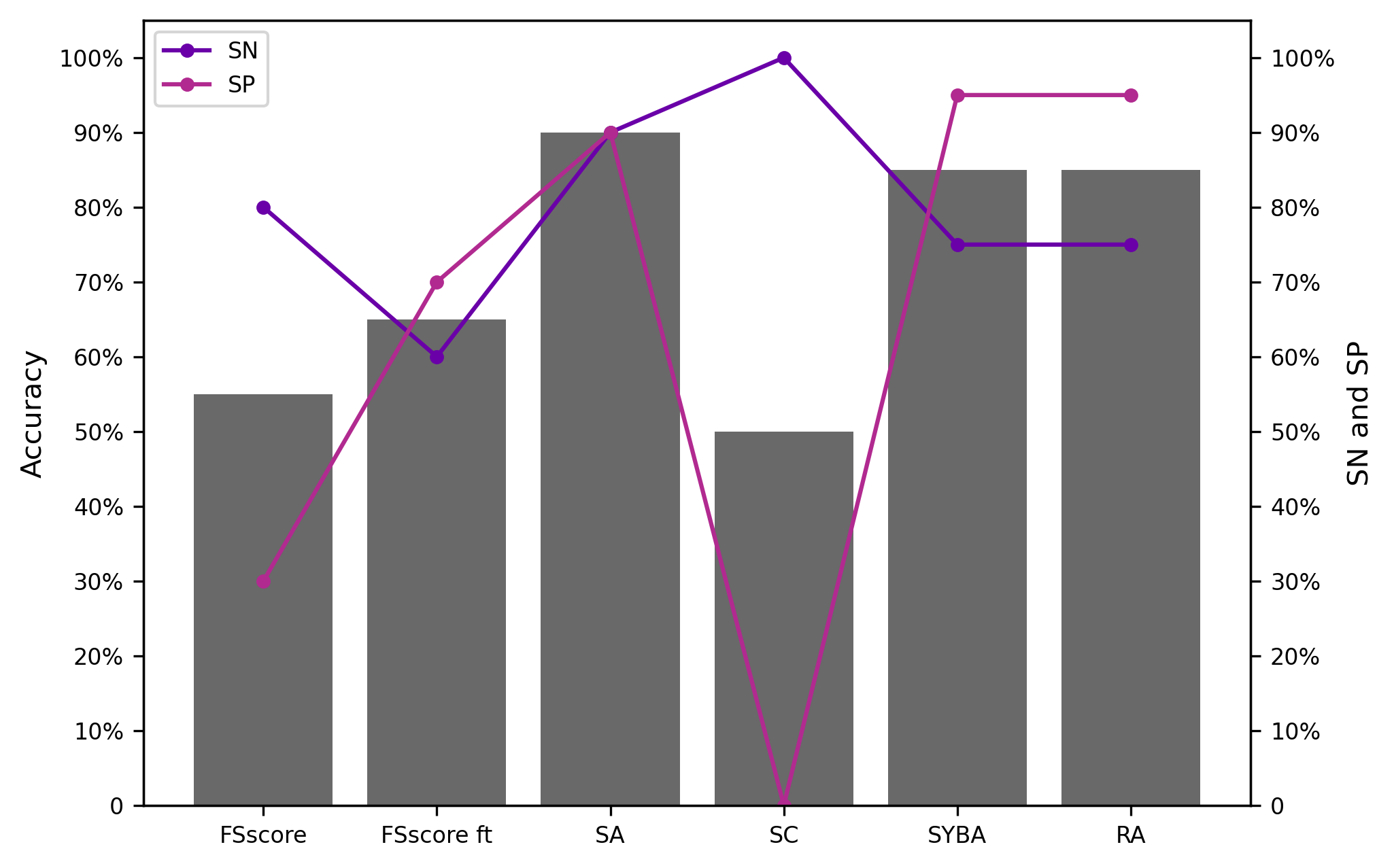}
		\caption{Graph (GGLGGL)}
		\label{subfig:SI_MC_acc_graph}
	\end{subfigure} \hfill
	\begin{subfigure}[t]{0.49\textwidth}
		\centering
		\includegraphics[width=\textwidth]{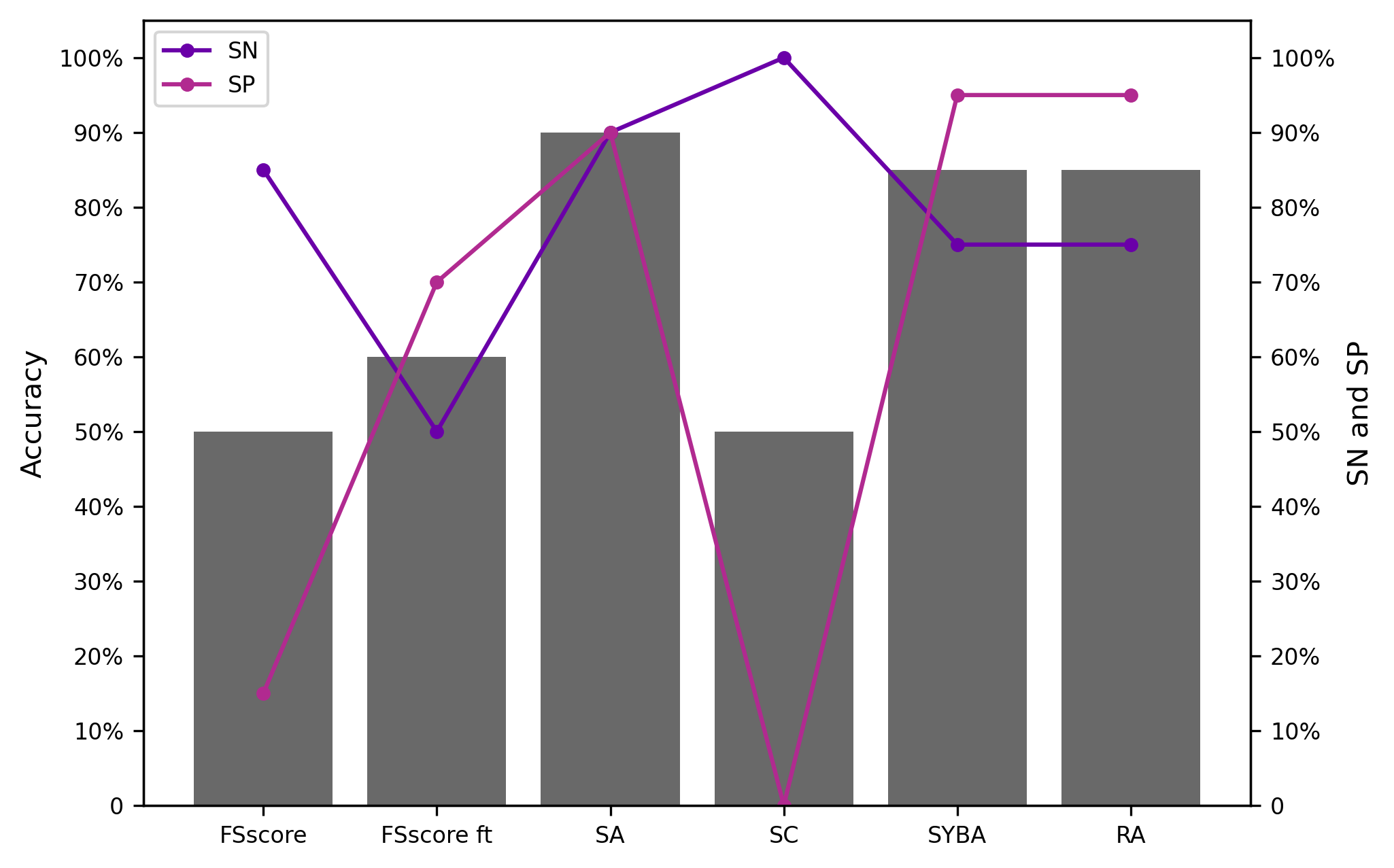}
		\caption{Morgan counts}
		\label{subfig:SI_MC_acc_fp}
	\end{subfigure} 
	\caption{Summary of \textit{Acc}, \textit{SN} and \textit{SP} for classifying the molecules from the MC test set as either HS or ES. The fine-tuning was done with 20~pairs and these 40~molecules were excluded from the plots.}
	\label{fig:SI_MC_acc}
\end{figure}

\subsection{meanComplexity dataset}

\begin{table}[H]
\caption{Performance metrics showcasing the improvement on the meanComplexity dataset (\citet{sheridan_modeling_2014}) after fine-tuning at different fine-tuning dataset sizes with (unique = yes) or without overlapping pairs (molecules can appear in multiple pairs). No FT refers to the performance of the pre-trained model on those datasets. \(Acc_{pt}\) and \(AUC_{pt}\) are determined on the pre-training test set and are based on the score difference (as during training) not the score itself. The values for the fine-tuned versions always show metrics on the dataset excluding the training molecules (with evaluation size \bm{$n_{eval}$}) and the full dataset in brackets.
}
\begin{center}
\label{tab:SI_ft_sheridan}
\begin{tabular}{llllllll}
\toprule

\textbf{dataset}        &\textbf{unique} & \textbf{mode}& \bm{$p_{ft}$} & \bm{$n_{eval}$} & \textbf{\textit{PCC}} & \bm{$Acc_{pt}$} & \bm{$AUC_{pt}$}    \\ \midrule
\multirow{20}{*}{meanComplexity \cite{sheridan_modeling_2014}} &
\multirow{10}{*}{yes} & \multirow{5}{*}{graph}  & no FT & 1681 & 0.51  & 0.905  & 0.971  \\ \cmidrule(l){4-8}
&                       &                       & 20    & 1641 & 0.61  & 0.9034 & 0.9701 \\
&                       &                       & 30    & 1621 & 0.63  & 0.9035 & 0.9697 \\
&                       &                       & 40    & 1601 & 0.66  & 0.9033 & 0.9697 \\
&                       &                       & 50    & 1581 & 0.66  & 0.9028 & 0.9689 \\ \cmidrule(l){3-8}
&                       & \multirow{5}{*}{fp}   & no FT & 1681 & 0.49  & 0.880  & 0.959  \\ \cmidrule(l){4-8}
&                       &                       & 20    & 1641 & 0.58  & 0.8768 & 0.9569 \\
&                       &                       & 30    & 1621 & 0.68  & 0.8668 & 0.9492 \\ 
&                       &                       & 40    & 1601 & 0.68  & 0.8483 & 0.9353 \\
&                       &                       & 50    & 1581 & 0.67  & 0.8674 & 0.9502 \\ \cmidrule(l){2-8}
&\multirow{10}{*}{no}   & \multirow{5}{*}{graph}& no FT & 1681 & 0.51  & 0.905  & 0.971  \\ \cmidrule(l){4-8}
&                       &                       & 20    & 1649 & 0.55  & 0.8995 & 0.9615 \\
&                       &                       & 50    & 1582 & 0.51  & 0.9017 & 0.963  \\
&                       &                       & 100   & 1552 & 0.73  & 0.8921 & 0.9564 \\
&                       &                       & 500   & 1291 & 0.84  & 0.8765 & 0.9447 \\ \cmidrule(l){3-8}
&                       & \multirow{5}{*}{fp}   & no FT & 1681 & 0.49  & 0.880  & 0.959  \\ \cmidrule(l){4-8}
&                       &                       & 20    & 1648 & 0.51  & 0.8809 & 0.9513 \\
&                       &                       & 50    & 1584 & 0.59  & 0.8763 & 0.9475 \\
&                       &                       & 100   & 1538 & 0.7   & 0.8632 & 0.9377 \\
&                       &                       & 500   & 1292 & 0.8   & 0.8412 & 0.9212 \\ \bottomrule
\end{tabular}
\end{center}
\end{table}

 \begin{figure}[H]
	\centering
	\begin{subfigure}[t]{0.32\textwidth}
		\centering
		\includegraphics[width=\textwidth]{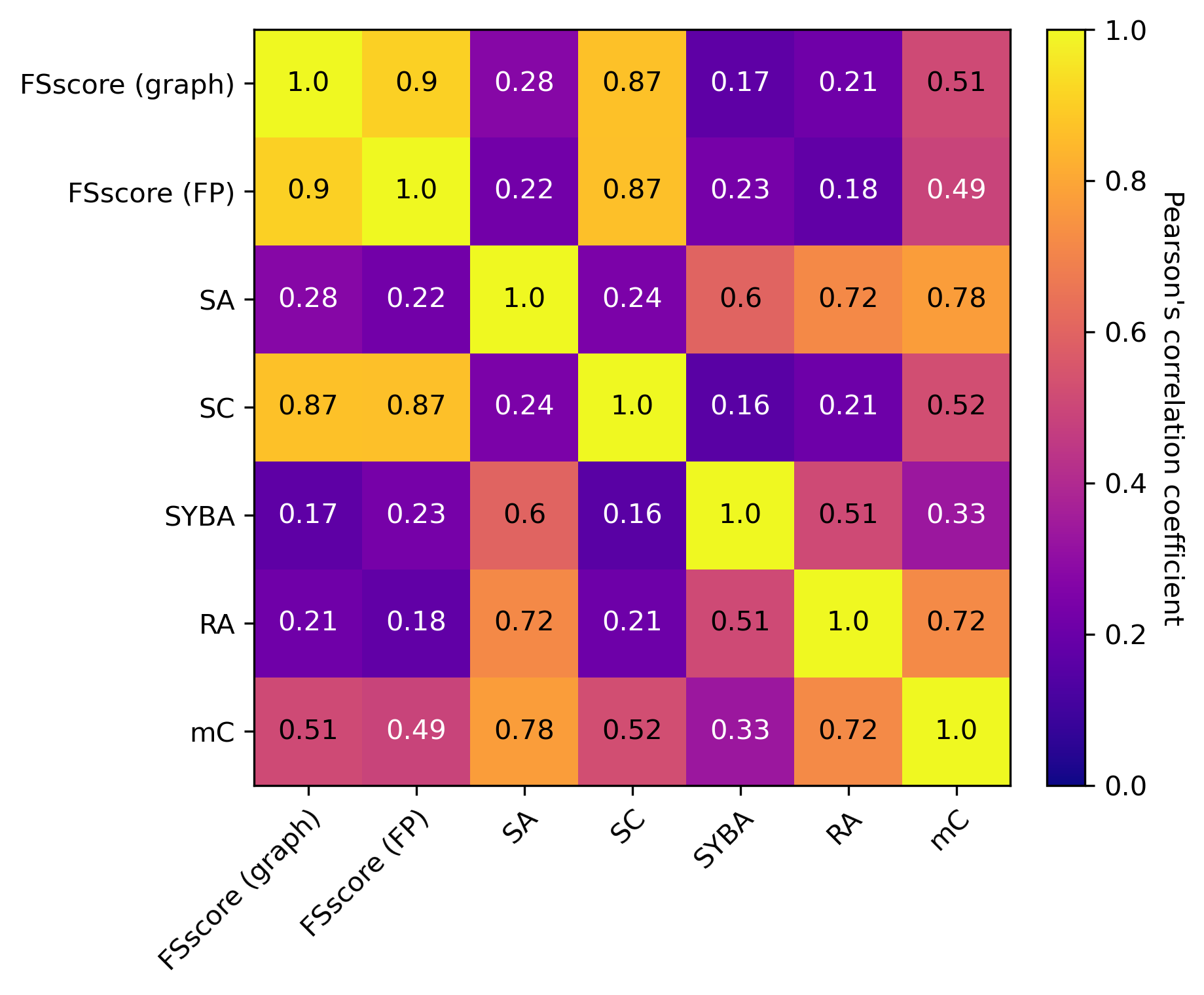}
		\caption{Pre-trained model}
		\label{subfig:SI_sheridan_pcc_pt}
	\end{subfigure} \hfill
	\begin{subfigure}[t]{0.32\textwidth}
		\centering
		\includegraphics[width=\textwidth]{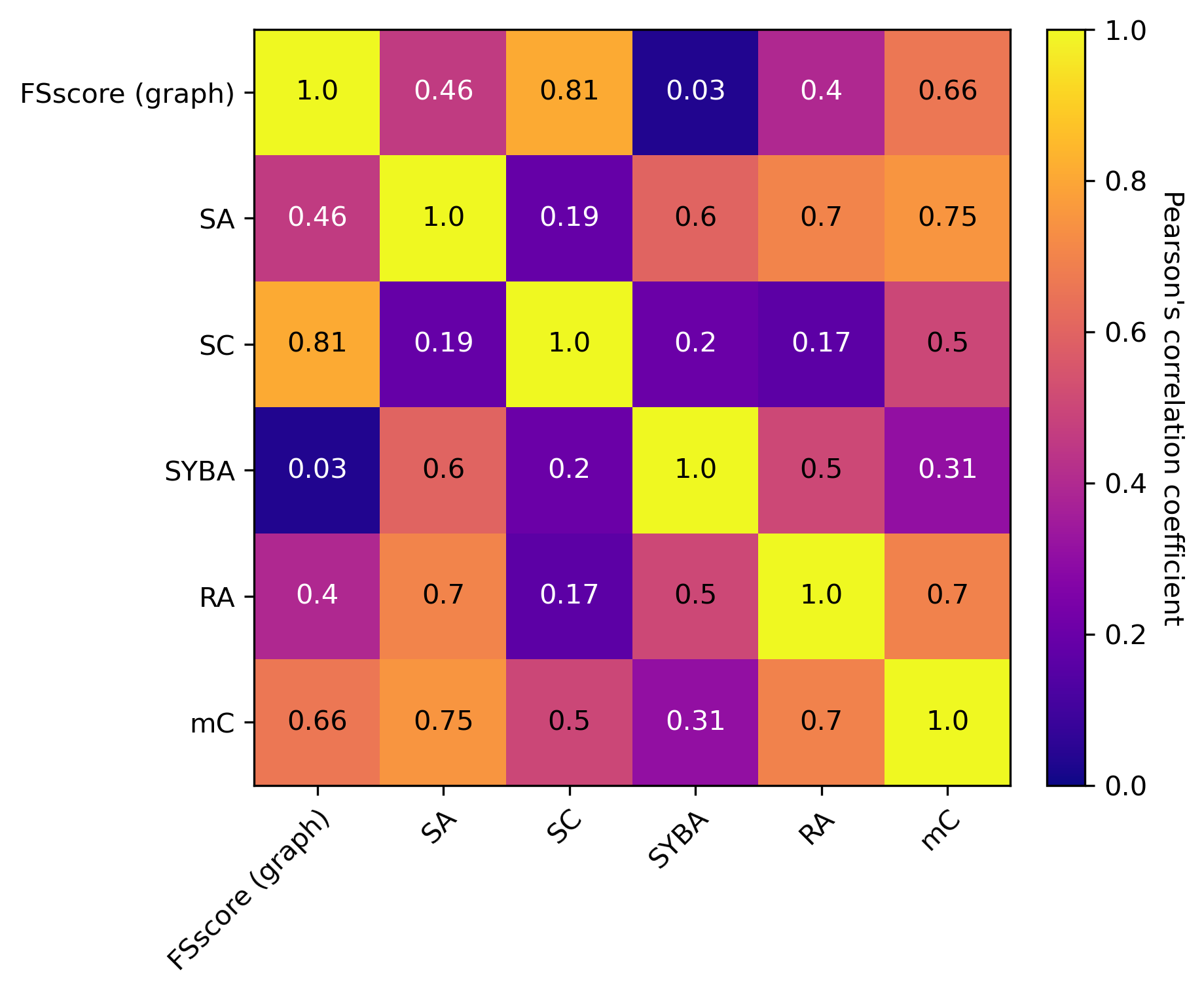}
		\caption{Fine-tuned graph-based model}
		\label{subfig:SI_sheridan_pcc_graph_ft}
	\end{subfigure} 
    \begin{subfigure}[t]{0.32\textwidth}
		\centering
		\includegraphics[width=\textwidth]{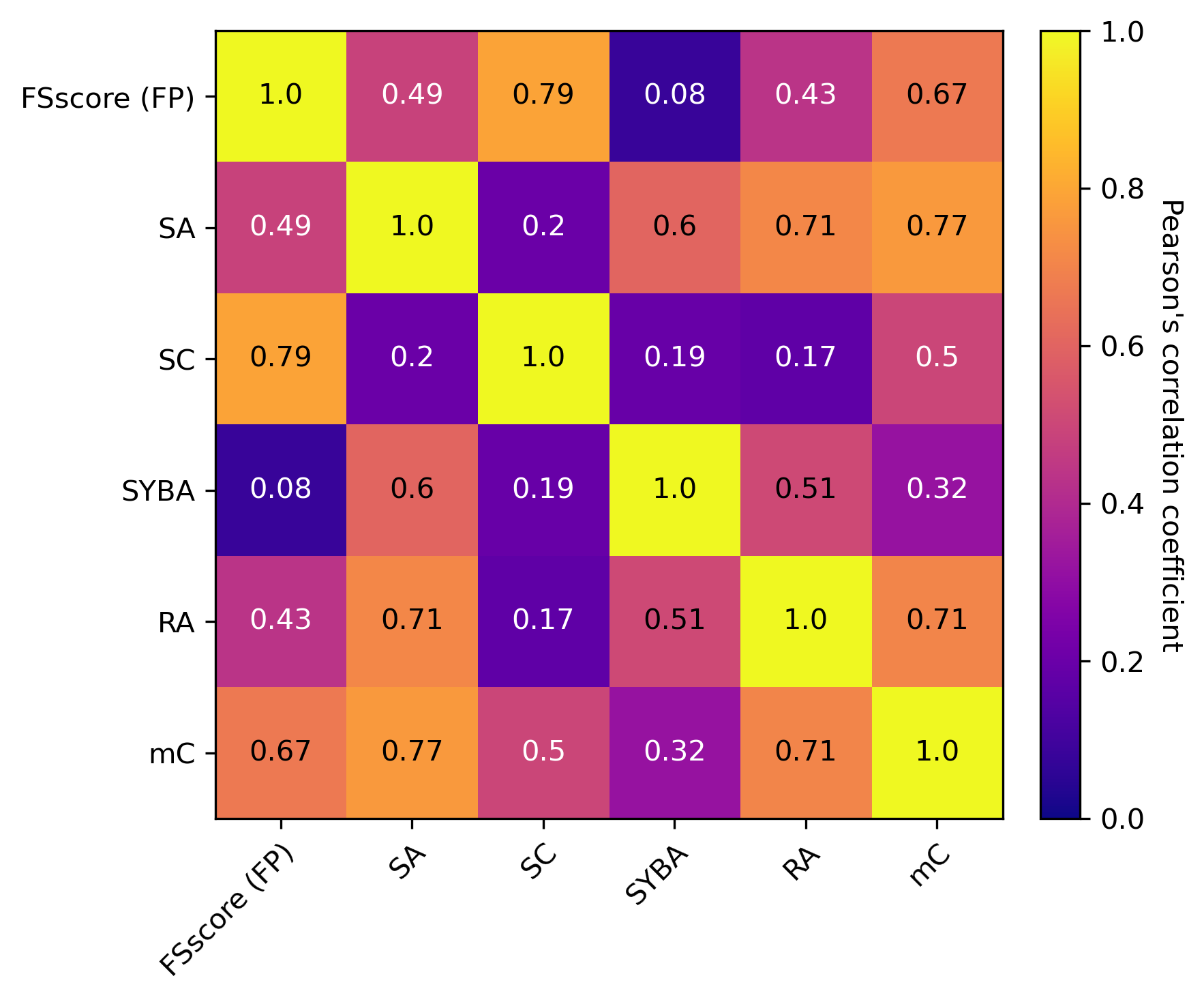}
		\caption{Fine-tuned fp-based model}
		\label{subfig:SI_sheridan_pcc_fp_ft}
	\end{subfigure} \hfill
	\caption{Heat maps displaying the correlations (PCC) between all scores including the meanComplexity~(mC) obtained on molecules from the meanComplexity dataset. The plots based on fine-tuned models excluded the fine-tuning data (50~pairs).}
	\label{fig:SI_sheridan_pcc}
\end{figure}

\begin{figure}[!ht]
    \centering
    \includegraphics[width=0.9\textwidth]{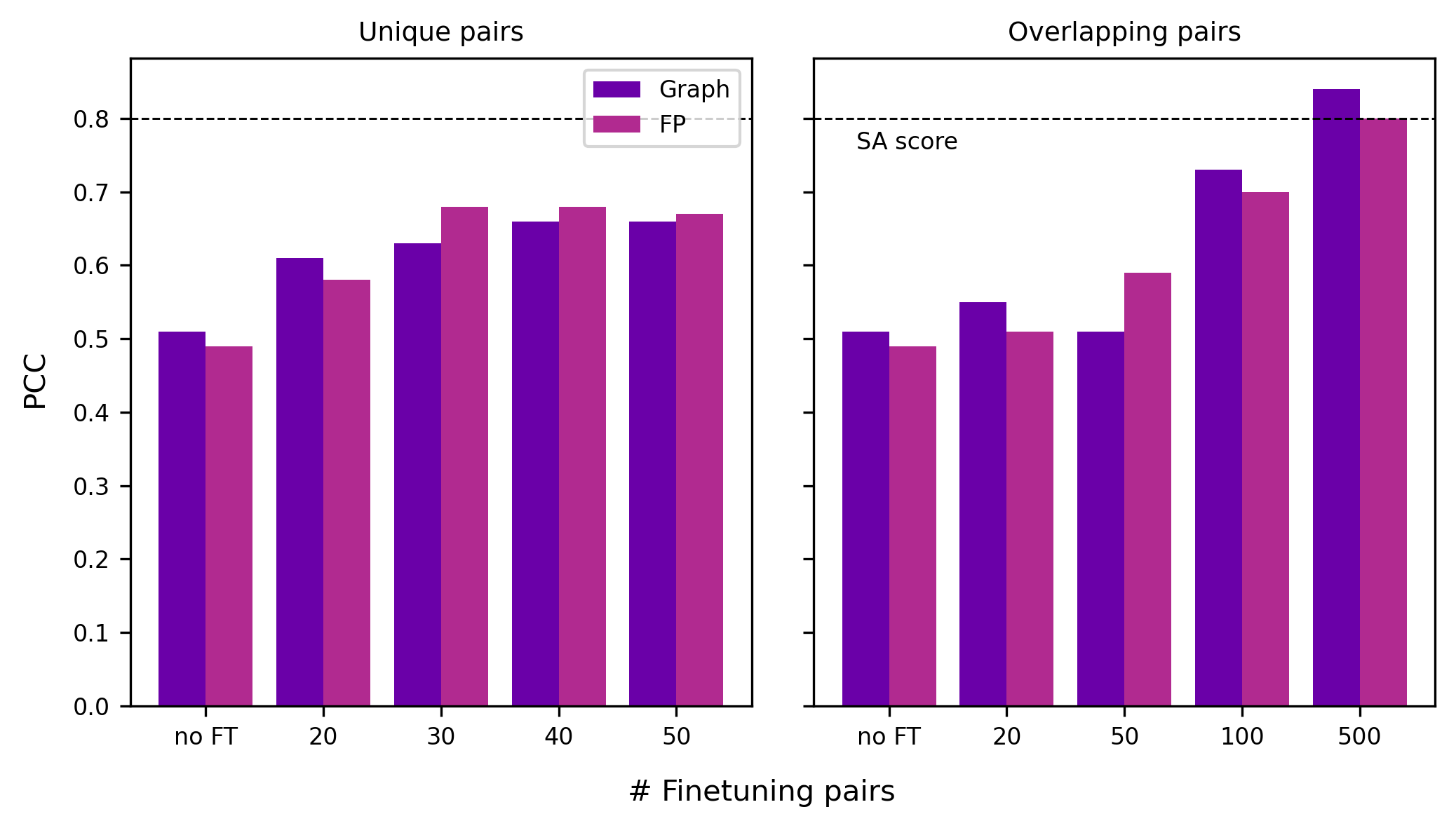}
    \caption{Correlations (PCC) between the meanComplexity and the FSscore fine-tuned with different dataset sizes and with unique (\textit{left}) or overlapping (\textit{right}) pairs for both graph and fingerprint (FP) representations. The dashed line indicates the PCC of the SA score to the meanCompelxity being the best performing score we have tested.}
    \label{fig:SI_sheridan_barplot}
\end{figure}

\section{Example structures}
\label{sec:SI_strucs}

\subsection{Hold-out test set}
\begin{figure}[H]
    \centering
    \includegraphics[width=1\linewidth]{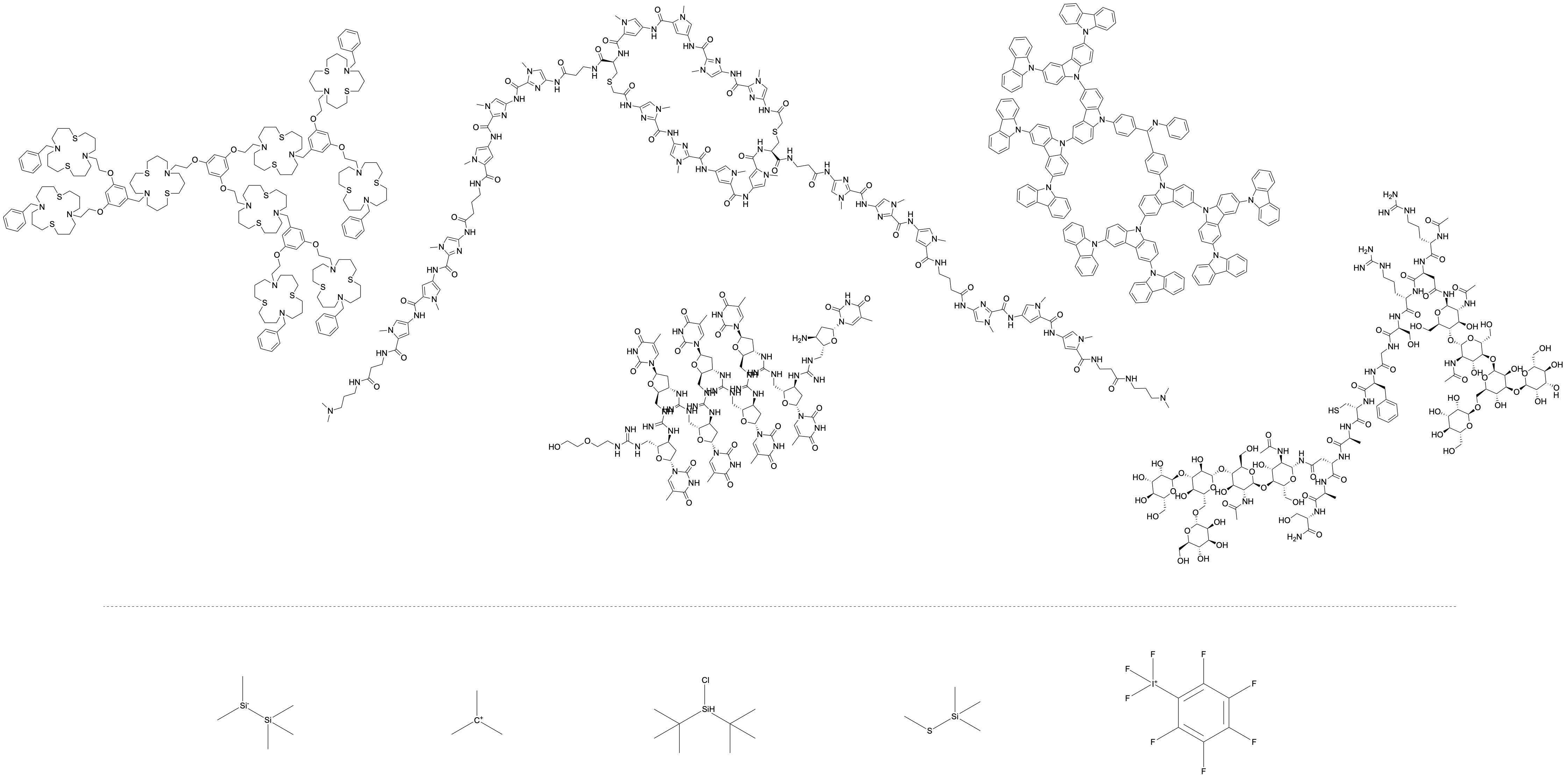}
    \caption{Structures from the hold-out test set scored by the pre-trained graph (\textbf{GGLGGL}) model. \textit{Upper row:} Low FSscore (low synthetic feasibility). \textit{Lower row:} High FSscore (high synthetic feasibility).}
    \label{fig:SI_strucs_combo_graph}
\end{figure}

\begin{figure}[H]
    \centering
    \includegraphics[width=1\linewidth]{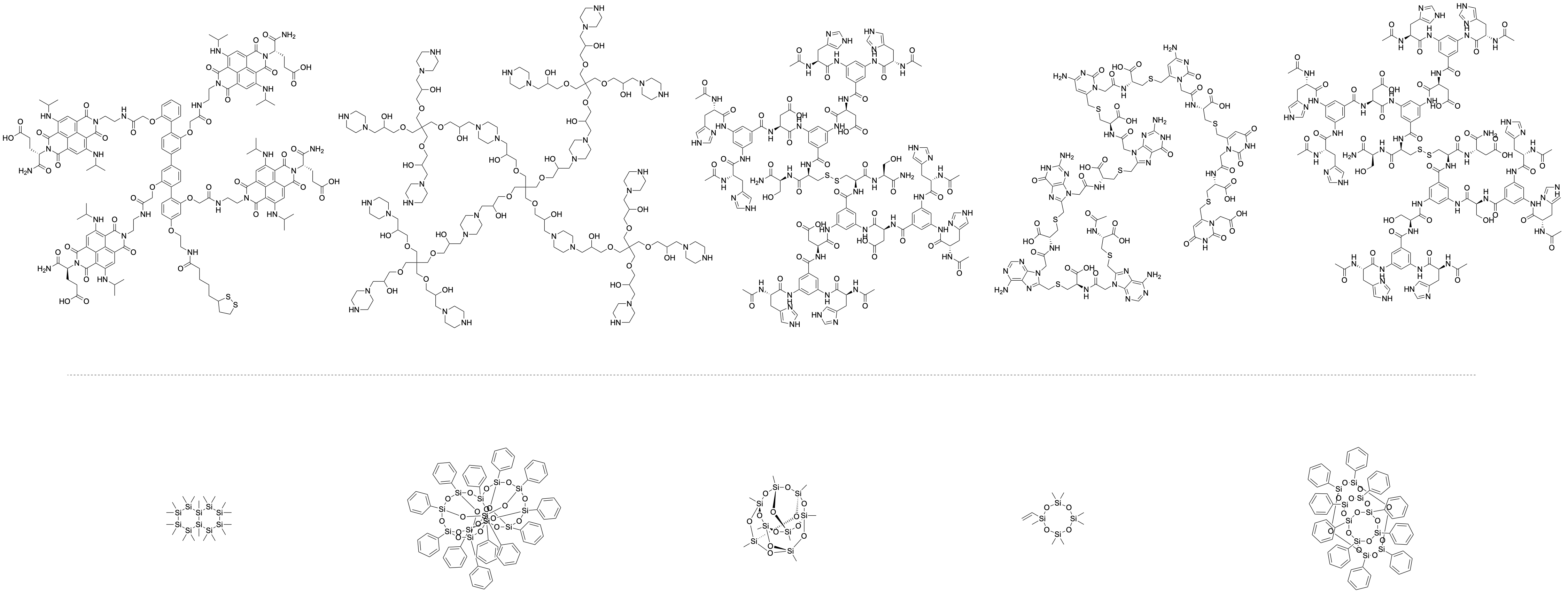}
    \caption{Structures from the hold-out test set scored by the pre-trained fingerprint (\textbf{Morgan counts}) model. \textit{Upper row:} Low FSscore (low synthetic feasibility). \textit{Lower row:} High FSscore (high synthetic feasibility).}
    \label{fig:SI_strucs_combo_morgan_count}
\end{figure}

\begin{figure}[H]
    \centering
    \includegraphics[width=1\linewidth]{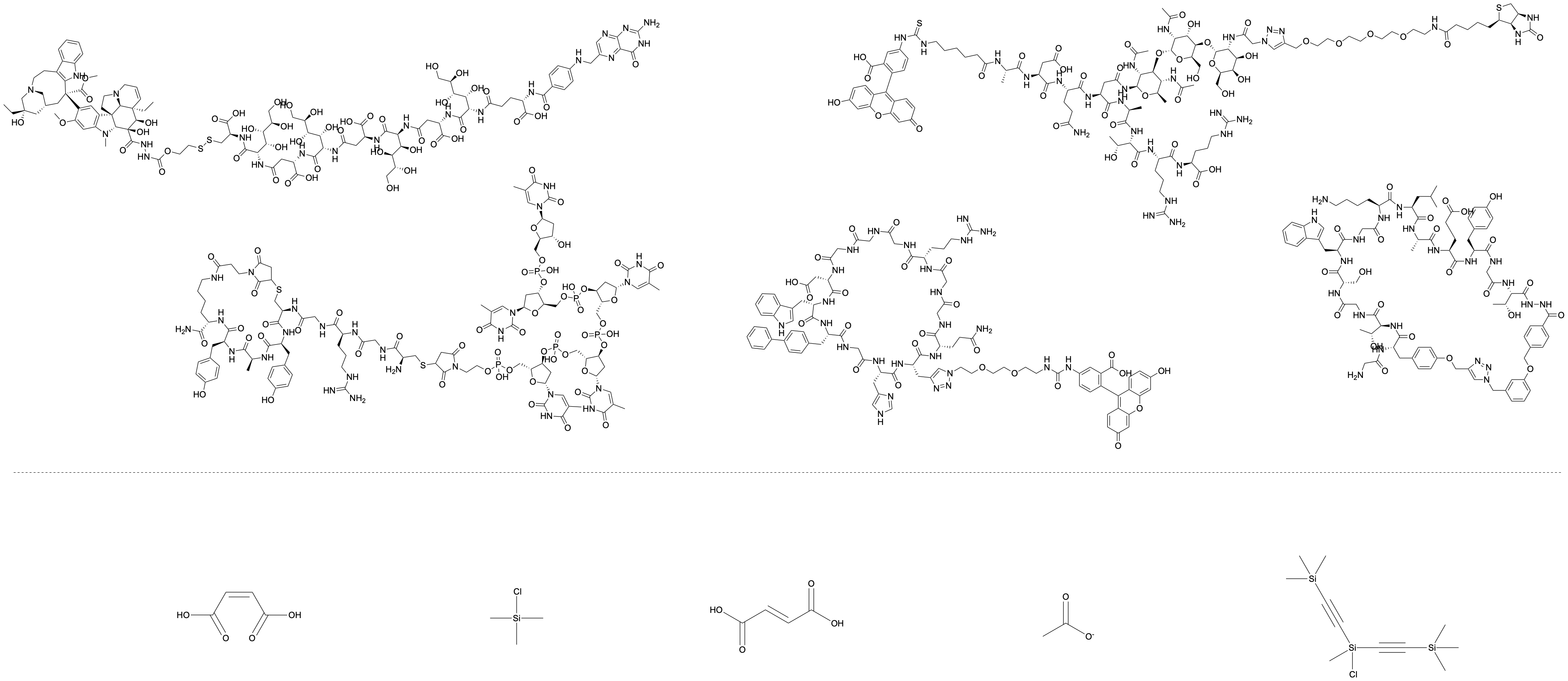}
    \caption{Structures from the hold-out test set scored by the pre-trained fingerprint (\textbf{Morgan boolean}) model. \textit{Upper row:} Low FSscore (low synthetic feasibility). \textit{Lower row:} High FSscore (high synthetic feasibility).}
    \label{fig:SI_strucs_combo_morgan}
\end{figure}

\subsection{CP and MC test sets}
\begin{figure}[H]
    \centering
    \includegraphics[width=1\linewidth]{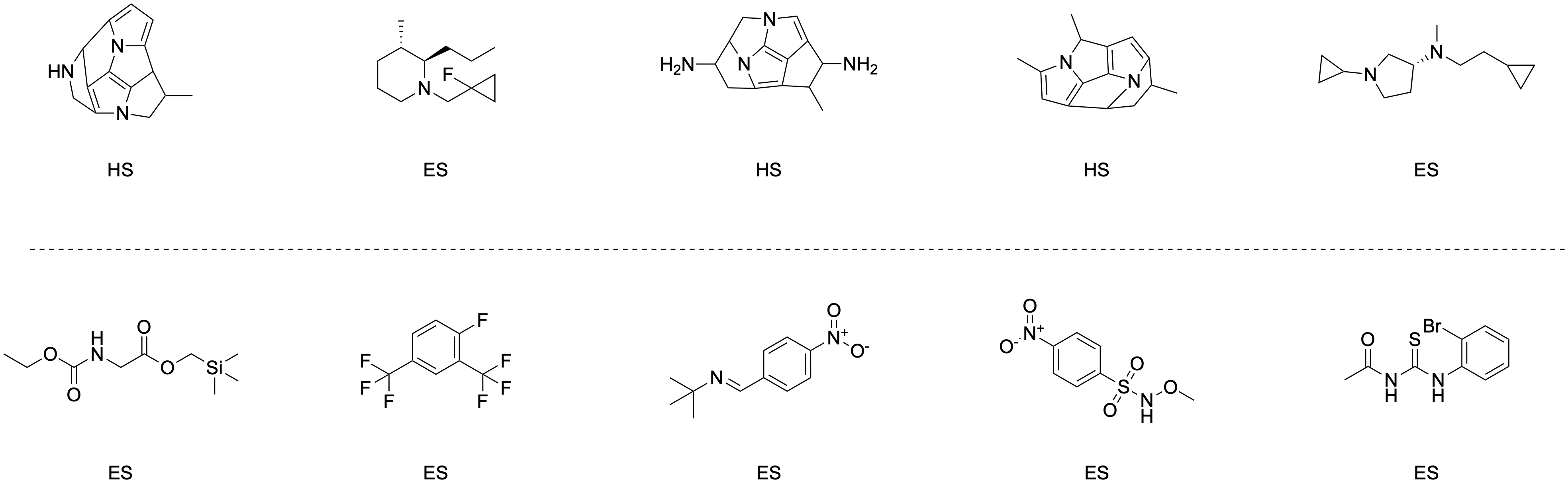}
    \caption{Structures from the CP test set scored by the \textbf{pre-trained} graph (\textbf{GGLGGL}) model. \textit{Upper row:} Low FSscore (low synthetic feasibility). \textit{Lower row:} High FSscore (high synthetic feasibility). The labels below indicate the ground truth.}
    \label{fig:SI_strucs_CP_graph_pt}
\end{figure}

\begin{figure}[H]
    \centering
    \includegraphics[width=1\linewidth]{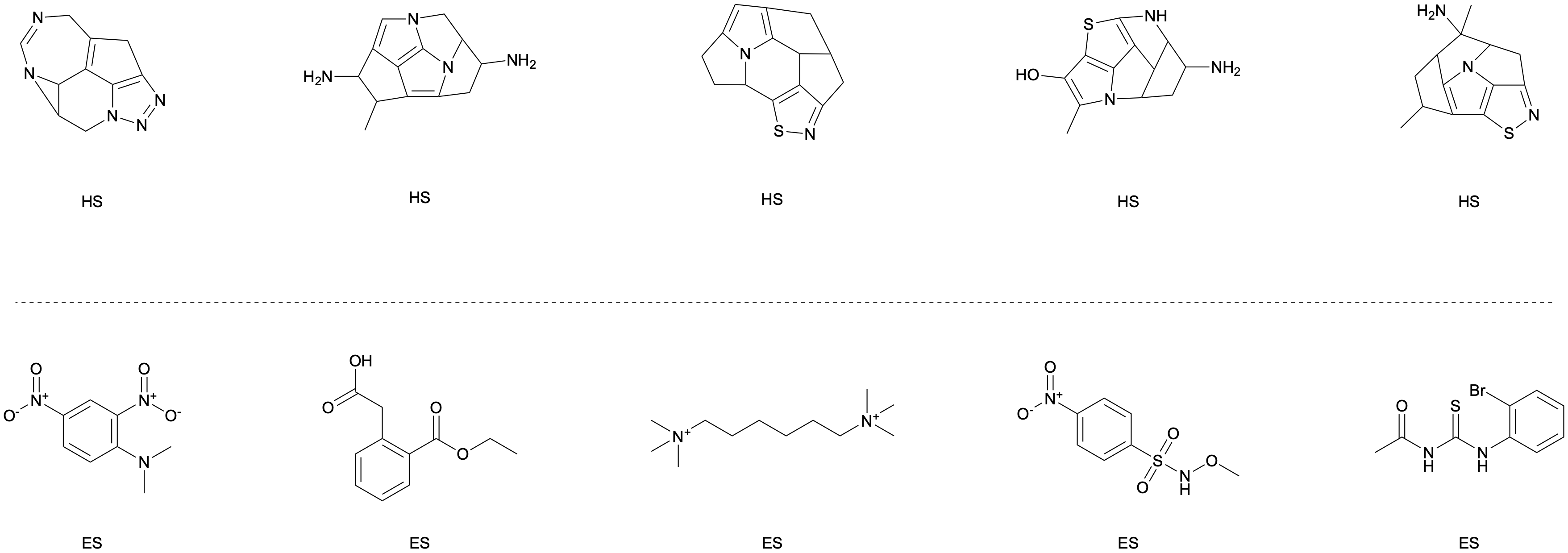}
    \caption{Structures from the CP test set scored by the \textbf{pre-trained} fingerprint (\textbf{Morgan counts}) model. \textit{Upper row:} Low FSscore (low synthetic feasibility). \textit{Lower row:} High FSscore (high synthetic feasibility). The labels below indicate the ground truth.}
    \label{fig:SI_strucs_CP_fp_pt}
\end{figure}

\begin{figure}[H]
    \centering
    \includegraphics[width=1\linewidth]{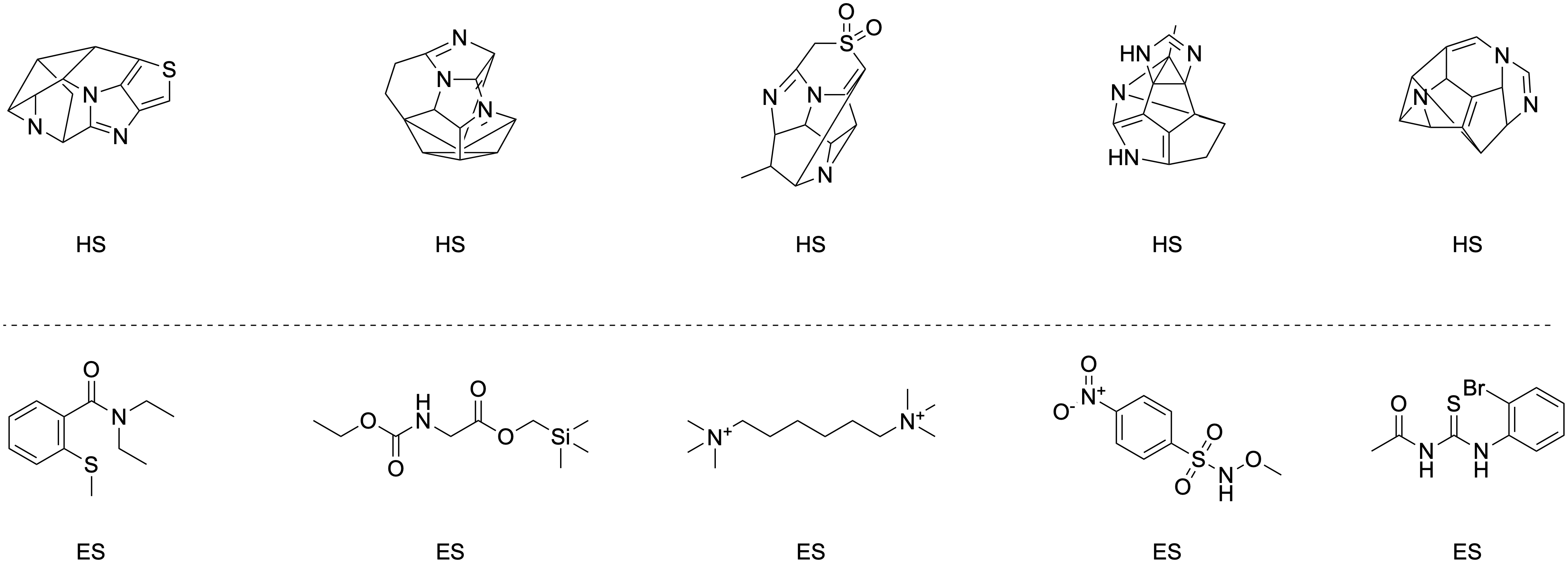}
    \caption{Structures from the CP test set scored by the \textbf{fine-tuned} graph (\textbf{GGLGGL}) model using 50 pairs. \textit{Upper row:} Low FSscore (low synthetic feasibility). \textit{Lower row:} High FSscore (high synthetic feasibility). The labels below indicate the ground truth.}
    \label{fig:SI_strucs_CP_graph_ft}
\end{figure}

\begin{figure}[H]
    \centering
    \includegraphics[width=1\linewidth]{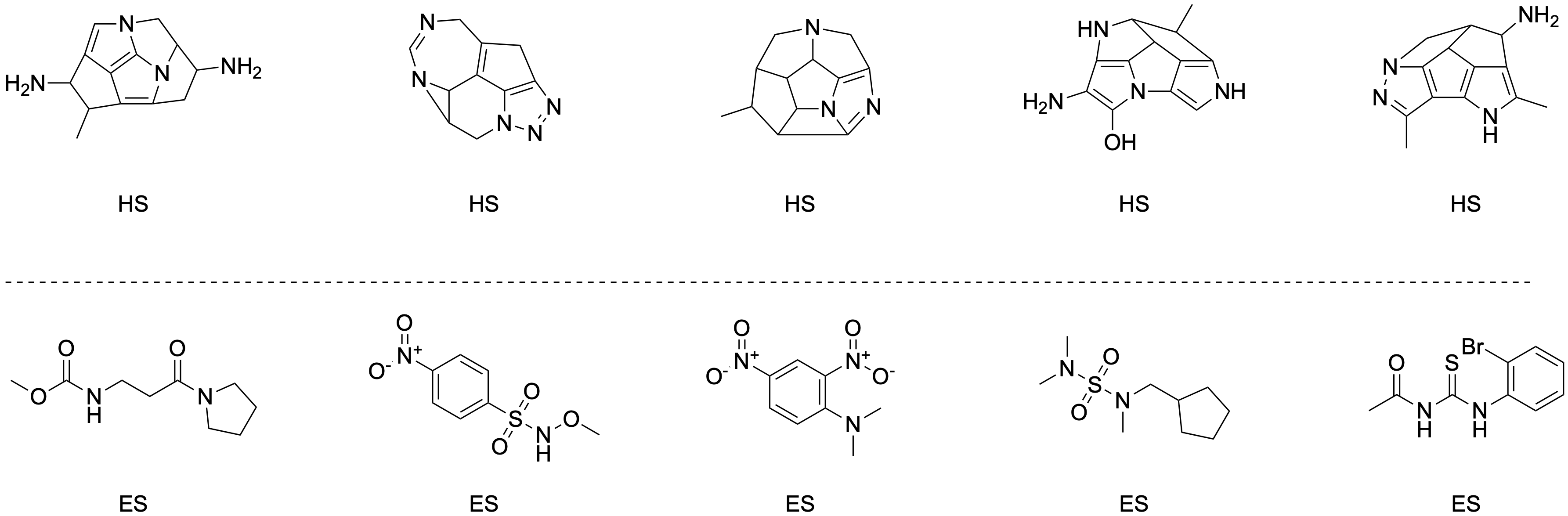}
    \caption{Structures from the CP test set scored by the \textbf{fine-tuned} fingerprint (\textbf{Morgan counts}) model using 50 pairs. \textit{Upper row:} Low FSscore (low synthetic feasibility). \textit{Lower row:} High FSscore (high synthetic feasibility). The labels below indicate the ground truth.}
    \label{fig:SI_strucs_CP_fp_ft}
\end{figure}


\begin{figure}[H]
    \centering
    \includegraphics[width=1\linewidth]{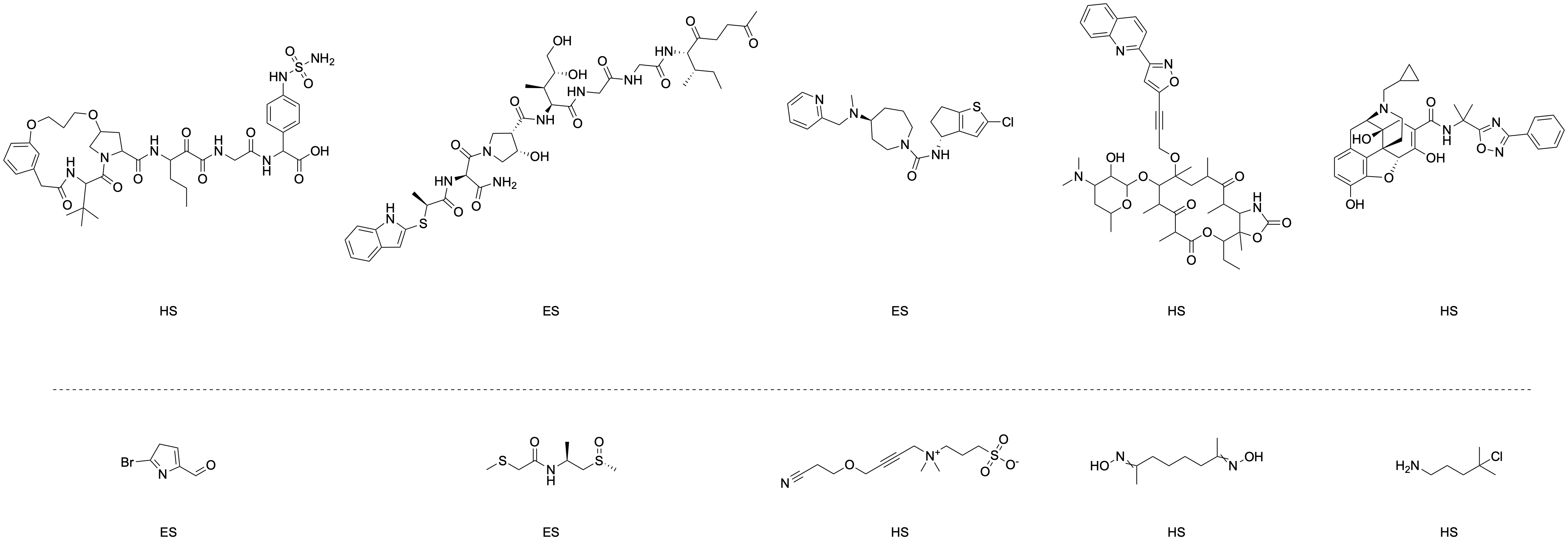}
    \caption{Structures from the MC test set scored by the \textbf{pre-trained} graph (\textbf{GGLGGL}) model. \textit{Upper row:} Low FSscore (low synthetic feasibility). \textit{Lower row:} High FSscore (high synthetic feasibility). The labels below indicate the ground truth.}
    \label{fig:SI_strucs_MC_graph_pt}
\end{figure}

\begin{figure}[H]
    \centering
    \includegraphics[width=1\linewidth]{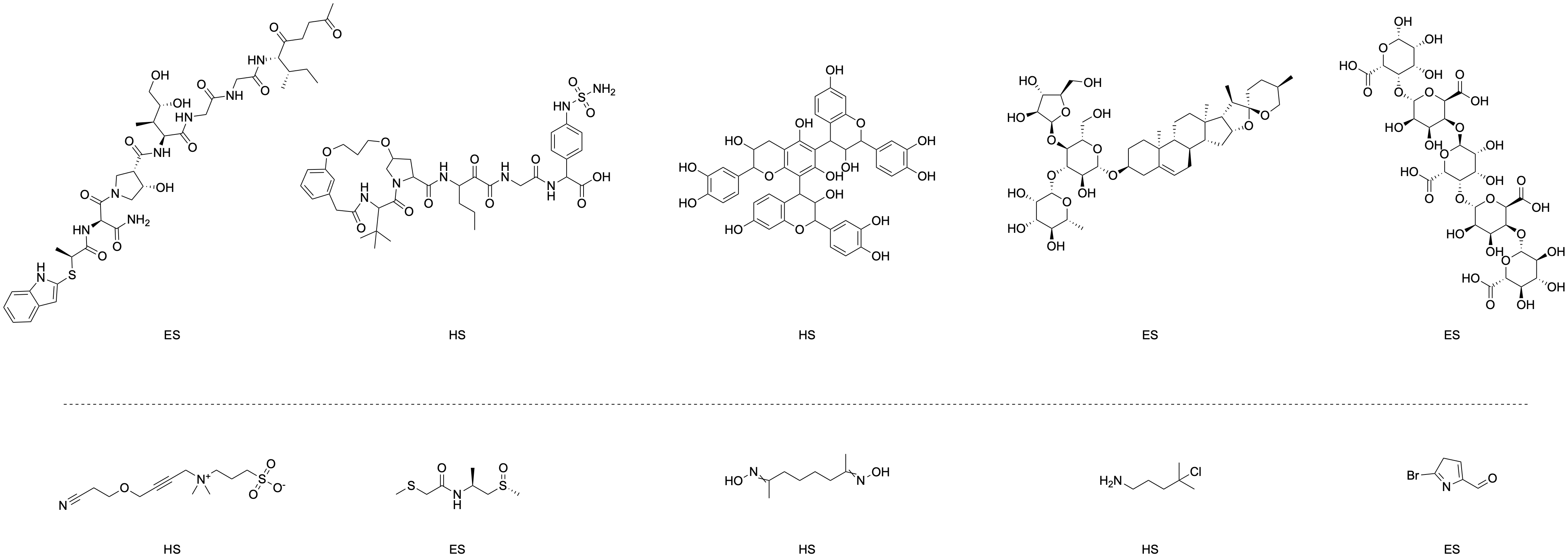}
    \caption{Structures from the MC test set scored by the \textbf{pre-trained} fingerprint (\textbf{Morgan counts}) model. \textit{Upper row:} Low FSscore (low synthetic feasibility). \textit{Lower row:} High FSscore (high synthetic feasibility). The labels below indicate the ground truth.}
    \label{fig:SI_strucs_MC_fp_pt}
\end{figure}

\begin{figure}[H]
    \centering
    \includegraphics[width=1\linewidth]{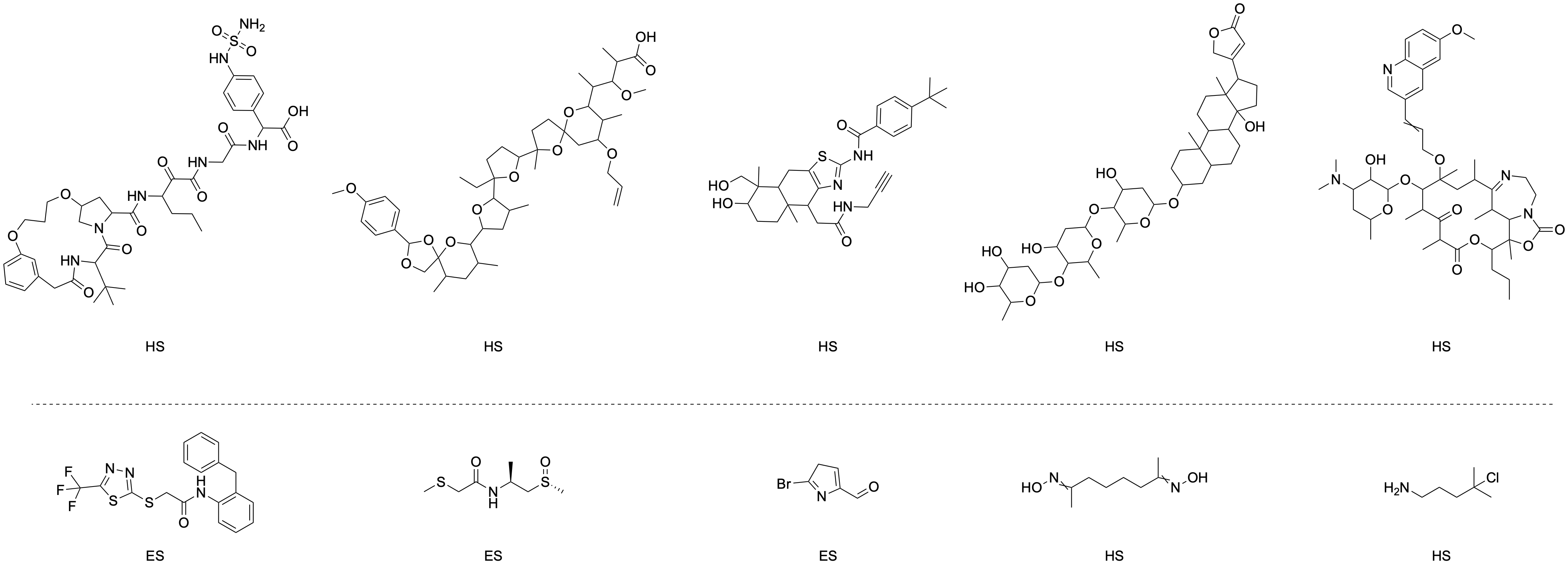}
    \caption{Structures from the MC test set scored by the \textbf{fine-tuned} graph (\textbf{GGLGGL}) model using 30 pairs. \textit{Upper row:} Low FSscore (low synthetic feasibility). \textit{Lower row:} High FSscore (high synthetic feasibility). The labels below indicate the ground truth.}
    \label{fig:SI_strucs_MC_graph_ft}
\end{figure}

\begin{figure}[H]
    \centering
    \includegraphics[width=1\linewidth]{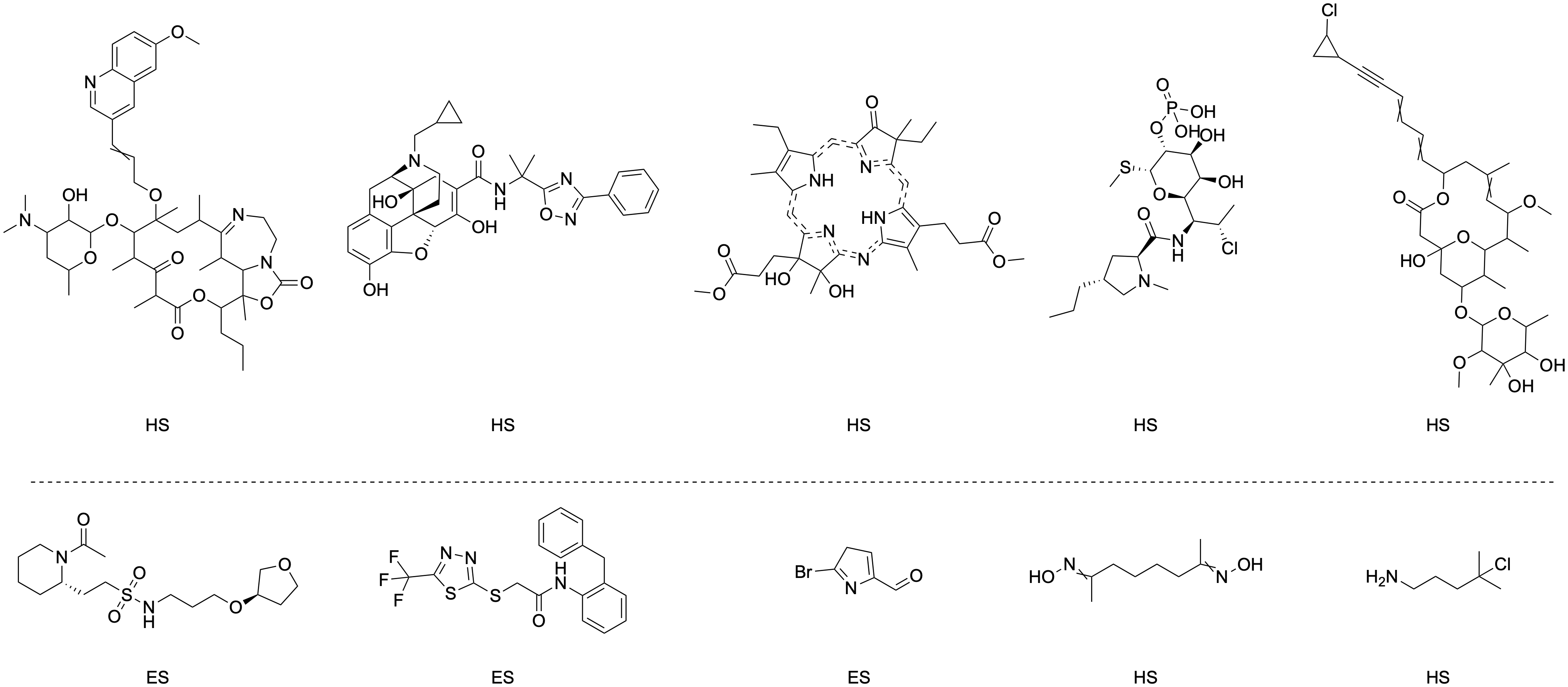}
    \caption{Structures from the MC test set scored by the \textbf{fine-tuned} fingerprint (\textbf{Morgan counts}) model using 30 pairs. \textit{Upper row:} Low FSscore (low synthetic feasibility). \textit{Lower row:} High FSscore (high synthetic feasibility). The labels below indicate the ground truth.}
    \label{fig:SI_strucs_MC_fp_ft}
\end{figure}

\subsection{PROTAC-DB}
\begin{figure}[H]
    \centering
    \includegraphics[width=1\linewidth]{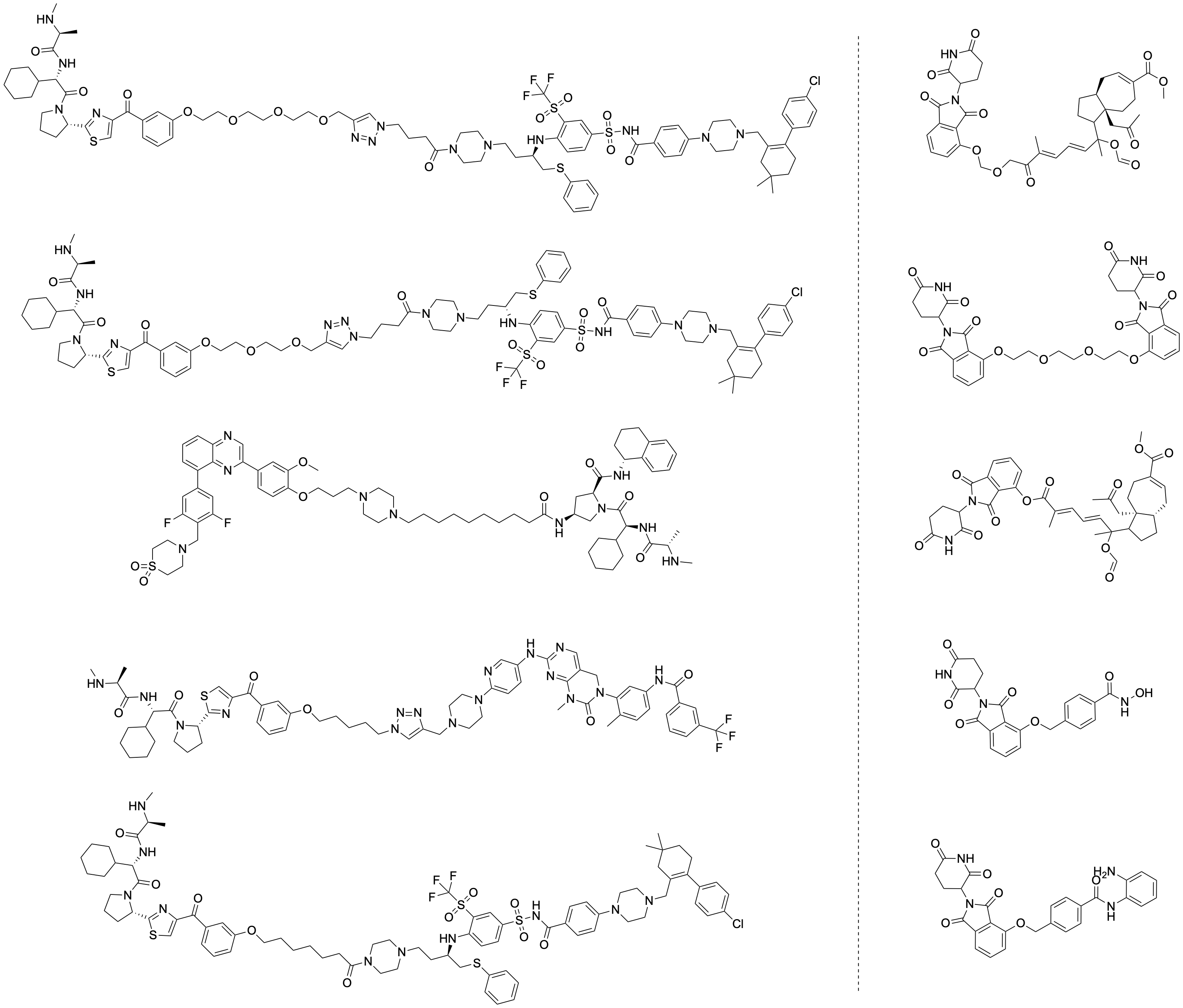}
    \caption{Structures from PROTAC-DB scored by the \textbf{pre-trained} graph (\textbf{GGLGGL}) model. \textit{Upper row:} Low FSscore (low synthetic feasibility). \textit{Lower row:} High FSscore (high synthetic feasibility).}
    \label{fig:SI_strucs_protac_pt}
\end{figure}

\begin{figure}[H]
    \centering
    \includegraphics[width=1\linewidth]{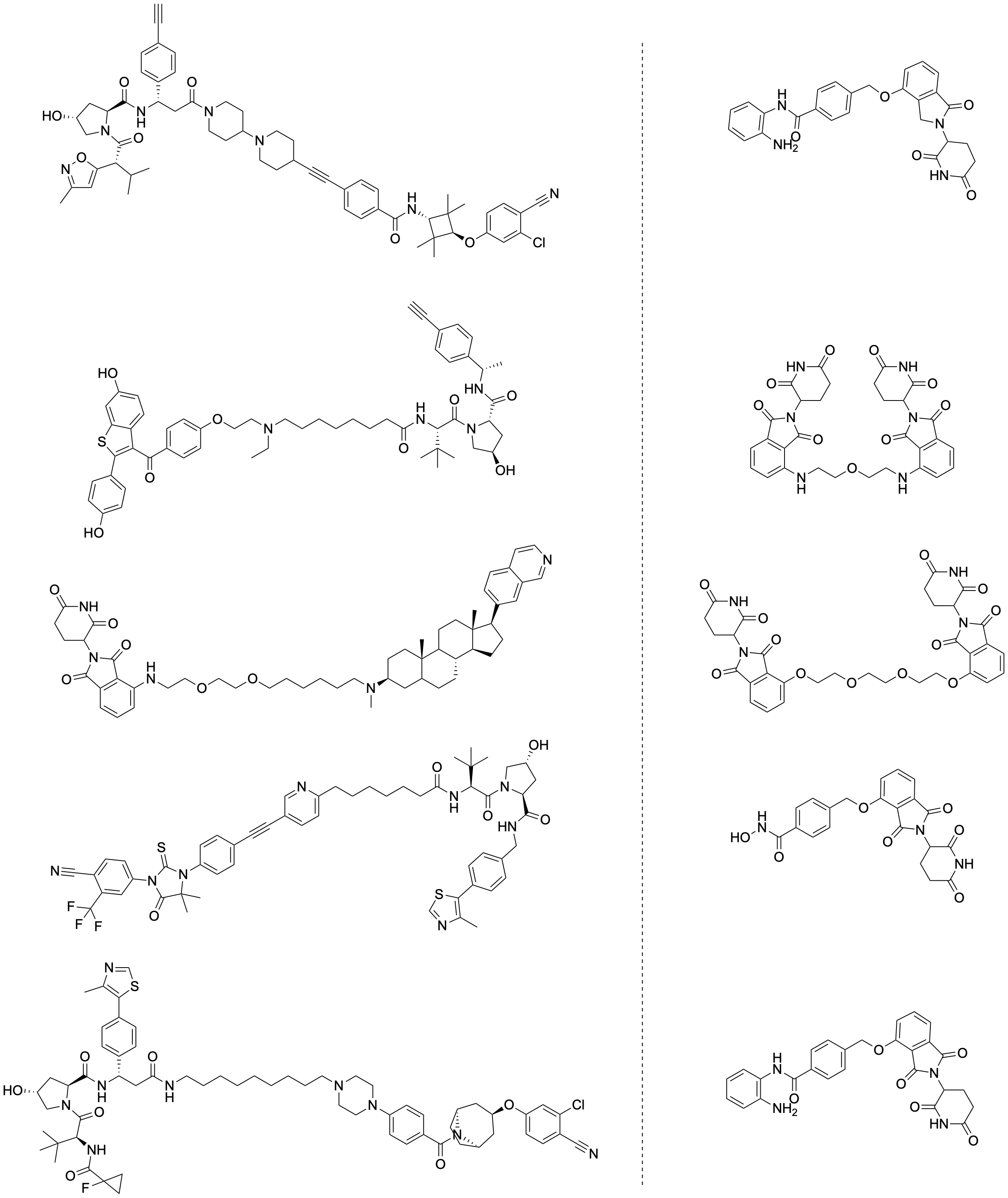}
    \caption{Structures from PROTAC-DB scored by the \textbf{fine-tuned} graph (\textbf{GGLGGL}) model using 50 pairs. \textit{Upper row:} Low FSscore (low synthetic feasibility). \textit{Lower row:} High FSscore (high synthetic feasibility).}
    \label{fig:SI_strucs_protac_ft}
\end{figure}

\subsection{REINVENT case study}
\begin{figure}[H]
    \centering
    \includegraphics[width=1\linewidth]{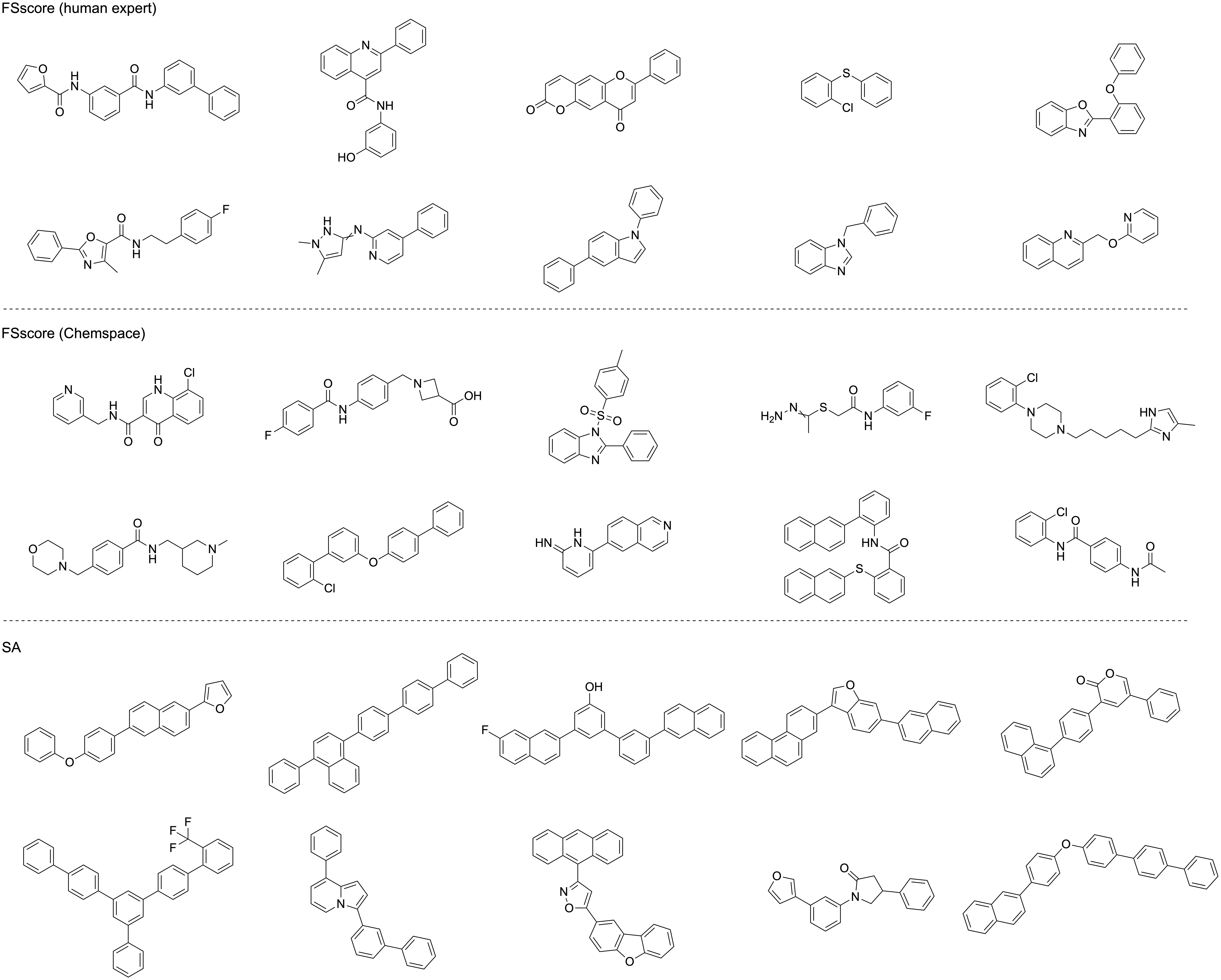}
    \caption{Random Samples from the three agents trained in the REINVENT case study. \textit{Upper row:} Samples from the first agent optimized for docking to DRD2. \textit{Middle row:} Samples from agent optimized to the FSscore.\textit{Lower row:} Samples from agent optimized for SA score.}
    \label{fig:SI_strucs_reinvent}
\end{figure}

\section{Learning curves}
\label{sec:SI_learning_curves}

\begin{figure}[H]
    \centering
    \includegraphics[width=1\linewidth]{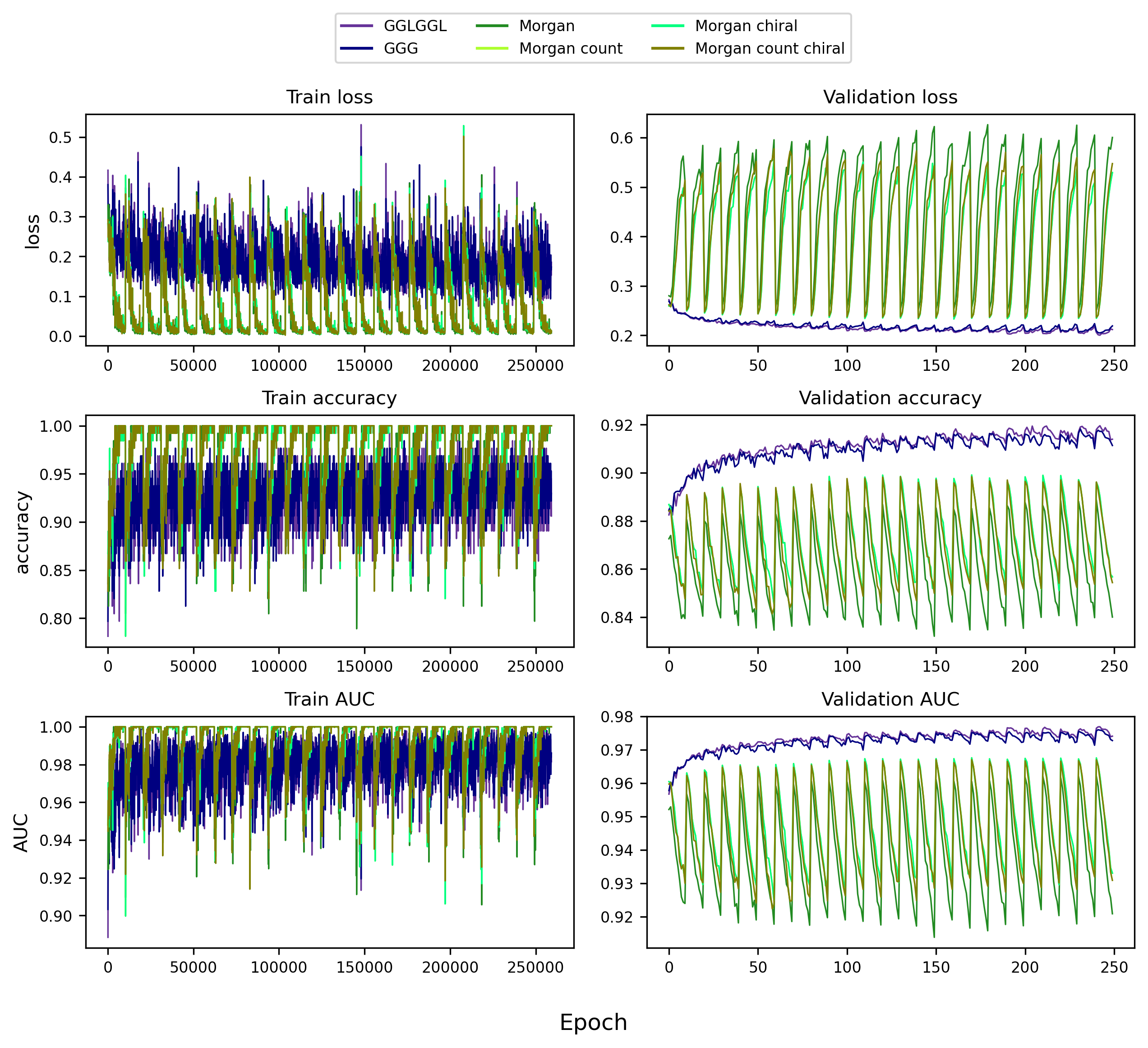}
    \caption{Learning curves of the training and validation set for all model variations during \textbf{pre-training} for 250 epochs. The training set was split into 25~subsets and the model was trained on each of them for 10~epochs explaining the oscillating nature of the curves.}
    \label{fig:SI_lc_pt}
\end{figure}

\begin{figure}[H]
    \centering
    \includegraphics[width=1\linewidth]{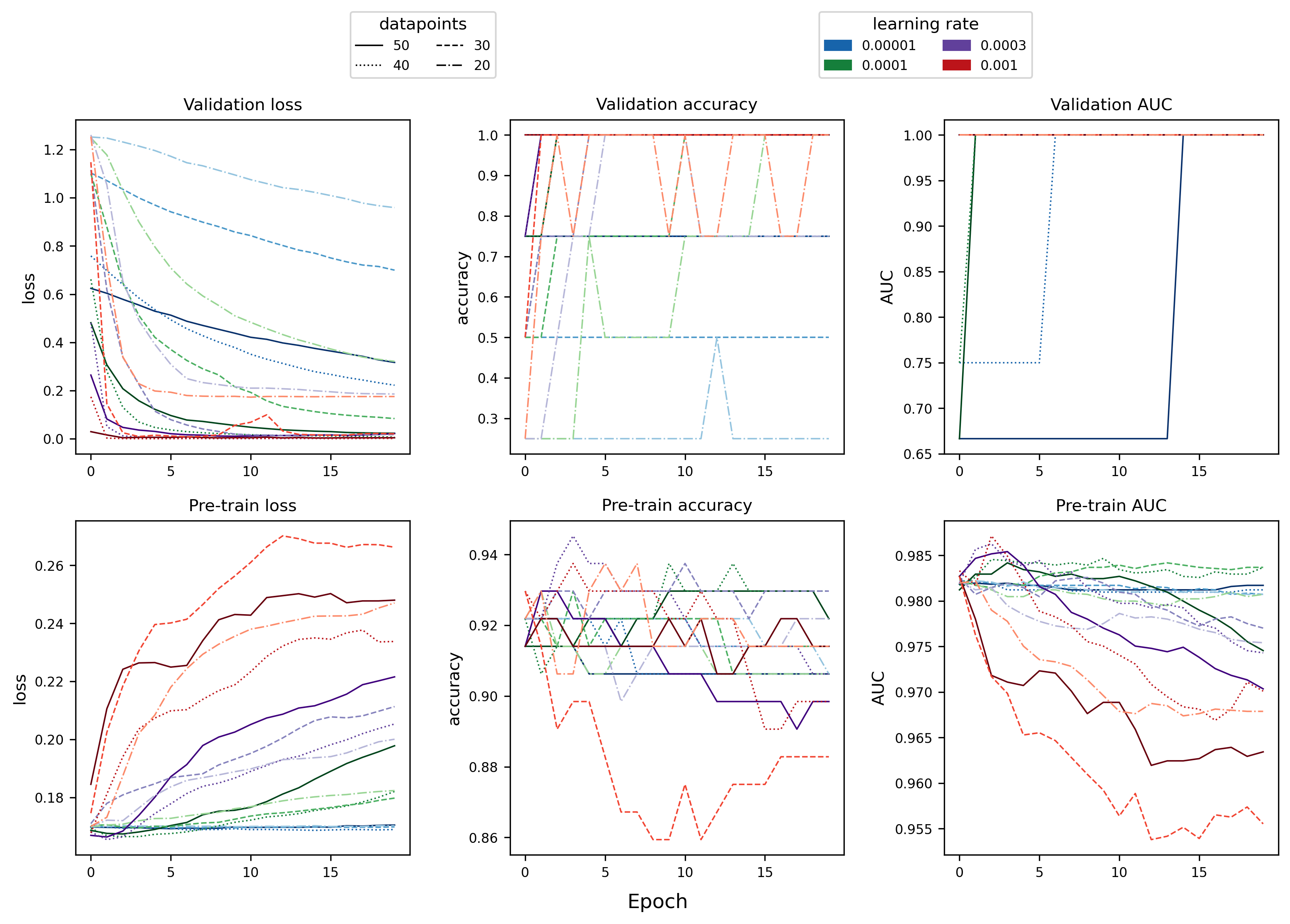}
    \caption{Learning curves of fine-tuning the graph-based (GGLGGL) model with varying learning rate and number of pairs used from the \textbf{chirality} test set. \textit{Upper row:} Training metrics. \textit{Middle row:} Validation metrics. \textit{Lower row:} Metrics on 5,000 samples from the pre-training test set.}
    \label{fig:SI_lc_chirality_graph}
\end{figure}

\begin{figure}[H]
    \centering
    \includegraphics[width=1\linewidth]{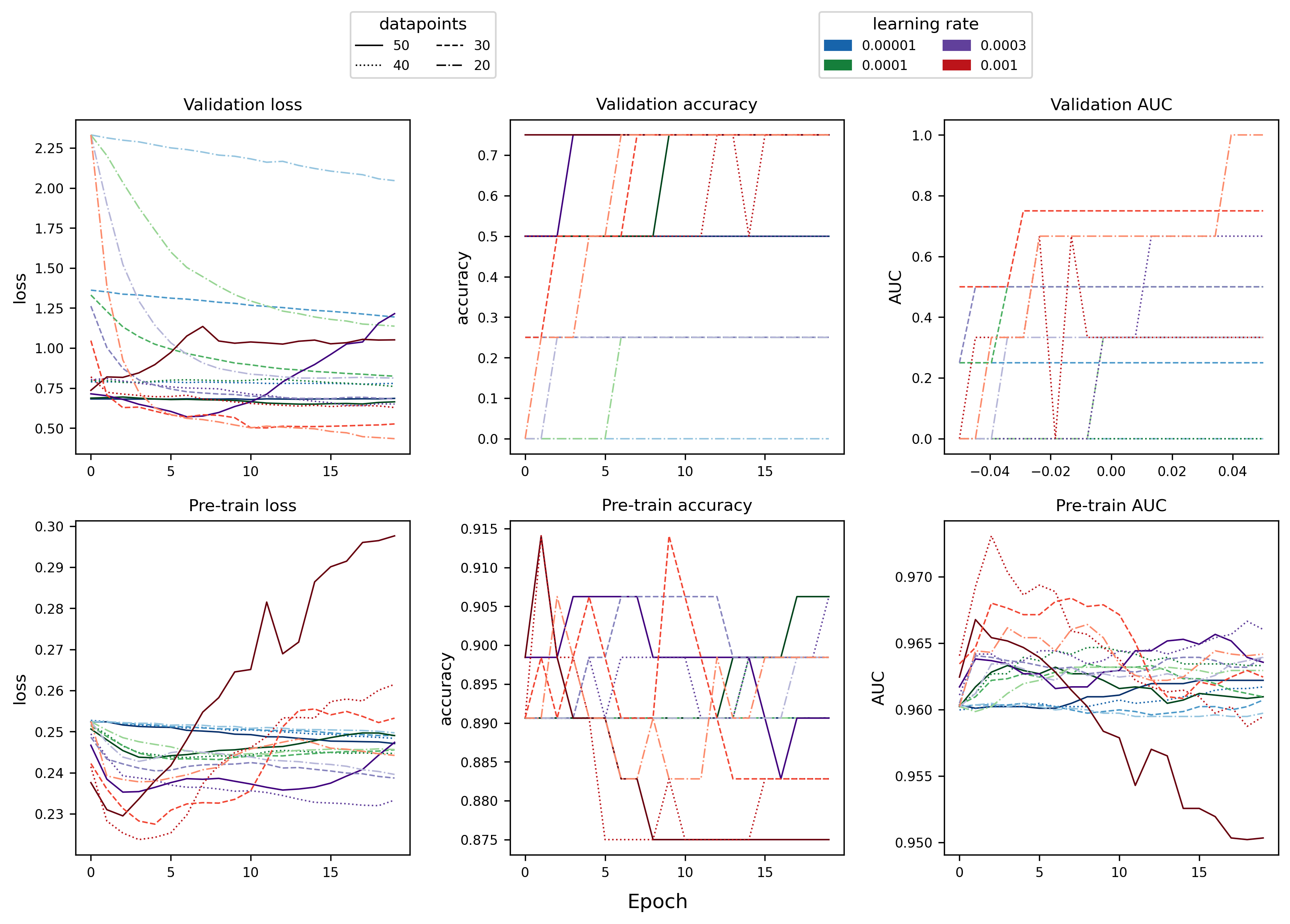}
    \caption{Learning curves of fine-tuning the fp-based (Morgan chiral count) model with varying learning rate and number of pairs used from the \textbf{chirality} test set. \textit{Upper row:} Training metrics. \textit{Middle row:} Validation metrics. \textit{Lower row:} Metrics on 5,000 samples from the pre-training test set.}
    \label{fig:SI_lc_chirality_fp}
\end{figure}

\begin{figure}[H]
    \centering
    \includegraphics[width=1\linewidth]{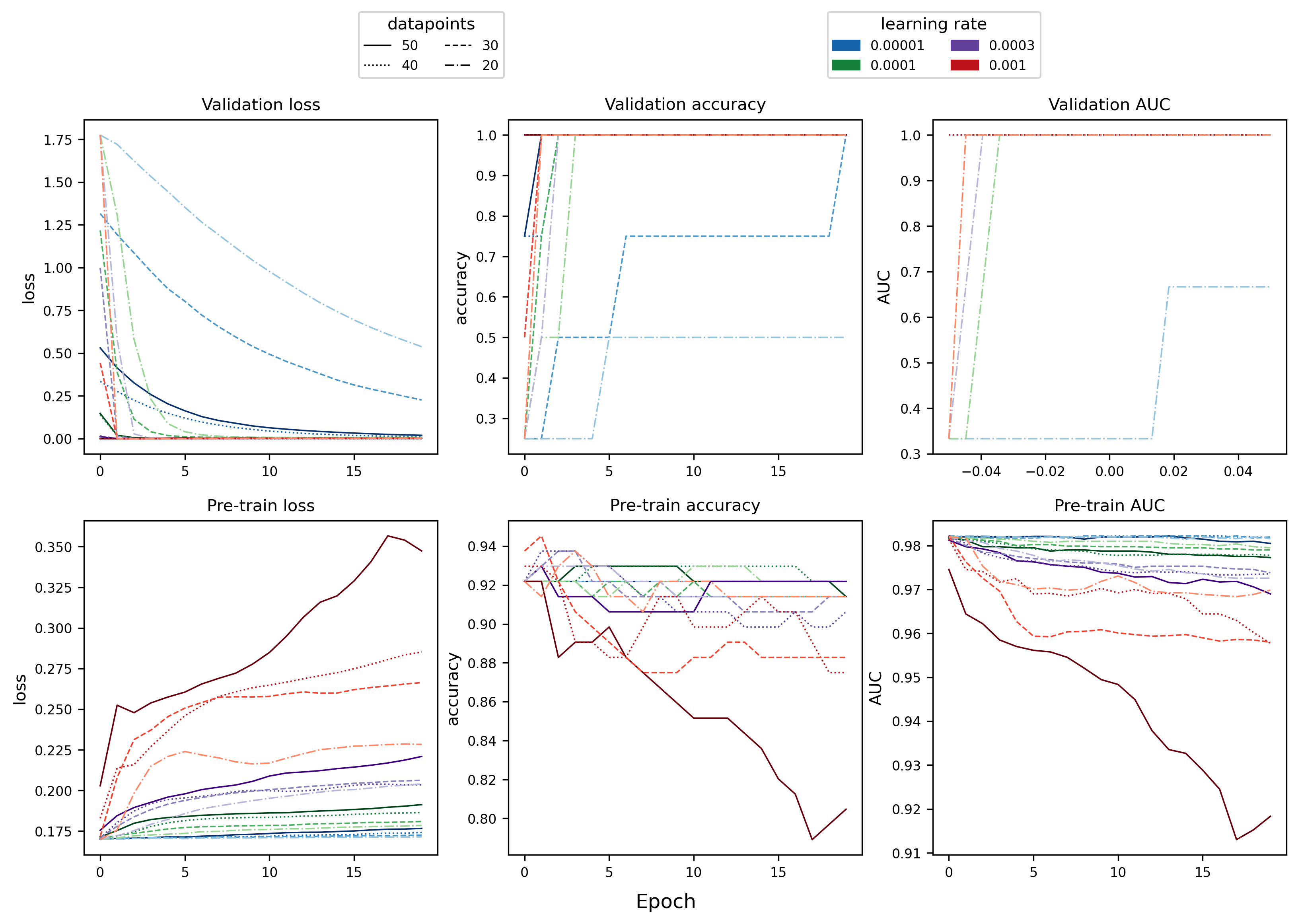}
    \caption{Learning curves of fine-tuning the graph-based (GGLGGL) model with varying learning rate and number of pairs used from the \textbf{CP} test set. \textit{Upper row:} Training metrics. \textit{Middle row:} Validation metrics. \textit{Lower row:} Metrics on 5,000 samples from the pre-training test set.}
    \label{fig:SI_lc_CP_graph}
\end{figure}

\begin{figure}[H]
    \centering
    \includegraphics[width=1\linewidth]{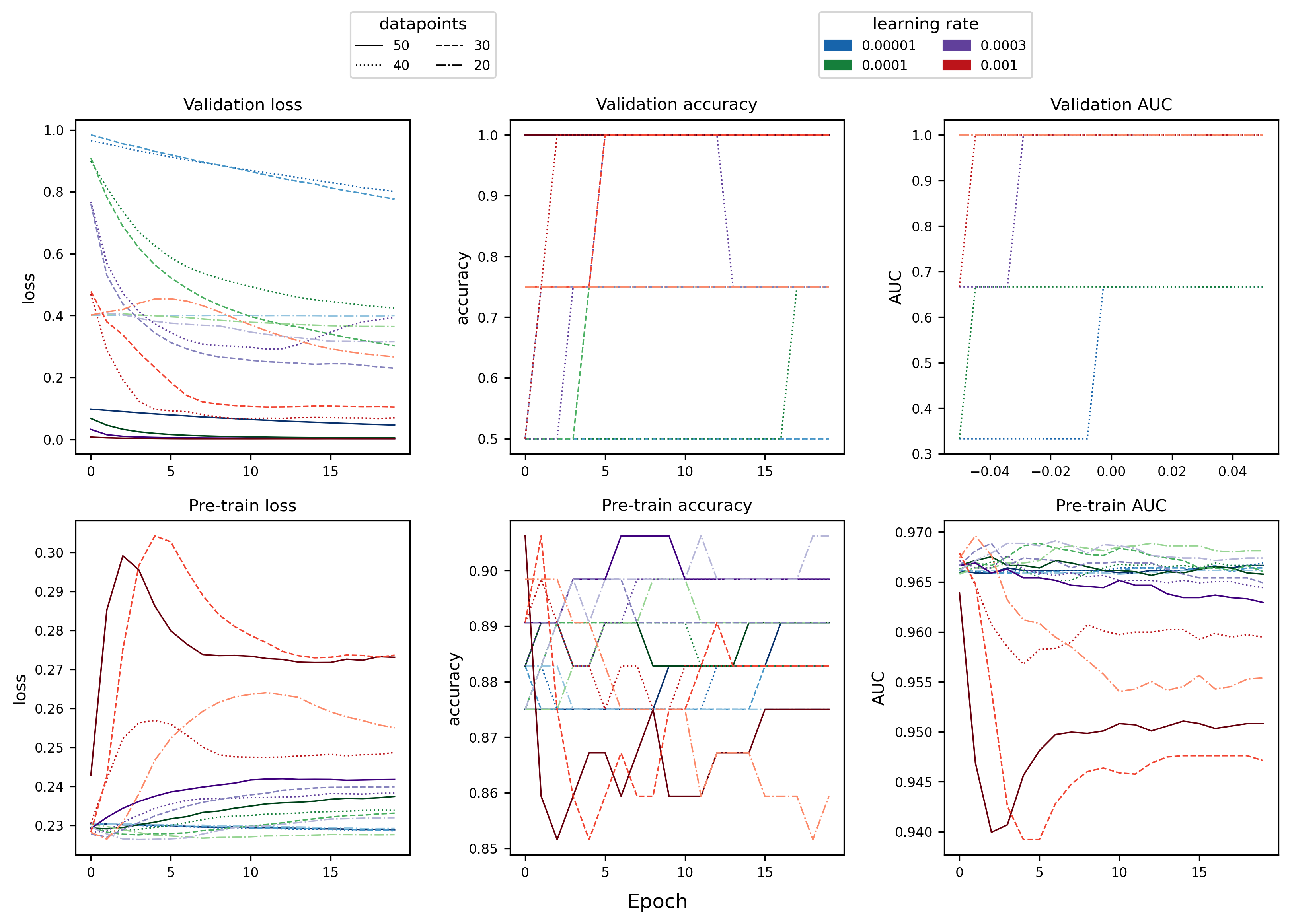}
    \caption{Learning curves of fine-tuning the fp-based (Morgan count) model with varying learning rate and number of pairs used from the \textbf{CP} test set. \textit{Upper row:} Training metrics. \textit{Middle row:} Validation metrics. \textit{Lower row:} Metrics on 5,000 samples from the pre-training test set.}
    \label{fig:SI_lc_CP_fp}
\end{figure}

\begin{figure}[H]
    \centering
    \includegraphics[width=1\linewidth]{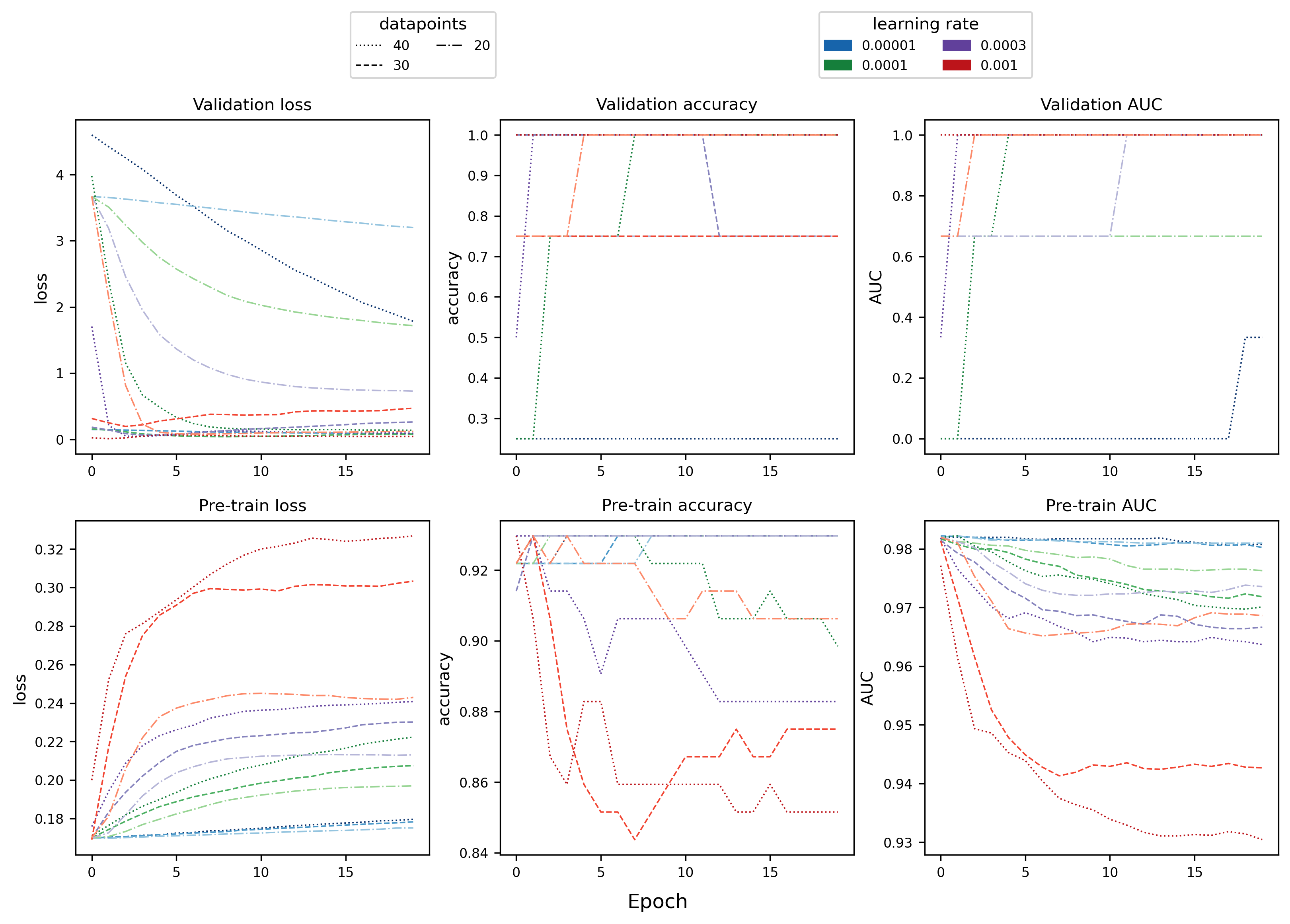}
    \caption{Learning curves of fine-tuning the graph-based (GGLGGL) model with varying learning rate and number of pairs used from the \textbf{MC} test set. \textit{Upper row:} Training metrics. \textit{Middle row:} Validation metrics. \textit{Lower row:} Metrics on 5,000 samples from the pre-training test set.}
    \label{fig:SI_lc_MC_graph}
\end{figure}

\begin{figure}[H]
    \centering
    \includegraphics[width=1\linewidth]{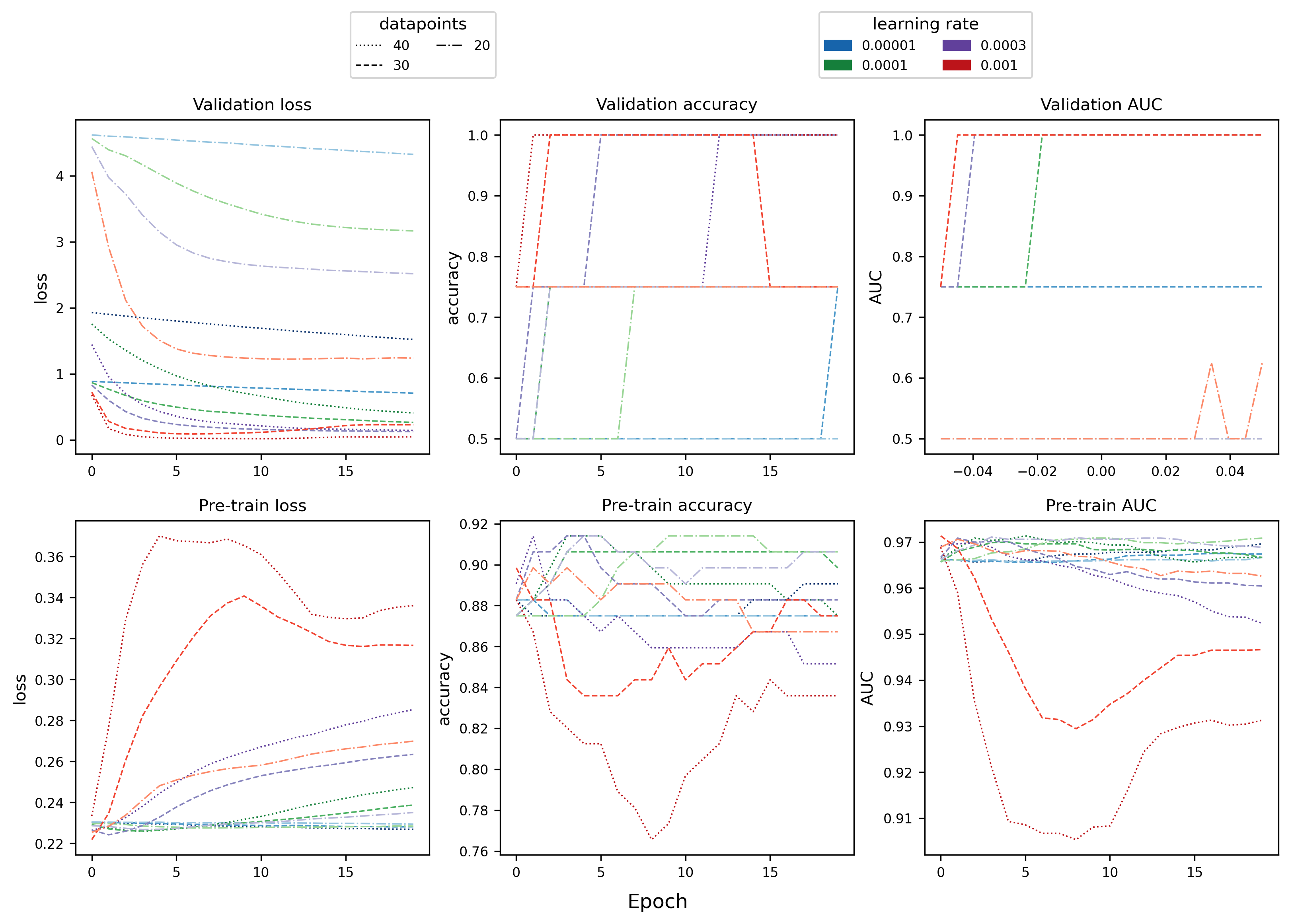}
    \caption{Learning curves of fine-tuning the fp-based (Morgan count) model with varying learning rate and number of pairs used from the \textbf{MC} test set. \textit{Upper row:} Training metrics. \textit{Middle row:} Validation metrics. \textit{Lower row:} Metrics on 5,000 samples from the pre-training test set.}
    \label{fig:SI_lc_MC_fp}
\end{figure}

\begin{figure}[H]
    \centering
    \includegraphics[width=1\linewidth]{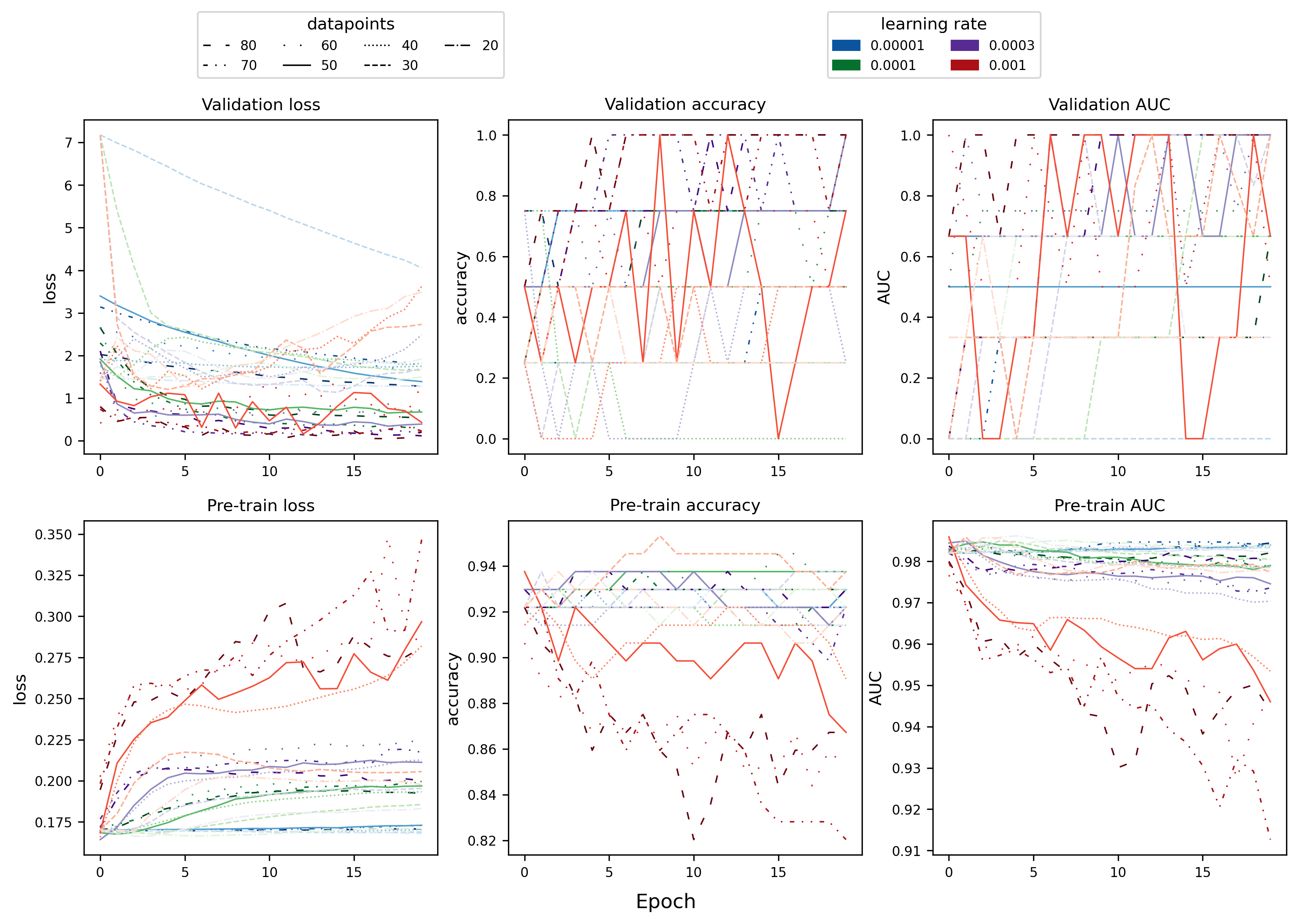}
    \caption{Learning curves of fine-tuning the graph-based (GGLGGL) model with varying learning rate and number of pairs used from the \textbf{PROTAC-DB} sample. \textit{Upper row:} Training metrics. \textit{Middle row:} Validation metrics. \textit{Lower row:} Metrics on 5,000 samples from the pre-training test set.}
    \label{fig:SI_lc_protacdb}
\end{figure}

\begin{figure}[H]
    \centering
    \includegraphics[width=1\linewidth]{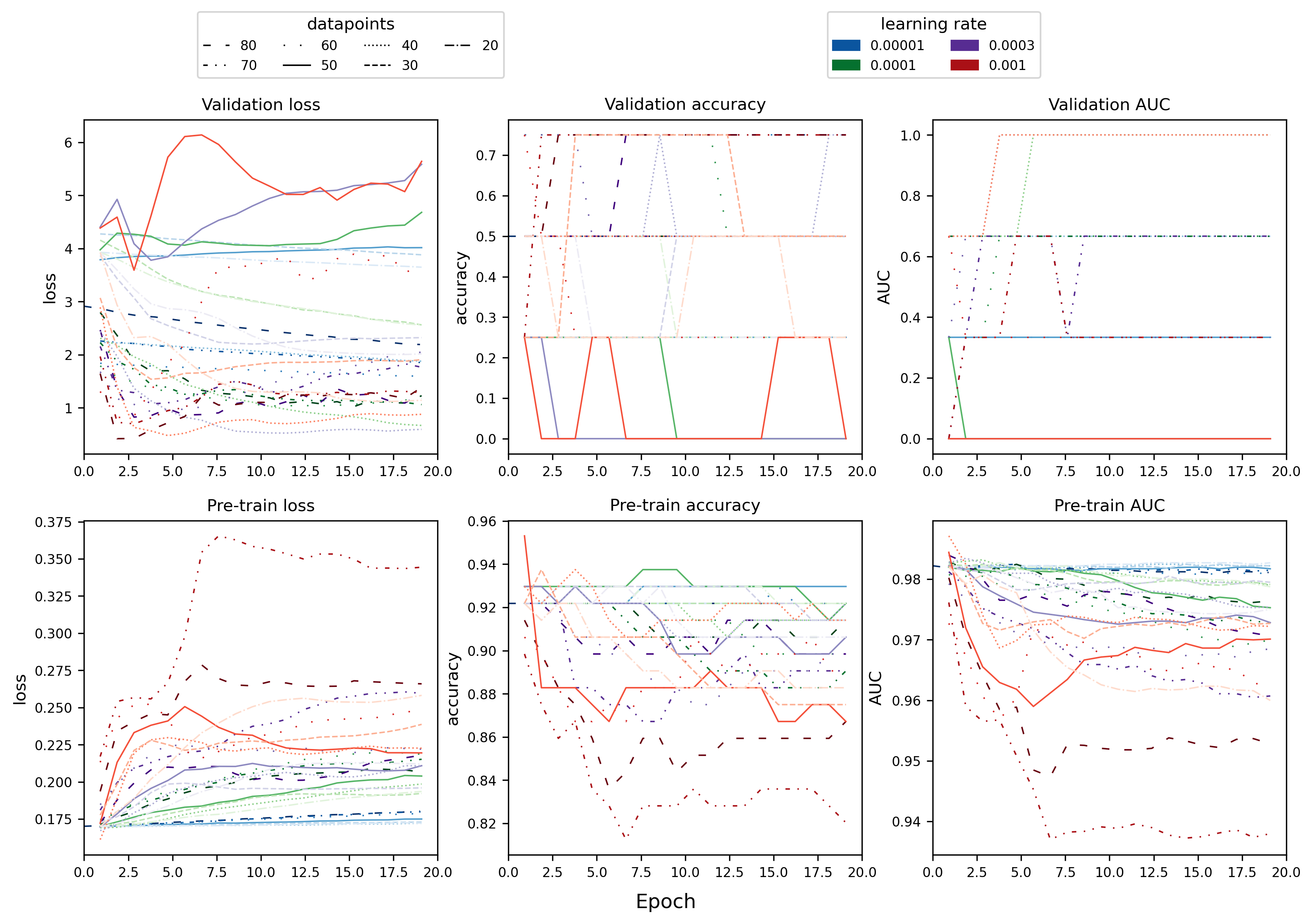}
    \caption{Learning curves of fine-tuning the graph-based (GGLGGL) model with varying learning rate and number of pairs used from the \textbf{REINVENT} case study. \textit{Upper row:} Training metrics. \textit{Middle row:} Validation metrics. \textit{Lower row:} Metrics on 5,000 samples from the pre-training test set.}
    \label{fig:SI_lc_reinvent}
\end{figure}

\bibliography{ref}